\documentclass[manuscript,screen]{acmart}

\usepackage{enumitem}
\usepackage{pifont}
\usepackage{multirow} 
\usepackage{graphicx}
\usepackage{subcaption}
\usepackage[normalem]{ulem}
\usepackage{xcolor}
\usepackage{xspace}
\usepackage{array}

\newcommand{\kpd}{\texttt{KP3D}\xspace}
\newcommand{\orbslamS}{\texttt{ORB-SLAM3} \texttt{(S)}\xspace}
\newcommand{\orbslamM}{\texttt{ORB-SLAM3} \texttt{(M)}\xspace}
\newcommand{\dvfo}{\texttt{DVFO}\xspace}
\newcommand{\deepvo}{\texttt{DeepVO}\xspace}

\renewcommand\footnotetextcopyrightpermission[1]{}

\usepackage{tikz}
\definecolor{colorbullets}{HTML}{D6E8D5}

\newcommand*\circled[1]{\tikz[baseline=(char.base)]{
            \node[shape=circle,draw,inner sep=1pt, fill=colorbullets] (char) {#1};}}

\AtBeginDocument{%
  }

\acmJournal{TOSN}
\acmVolume{1}
\acmNumber{1}
\acmArticle{1}
\acmMonth{1}

\begin{document}

\title[Lost in Tracking Translation]{Lost in Tracking Translation: A Comprehensive Analysis of Visual SLAM in Human-Centered XR and IoT Ecosystems}

\author{Yasra Chandio}
\affiliation{%
 \institution{University of Massachusetts Amherst}
 \country{United States}}
 
\author{Khotso Selialia}
\affiliation{%
 \institution{University of Massachusetts Amherst}
 \country{United States}}
 
 \author{Joseph DeGol}
\affiliation{%
 \institution{Steg AI}
 \country{United States}}
 
 \author{Luis Garcia}
\affiliation{%
 \institution{University of Utah}
 \country{United States}}

\author{Fatima M. Anwar}
\affiliation{%
 \institution{University of Massachusetts Amherst}
 \country{United States}}

\renewcommand{\shortauthors}{Chandio et al.}

\begin{abstract}
Advances in tracking algorithms have enabled applications from steering autonomous vehicles to guiding robots to enhancing augmented reality experiences for users. 
However, these algorithms are application-specific and do not work across applications with different types of motion; even a tracking algorithm designed for a given application does not work in scenarios deviating from highly standard conditions. 
For example, a tracking algorithm designed for indoor robot navigation will not work for tracking the same robot in an outdoor environment. 
To demonstrate this problem, we evaluate the performance of the state-of-the-art tracking methods across various applications and scenarios. 
To inform our analysis, we first categorize environmental, locomotion-related, and algorithmic challenges faced by tracking algorithms. 
We quantitatively evaluate the performance using multiple tracking algorithms and representative datasets for a wide range of Internet of Things (IoT) and Extended Reality (XR) applications, including autonomous vehicles, drones, and humans.
Our analysis shows that no tracking algorithm works across all applications and even in diverse scenarios within the same application. 
Ultimately, using the insights generated from our analysis, we discuss multiple approaches to improving the tracking performance using input data characterization, leveraging intermediate information, and output evaluation. 

\end{abstract}





\maketitle

\section{Introduction}
\label{sec:introduction}
Tracking systems are fundamental to immersive Extended Reality (XR) applications, facilitating accurate and real-time navigation and mapping that are crucial for creating immersive and interactive experiences~\cite{AR-SLAM-survey-cite-billinghurst2015survey, SLAMAR-klein2007parallel}. However, various challenges must be addressed for accurate tracking, particularly in human-centered scenarios like XR~\cite{cadena2016past, why-slam-fail}. 
Tracking systems in XR face additional challenges due to human factors such as unpredictable movements, inter-individual variability, contextual factors, cognitive load, occlusions from body parts, physical safety concerns, adaptive requirements, and the need for real-time interaction. These elements introduce layers of unpredictability and complexity, further complicating the tracking process.
These challenges are intertwined, involving (1) the environment's complexity and the inherent limitations of sensing systems, (2) various locomotion demands, and (3) the inherent limitations of tracking systems. As a result, while tracking methods are often presented as generic, their performance significantly varies across different environments, locomotion scenarios, and application settings, such as drones~\cite{burri2016euroc}, autonomous vehicles~\cite{kitti-geiger2013vision}, robotics~\cite{smart-robot}, and other human-centered~\cite{cps-work-luis-human-cyber-safety-crtical, surgery_review,activityaware-fu2021hawatcher} and non-human-centered environments~\cite{aqel2016review-slam-vo-survey, automated-driving-tracking-survey-cvpr-w-milz2018visual}. 

To understand these challenges, it is essential to examine the specific factors contributing to the complexity of broader tracking systems and how human factors add to their complexity.
First, the complexity of the environment can vary with the number of objects, lighting conditions, occlusions, weather, reflective surfaces, and scene changes. For instance, tracking in a crowded urban setting with changing lighting and reflective surfaces is particularly difficult, especially given varying sensing capabilities across different hardware platforms. In XR applications, this complexity is heightened by the unpredictability of human interactions and the dynamic nature of the environment. Additionally, humans can seamlessly transition between different environments, such as walking from a room to a corridor to the outdoors, without a break. This continuous movement across varied settings introduces additional challenges for tracking systems, as they must constantly adapt to new conditions and maintain accuracy.
Second, locomotion differs across applications. Vehicles, robots, and humans move differently, each posing unique challenges. In human-centric applications, such as XR, abrupt movements can cause blurred images. Even in-vehicle navigation typically involves fewer abrupt movements, maintaining tracking accuracy is difficult due to varying speeds and accelerations in parking lots, urban areas, and highways~\cite{automated-driving-tracking-survey-cvpr-w-milz2018visual}. Tracking human movement adds another layer of complexity, as humans frequently stop, walk, and change speeds unpredictably, making consistent speed maintenance challenging.
Third, sensors such as inertial measurement units (IMU) sensors, depth cameras, and RGB cameras each have specific issues~\cite{chong2015sensor}. IMU sensors can drift over time~\cite{anwar2019securing, walnut}, depth cameras struggle with lighting and reflective surfaces~\cite{depth-camera-issues-halmetschlager2018empirical}, and RGB cameras are affected by lighting variations. In XR applications, these sensor limitations are compounded by the need to integrate data from multiple sources in real-time~\cite{ungureanu2020hololens2r}.

To overcome these challenges, prior work has developed tracking algorithms that leverage various computational approaches. For example, traditional SLAM methods heavily depend on carefully engineered features and manually designed system components~\cite{ORBSLAM3_TRO, DSM-Zubizarreta2020, schops2019badslam, vins-mono-qin2017vins}. These methods often lack robustness and struggle to maintain accuracy and reliability in dynamic and diverse real-world scenarios where conditions can vary significantly. Factors such as changing lighting conditions, moving objects, and varying environmental textures can degrade their performance. Conversely, end-to-end learning approaches~\cite{ZUB18deeptam, bloesch2018codeslam, tartanvo2020corl} learn system components directly from data, which can lead to improved adaptability. However, these approaches can also face robustness issues, as they may fail to generalize when encountering unfamiliar situations or environments not represented in their training data. Hybrid approaches~\cite{tang2020kp3d, gradslam, sfmlearner-zhou2017unsupervised} aim to enhance overall performance by combining traditional and learning-based methods, leveraging both strengths.
While this improves average performance, it can reduce the peak performance of individual approaches.

To comprehensively address these challenges, it is important not only to evaluate tracking systems within XR environments but also to compare their performance against other application domains, such as autonomous vehicles and drones. Evaluating tracking methods across these varied domains provides a broader perspective on the strengths and weaknesses of different approaches. Autonomous vehicles and drones present unique challenges, such as high-speed movement and indoor-outdoor environmental variability, which can inform improvements in XR tracking systems. By understanding how these systems perform in different contexts, we can derive insights that contribute to developing more robust and versatile tracking solutions that can be applied across multiple domains, including but not limited to XR. Additionally, it is crucial to examine how these algorithms behaved in their original use cases~\cite{valente2019improving-visual-challenges, rgbd-survey, automated-driving-tracking-survey-cvpr-w-milz2018visual, deep-vo-survey-, chen2020survey} before XR became prominent. Understanding their foundational performance and limitations in traditional applications will provide a deeper insight into their adaptability and potential enhancements needed for XR environments.

This paper aims to address these challenges by systematically understanding the challenges, technical requirements, bottlenecks, and potential solution directions needed to enhance tracking performance in XR and beyond. In doing so, we make the following contributions:

\begin{enumerate}[topsep=4pt, leftmargin=*, itemsep=0.6ex]
    \item \textbf{Taxonomy of challenges.} We categorize the algorithmic, environmental, and locomotion-related challenges that tracking systems face and their impact on XR applications. This taxonomy provides a structured overview of the difficulties inherent in Visual SLAM tracking by highlighting the specific issues that need to be addressed to improve tracking performance in various human-in-loop and other Internet of Things (IoT) systems.
    
    \item \textbf{Charting tracking performance.} We quantitatively evaluate the performance of state-of-the-art tracking algorithms across three distinct datasets, each representing a different application domain, environment, motion, and tracking target with unique complexities, including representative IoT systems like autonomous vehicles and drones and human-in-the-loop systems such as XR.
    
    \item \textbf{Dataset characterization.} Building on observations from our quantitative evaluation across traditional, end-to-end learning-based, and hybrid tracking systems, we conduct a preliminary proof-of-concept data characterization. This analysis highlights the importance of understanding how dataset properties impact tracking performance and identifies potential adaptive solutions for specific environments and use cases. 
\end{enumerate}

Unlike existing surveys that focus on specific applications or isolated aspects of tracking systems, our comprehensive evaluation empirically examines a broader range of scenarios and system types. This approach systematically presents challenges and performance bottlenecks across diverse contexts, providing a robust foundation for developing adaptable and reliable tracking solutions. These insights are especially valuable for XR applications, where tracking systems must adapt to the unpredictability of human behavior. By addressing current challenges and conducting proof-of-concept case studies, this paper serves as both a reference point for researchers and a springboard for future innovations in Visual SLAM tracking in XR and beyond.

\section{Background and Motivation}
\label{sec:background}
\subsection{XR Tracking vs. Other CPS Systems}
XR tracking systems differ from other cyber-physical systems like autonomous vehicles, drones, and robotics due to their need to integrate virtual and real-world elements in real-time~\cite{what-is-mr-chi}. Each of these systems presents distinct tracking challenges, but XR systems face additional complexities related to human interaction~\cite{interaction-performace-bhargava2018evaluating, abstract-real, biocca-interaction} and environmental variability~\cite{fua2007vision-mr-tracking}.
Autonomous vehicles rely heavily on sensors like GPS, LiDAR, and cameras to navigate and avoid obstacles in dynamic environments~\cite{automated-driving-tracking-survey-cvpr-w-milz2018visual}. While real-time data processing is crucial in both XR and vehicle systems, XR tracking demands more precision and low latency to maintain user immersion and visual coherence between virtual and real-world elements~\cite{perfromance-presence, context}. Unlike vehicle systems, which only focus on navigating roads and traffic~\cite{automated-driving-tracking-survey-cvpr-w-milz2018visual}, XR systems must blend virtual objects with the real world in a visually coherent manner, necessitating accurate spatial understanding~\cite{context, man-in-the-middle-attack-sluganovic2020tap, spatial-misalogment, huapple-vs-meta-spatial-tracking-XR-maria} and low latency to maintain immersion~\cite{perfromance-presence}.

Similarly, drone navigation involves flight control, stabilization, and obstacle avoidance, often in outdoor environments with varying weather conditions and terrains~\cite{drones}. Drones utilize GPS, IMUs, and cameras to perform tasks like localization and mapping. The primary challenges in drone tracking include maintaining stability, managing energy consumption, and ensuring safety in uncontrolled airspaces~\cite{acoustic-dronesson2015rocking}. In contrast, XR tracking systems must handle more complex human interactions and diverse environments~\cite{large-indoor-survey-muravyev2022evaluation, indoor-feature, guo2022lidar-slam-featrue-and-kitti-scene-info}. For example, XR applications often require users to transition seamlessly between different settings~\cite{hubner2020evaluation}, such as moving from a room to a corridor to an outdoor space~\cite{merging-real-vr-lok2004toward, why-slam-fail}, necessitating quick adaptation to varying lighting conditions~\cite{lighting-ost}, occlusions~\cite{occlusion-attack-petit2015remote}, and reflective surfaces~\cite{yan2022dgs-reflective-texture}.
Robotic tracking systems, used in applications ranging from industrial automation~\cite{saeedi2016multiple-robot-survey} to service robots in healthcare~\cite{smart-robot} and hospitality~\cite{zhang2021pose}, rely on sensors like LiDAR, cameras, and ultrasonic sensors to navigate and interact with their environment. While robotic and XR tracking share similarities in sensor usage, the key difference lies in the nature of interaction. XR tracking demands higher precision and real-time processing~\cite{output_reality}. Additionally, XR systems must account for the unpredictability of human behavior~\cite{huapple-vs-meta-spatial-tracking-XR-maria}, such as sudden movements and changes in speed~\cite{why-slam-fail, washigntonxrsecurityap}, adding complexity not typically encountered in robotic tracking~\cite{skarbez2020immersion}.

\subsection{Unique Challenges in XR Tracking} 
Imagine a user exploring an XR-guided tour in a museum~\cite{museums-trunfio2022mixed}. As they approach a detailed exhibit, they make quick, abrupt movements to get a closer look, causing the tracking system to lose accuracy momentarily. The user then moves through a corridor with mixed lighting conditions, further challenging the system. Exiting into the courtyard, the bright sunlight causes reflections and shadows that the depth cameras must adjust to. Finally, the user walks into a park, where moving objects like trees and other visitors introduce occlusions, requiring the system to recalibrate constantly.
This example scenario illustrates the complexity of developing robust and accurate XR tracking systems capable of handling dynamic environments, varied locomotion~\cite{yu2021visual-locomotion} and a range of human factors, including:

\noindent \textbf{Unpredictable Movements:} Human actions are often abrupt and erratic, such as sudden turns or rapid gestures, which can disrupt tracking accuracy. Users can make sudden turns, quick gestures, or rapid changes in walking speed, which can momentarily disrupt the tracking system's accuracy. XR systems must quickly adapt to these changes to maintain an immersive experience~\cite{cadena2016past}. 

\noindent \textbf{Inter-individual Variability:} Users interact differently with XR systems, depending on their familiarity with the technology, physical abilities, and personal preferences~\cite{chandio-vr-24-human-factors}. XR tracking must accommodate this variability to deliver a consistent and intuitive user experience~\cite{cps-work-luis-human-cyber-safety-crtical}.

\noindent \textbf{Environmental Variability:} XR systems operate in diverse environments, ranging from well-lit indoor spaces to outdoor areas with changing weather conditions. This variability demands constant recalibration without disrupting user interaction~\cite{hubner2020evaluation, yu2021visual-locomotion}. XR applications often require real-time adaptation to changes in the environment or user activity, placing demands on system responsiveness~\cite{context-of-adaptation}.

\noindent \textbf{Cognitive Load and Real-Time Processing:} Complex interactions in XR applications can increase cognitive load, requiring tracking systems to be intuitive and minimally intrusive to avoid adding strain~\cite{gamers-enhanced-cognitive-ablities-green2003action}. Low latency is critical to XR immersion~\cite{gonzalez2017immersive}. Efficient algorithms that manage real-time data processing, feature extraction, and environmental mapping are necessary to avoid discomfort or disorientation.

\noindent \textbf{Occlusions from Body Parts:} In XR, users' hands, arms, or other body parts can occlude sensors, interrupting tracking. The system must recalibrate quickly to regain accuracy~\cite{occlusion-attack-petit2015remote}.

\noindent \textbf{Physical Safety Concerns:} XR systems must maintain situational awareness to prevent collisions or unsafe interactions with real-world objects~\cite{fan2024cueing-trajectory-safety-prediction}, ensuring users' physical safety during immersive experiences~\cite{Chandio2024-Safetyphysical-cognitive-presence}. This also involves mitigating risks from internal system errors and external attacks~\cite{chandio-aivr-2024, Chandio2020SpatiotemporalSecurity-poster-sensys}, which can compromise the system's accuracy or lead to unsafe scenarios~\cite{washigntonxrsecurityap}.

These challenges emphasize the complexity of XR tracking, which must integrate human factors, environmental variability, and real-time data processing to ensure accurate, immersive experiences. Algorithms must handle environmental changes~\cite{chen2021dynanet}, manage scale~\cite{tartanvo2020corl} and depth perception~\cite{depth-perception-diaz2017designing}, and perform efficient, real-time feature extraction~\cite{amraoui2019features-survey-vslam} and association~\cite{Chandio2024-nfex-IROS, amraoui2019features-survey-vslam}.

\begin{table}[]
\centering
\scriptsize
\caption{\textbf{\emph{Summary of studies conducting a comparative analysis of tracking methods. Terminologies: deep learning, empirical, method-level, dataset-level,
sequence-level, sample-level, and environmental representation.}}}
\vspace{-0.25cm}
\begin{tabular}{|c|c|c|c|c|cccc|}
\hline
\multicolumn{1}{|c|}{\multirow{2}{*}{\textbf{Reference}}} & \multicolumn{1}{c|}{\multirow{2}{*}{\textbf{\begin{tabular}[c]{@{}c@{}}SLAM\\ /VO\end{tabular}}}} & \multicolumn{1}{c|}{\multirow{2}{*}{\textbf{Deep Learning}}} & \multicolumn{1}{c|}{\multirow{2}{*}{\textbf{Empirical}}} & \multicolumn{1}{c|}{\multirow{2}{*}{\textbf{Environmental}}} & \multicolumn{4}{c|}{\textbf{Generalizability}} \\ \cline{6-9} 
\multicolumn{1}{|c|}{} & \multicolumn{1}{c|}{} & \multicolumn{1}{c|}{} & \multicolumn{1}{c|}{} & \multicolumn{1}{c|}{} & \multicolumn{1}{c|}{\textbf{Method}} & \multicolumn{1}{c|}{\textbf{Dataset}} & \multicolumn{1}{c|}{\textbf{Sequence}} & \multicolumn{1}{c|}{\textbf{Sample}} \\ \hline \hline
 ~\cite{fuentes2015visual-survey,aqel2016review-slam-vo-survey}& VO &   &  &  & \multicolumn{1}{l|}{} & \multicolumn{1}{l|}{} & \multicolumn{1}{l|}{} &  \\ \hline
 ~\cite{deep-vo-survey-jeong2021comparison,deep-vo-survey-}& VO &  \checkmark  & \checkmark &  & \multicolumn{1}{l|}{} & \multicolumn{1}{l|}{} & \multicolumn{1}{l|}{} &  \\ \hline
  ~\cite{dataset-characterization-ali2022we, saeedi2019characterizing,liu2021simultaneous-datasets-survey}& SLAM &  & \checkmark &  & \multicolumn{1}{l|}{} & \multicolumn{1}{l|}{\checkmark} & \multicolumn{1}{l|}{} &  \\ \hline
\multicolumn{1}{|c|}{\cite{rgbd-survey, rogers2020test-your-slam, dataset-characterization-ali2022we}} & \multicolumn{1}{c|}{SLAM} & \multicolumn{1}{c|}{} & \multicolumn{1}{c|}{\checkmark} & \multicolumn{1}{c|}{} & \multicolumn{1}{c|}{\checkmark} & \multicolumn{1}{c|}{} & \multicolumn{1}{c|}{} & \multicolumn{1}{c|}{} \\ \hline
 ~\cite{traditional-semantic-survey,azzam2020feature,loop-clouser-arshad2021role,tsintotas2022revisiting-slam-loop-clouser,aulinas2008slam-filtering-appraches,survey-keyframe-younes2017keyframe,saeedi2016multiple-robot-survey}& SLAM &  &  &  & \multicolumn{1}{l|}{\checkmark} & \multicolumn{1}{l|}{} & \multicolumn{1}{l|}{} &  \\ \hline
 \multicolumn{1}{|c|}{\cite{durrant2006simultaneous-part-1-survey,bailey2006simultaneous-part2-survey,boal2014topological-survey,macario2022comprehensive-survey,liu2021simultaneous-datasets-survey,taheri2021slam-survey,grisetti2010tutorial-slam,huang2016critique-survey}} & \multicolumn{1}{c|}{SLAM} & \multicolumn{1}{c|}{} & \multicolumn{1}{c|}{} & \multicolumn{1}{c|}{} & \multicolumn{1}{c|}{} & \multicolumn{1}{c|}{} & \multicolumn{1}{c|}{} & \multicolumn{1}{c|}{} \\ \hline
 \cite{automated-driving-tracking-survey-cvpr-w-milz2018visual,li2018ongoing-evolution-survey,chen2020survey,survey-sualeh2019simultaneous,li2022overview,mokssit2023deep,taketomi2017visual-survey-2010-to-2016,cadena2016past}& Both & \checkmark &  &\checkmark  & \multicolumn{1}{l|}{} & \multicolumn{1}{l|}{} & \multicolumn{1}{l|}{} &  \\ \hline
 ~\cite{bresson2017simultaneous-trends-AVs,sfm-dynamic-survey-saputra2018visual,large-indoor-survey-muravyev2022evaluation,drones6040085,chong2015sensor,servieres2021visual-vislam}& Both &  &  & \checkmark & \multicolumn{1}{l|}{} & \multicolumn{1}{l|}{} & \multicolumn{1}{l|}{} &  \\ \hline
  ~\cite{lidar-slam-survey-garigipati2022evaluation,li2022timing-slam-survey, zhao2020closedloop-system0benchmark-survey, system-level-quant-analysis-karthik-2022,nardi2015introducing-SLAMBENCh,survey-robustness-prokhorov2019measuring}&  Both&  & \checkmark &  \checkmark& \multicolumn{1}{l|}{} & \multicolumn{1}{l|}{} & \multicolumn{1}{l|}{} &  \\ \hline
  \hline
 \textbf{Ours}&\textbf{ Both} & \textbf{ \checkmark}&\textbf{\checkmark}  & \textbf{\checkmark} & \multicolumn{1}{l|}{\textbf{\checkmark}} & \multicolumn{1}{l|}{\textbf{\checkmark}} & \multicolumn{1}{l|}{\textbf{\checkmark}} & \textbf{\checkmark} \\ \hline \hline
\end{tabular}%
\vspace{-0.5cm}
\label{tab:survey-table}
\end{table}

\subsection{Contributions Beyond Related Work} 
Previous comparative analyses of SLAM tracking methods often focus on specific methodologies or application domains, typically limited to one application such as robotics, drones, or autonomous vehicles. For instance, \cite{deep-vo-survey-jeong2021comparison, deep-vo-survey-} primarily explore visual odometry\footnote{Visual Odometry (VO) estimates the camera’s motion (trajectory) over time by analyzing sequential visual data, without explicitly building or maintaining a global map of the environment.} methods that utilize end-to-end learning, but within narrow, domain-specific contexts. In contrast, our analysis takes a more holistic approach, focusing on human-centered tracking in XR environments and highlighting the fundamental differences between conventional tracking algorithms for IoT/CPS systems and XR-based applications.  By evaluating performance across three diverse datasets, two representing IoT/CPS use cases (drones, cars) and one representing an XR use case, we provide a broader perspective on the unique challenges faced in human-in-the-loop tracking systems and other domains. This comparative approach reveals distinct subtleties across different application domains, offering insights that previous studies and surveys have not comprehensively addressed, as summarized in Table~\ref{tab:survey-table}.

First, we construct a \emph{taxonomy designed to classify challenges systematically} with an exploration of \emph{algorithmic}, \emph{environmental}, and \emph{locomotion} challenges in SLAM (\S\ref{sec:overview-challenges}). Unlike traditional SLAM comparisons, we explore how these factors interact within dynamic, unpredictable human-centered environments, such as those encountered in XR. We emphasize the interaction between the environment and movement and its impact on methods. We bridge the gap between traditional SLAM and contemporary end-to-end learning techniques, offering a unified perspective on their evolution and interrelation. Our approach combines both qualitative insights and quantitative evaluations, leading to practical recommendations and future research directions for XR-based tracking systems.

Second, we evaluate the performance of state-of-the-art visual tracking algorithms across applications and datasets \S\ref{sec:analysis}, analyzing four hierarchical dimensions: method, dataset, sequence, and sample levels. We explore trade-offs between SLAM approaches, from classical to deep learning models, in human-centered XR environments. At the dataset and sequence levels, we examine how environmental conditions and transitions between different settings affect tracking accuracy. At the sample level, we assess challenges like lighting changes and unpredictable user movements. We also compare SLAM, deep learning, and hybrid methods, and analyze the impact of human factors on tracking performance.

Third, we propose three strategies to enhance SLAM systems in dynamic environments (\S\ref{sec:discussion}): input profiling, intermediate insights, and output evaluation. Input profiling tailors tracking to specific environmental conditions, while intermediate insights allow for real-time adjustments. Output evaluation refines performance by learning from error patterns, improving system robustness and adaptability across diverse scenarios.

\subsection{Hypothesis Statement}
\label{sec:problem-statement}

We hypothesize that \emph{most tracking algorithms are designed for highly standard operating conditions for specific applications. 
As a result, they perform poorly across applications and scenarios within a given application in real-world settings.
} 
\vspace{-0.2cm}
\section{Visual SLAM: Charting the Uncharted Territories}
\label{sec:overview-challenges}
In this section, we first present an overview of the state-of-the-art tracking approaches that either leverage traditional visual Simultaneous Localization and Mapping (SLAM) methods, end-to-end learning approaches, or their combination. Visual SLAM refers to the task of estimating a camera’s trajectory (pose) while simultaneously constructing a map of the surrounding environment, using visual input such as monocular or stereo images. Unlike conventional SLAM systems that may integrate additional sensors (LiDAR, IMU), Visual SLAM relies primarily on visual data and is widely used in robotics, AR/VR/XR, and IoT applications.
We also detail our key contribution of conceptualizing and devising a taxonomy of challenges. 
The goal of analyzing the tracking methods and categorizing their challenges is to guide our evaluation of tracking approaches in~\autoref{sec:analysis}.

\begin{figure*}[t]
    \centering
    \includegraphics[width=.95\linewidth]{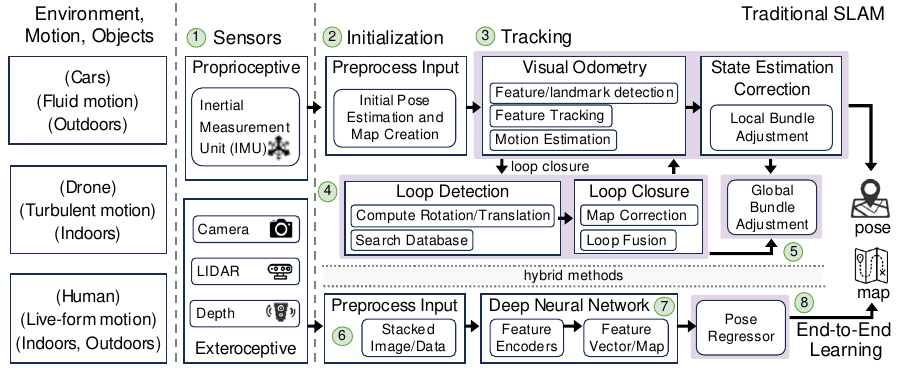} 
    \vspace{-0.35cm}
    \caption{\textbf{\emph{An overview of traditional SLAM- and end-to-end learning-based tracking methods and their components.}}}
    \vspace{-0.3cm}
    \label{fig:overview}
\end{figure*}

\subsection{Visual SLAM Tracking} 
\label{sec:tracking-approaches}

SLAM is a technique used by a wide range of subjects, such as XR headsets, robots, and autonomous vehicles, to build a map of an unknown environment (mapping) in which they navigate while simultaneously tracking their current location and orientation (localization)~\cite{cadena2016past}. \autoref{fig:overview} illustrates different components in a SLAM pipeline and various methods used for each step. 
The state-of-the-art SLAM methods can be broadly classified into three categories: (i) traditional SLAM methods, (ii) end-to-end learning-based methods, and (iii) hybrid methods. 
These methods differ in how they extract and process information from the sensor data. For example, traditional methods may extract features from images or estimate motion based on pixel movement, while learning-based methods may use raw pixel data. 
Below, we discuss the tasks performed by a tracking pipeline and outline how different SLAM methods perform these tasks. 

\vspace{-0.15cm}
\subsubsection{\textbf{Visual SLAM Components and Pipelines.}} 
\label{sec:slam-components}
The SLAM pipelines track various types of objects (human wearing an XR headset, drone, robot, car)
that have idiosyncratic motion patterns (live-form, turbulent, fluid) in various environments (indoor/outdoor, urban/rural), as shown in~\autoref{fig:overview}. 
While additional combinations of these attributes exist, we focus on these examples as they cover most scenarios where visual SLAM tracking is employed.

\begin{enumerate}
    \item The first example is of humans wearing headsets in XR environments. Humans have a live-form motion as they can make complex movements in many directions as often as they want. Humans experience the XR environment in both indoor and outdoor settings.
    \item The second example is of a drone in indoor settings, where they keep track of inventory, monitor an industrial plant, or track people across a factory floor. They have turbulent locomotion forms as they must navigate complex environments and avoid collisions with other objects or in-operation machinery.
    \item The third example is of an autonomous car on the street or a highway, where they generally navigate the environment in a streamlined and fluid motion, i.e., their motion is often predictable.
\end{enumerate}

Given the three example scenarios, we next discuss the visual SLAM pipeline, outlining each task. We outline the different steps in the pipeline when using the traditional SLAM method, based on deep learning-based end-to-end methods or hybrid methods that leverage both approaches. The first step of a tracking pipeline involves data collection from the environment, which is shared across SLAM categories. 

\begin{enumerate}[label=\protect\circled{\arabic*}]
    \item A SLAM pipeline starts with data collected from sensors that observe the object and the environment. There are two types of sensors employed in SLAM applications. 
    One type consists of proprioceptive sensors that observe a phenomenon produced and perceived within a subject. 
    Examples of such sensors include the Inertial Measurement Unit (IMU). 
    The second type is exteroceptive sensors, which observe stimuli external to the subject.
    The example of such sensors includes cameras, LIDAR, and depth sensors~\cite{chong2015sensor}. 
\end{enumerate}

Sensors are part of device hardware, such as XR headsets, and they pass the sensed data to the SLAM pipeline. Next, we describe the several key stages of a traditional SLAM pipeline. Traditional methods also differ depending on whether they only track motion, such as visual odometry, or also keep a global environmental map. We use the latter method for subsequent discussion of traditional SLAM pipelines.

\begin{enumerate}[label=\protect\circled{\arabic*}, start=2, itemsep=0.08cm]
    \item In this first step of the traditional SLAM pipeline, data from all sensors is cleaned and processed to estimate the initial pose of the object (such as a human wearing a headset in XR) and create an initial map of the environment (such as a dimly lit room or the outdoor environment) in which the object navigates.

    \item The initial pose and map, alongside the preprocessed data, are passed to the tracking module, which is the most important stage of the SLAM pipeline. The first step of the tracking stage leverages visual odometry to detect and track features or landmarks in the environment. These features can be keypoints, descriptors, or other significant patterns, and are used to estimate motion (pose over time~\cite{ORBSLAM3_TRO, vins-mono-qin2017vins, DSM-Zubizarreta2020}). Feature-based methods identify and use keypoints or distinct image features for matching, tracking, and reconstruction. In parallel, keyframe-based methods select a subset of frames as keyframes to incrementally build the map and perform bundle adjustment more efficiently. This reduces computational cost compared to using all frames, making real-time operation feasible on mobile systems~\cite{keyframe-similar-badslam, klein2019smartphone}. The tracking stage is also responsible for state estimation correction, which is performed by the local bundle adjustment module by minimizing the difference between the observed position of points and their estimated position.

    \item Another key component is the loop closure detection module, which determines whether the object has returned to a place it has visited before. The loop detection module keeps track of previously visited places in a database. 
    Since the object may not reach a previously visited location from the same direction and angle, the loop detection module uses various rotations and translations when matching the visited locations in the database. 
    Recognizing a previously visited place enables map correction and ensures the trajectory's long-term consistency~\cite{latif2013robustloopclousure}.
    
    \item Finally, the global bundle adjustment module adjusts camera poses and 3D point positions to minimize the sum of re-projection errors across all camera views. This module performs the same task as the local bundle adjustment but at the global level. The final output of the SLAM pipeline is a map of the environment in which the object navigates and the object's pose within that map.
\end{enumerate}

The traditional SLAM approaches rely on detailed physical models of the world to exploit the sequential nature of the observations of the environment. 
While this approach works well in stable environments, it can struggle in dynamic environments with many moving objects. 
Recent work has shown that deep learning-based approaches can be helpful in various aspects of the SLAM pipeline. 
Recent advances in deep learning have enabled end-to-end learning methods that map the raw sensor data to the desired output (the pose and the map) using neural networks for processing images, detecting features, estimating depth, and performing other related tasks. 
These direct learning-based methods work on the raw pixel intensities of images rather than on extracted features, enabling a dense environment model~\cite{automated-driving-tracking-survey-cvpr-w-milz2018visual}.
They are especially valuable in texture-less regions or when capturing dense information.
An end-to-end learning-based approach removes the need to manually design and tune the different stages of the SLAM pipeline. 

\begin{enumerate}[label=\protect\circled{\arabic*}, start=6, itemsep=0.1cm]
    \item The first step of the learning-based tracking pipeline preprocesses the raw data from various sensors. The deep learning approaches excel at extracting information from the image, high-level features, and their representations, which are then fed as input into the deep neural network-based model. 

    \item The preprocessed data from each sensor is generally fed to sensor-specific feature encoders. These feature encoders generate high-level representations of the data. 
    In some end-to-end learning pipelines, the outputs of sensor-specific feature encoders may be fed to another encoder to generate a combined high-level representation. 
    The high-level representations are called feature vectors, representing the environment and state of the tracked object within the environment. 
    The feature vectors are then passed to the next step of the pipeline.

    \item The combined feature vectors encapsulating the high-level representations from the neural network are then fed into a pose regressor. 
    The pose regressor is also a neural network trained to predict the pose of the object based on the feature vectors. 
    The combined feature vectors can also generate a map of the environment as well. 
    
\end{enumerate}

The hybrid tracking methods replace one or more components of the traditional SLAM pipeline with learning-based methods. 
This could involve replacing components like feature extraction, data association, mapping, loop closure, and pose estimation. 
The choice of component to replace depends on the application's specific requirements, the available data, and the computational resources. 
Hybrid approaches aim to leverage the explicit performance guarantees of traditional SLAM methods, which are often based on well-understood mathematical principles, and learning-based methods, which can learn complex mappings from data and be robust to environmental changes~\cite{scona2018staticfusion}.

\subsection{Taxonomy of Challenges}
\label{sec:taxonomy}
\begin{figure}[t]
    \centering
    \includegraphics[width=1\linewidth]{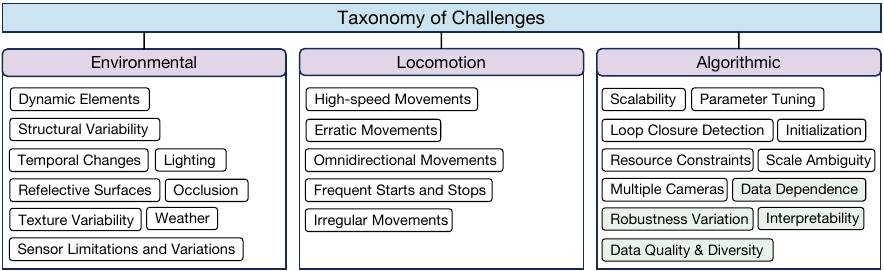} 
    \vspace{-0.7cm}
    \caption{\emph{\textbf{A taxonomy of challenges tracking algorithms face. The end-to-end learning challenges are highlighted in green.}}}
    \vspace{-0.2cm}
    \label{fig:taxonomy}
\end{figure}

\vspace{-0.1cm}
The unique human factors of XR, combined with the inherent complexities of SLAM, pose a broad set of intertwined challenges. To navigate this complex landscape, we categorize the challenges into environmental, locomotion, and algorithmic challenges. \autoref{fig:taxonomy} presents our taxonomy of challenges, whose different categories we next discuss.

\subsubsection{\textbf{Environmental Challenges}}
In this category, we include challenges that arise from the number and dynamics of elements in the environment, their perceptibility, and the perception capabilities of the sensors. The ability of the tracking systems heavily depends on the number of dynamic elements in the environment, the temporal characteristics of the environment, and its structural variability~\cite{zou2019structvio, structural-irrelagularite-li2021rgb, zhou2015structslam}. 
For example, tracking methods employed by autonomous vehicles perform better in highway settings with fewer elements that are also more predictable than in a chaotic city street environment. 
It is important to note that humans wearing XR headsets face environmental elements with unique characteristics and broader spatiotemporal dynamics than traditional tracking applications.
Humans in XR scenarios are potentially exposed to a broader range of environmental elements with different locomotion characteristics, such as pets, humans in the vicinity, inanimate objects, bicycles on the sidewalk, and vehicles on the road. 
On the other hand, a drone monitoring an industrial facility will likely not come across a teenager on a bicycle using her phone. 
Similarly, a car on a city street or a highway will not encounter a toddler fighting a pet.

The ability of an environment and its elements to be sensed is one of the most significant factors in determining the accuracy of the tracking systems. 
For example, lighting variations can destabilize a tracking system's consistency as light changes the image characteristics~\cite{taheri2021slam-survey}. 
Similarly, reflective surfaces and occlusions can impede tracking algorithms by presenting optical illusions and obstructing the field of view, which may appear as an object exiting the scene. 
The environment's textural characteristics are crucial in tracking features over time, and textural variations can result in poor tracking performance. 
Diurnal and seasonal changes to the environment can also impact accuracy, especially for learning-based methods, if they are not trained using a representative dataset.
Finally, an environment's ability to be sensed significantly varies with the weather as clouds, snow, and fog yield lighting changes, reflective surfaces, and occlusions~\cite{bresson2017simultaneous-trends-AVs}. 
Unfortunately, XR systems are exposed to environments that significantly vary in their characteristics, which determine how easy it is to sense objects. 
For example, humans effortlessly move from dimly lit bedrooms to streets with occlusions to hiking paths with ample sunlight or snow reflections.

Finally, the ability of the sensors to accurately sense the environmental elements adds another layer of complexity. Sensors' abilities to sense the environment vary based on their specifications, such as the frame rate they can support, their focal length, and sensor sizes. 
The sensor's capabilities can vary across application scenarios and impact the tracking accuracy. 
In addition, even if two sensors are the same based on all the metrics mentioned above, the sensor can have variations due to manufacturing defects or natural variations in the materials used to synthesize them~\cite{chong2015sensor, servieres2021visual-vislam, andreopoulos2011sensor}. Sensor limitations are an important component of environmental challenges. Hardware properties such as shutter type (rolling vs. global), resolution, field-of-view (FoV), frame rate, dynamic range, and sensor noise fundamentally constrain what tracking methods can achieve. For example, rolling shutter distortion during fast motion introduces geometric artifacts that degrade pose estimation unless explicitly modeled or corrected. Narrow FoV or low resolution reduces feature coverage and robustness, particularly in texture-poor scenes. A low frame rate exacerbates motion blur under fast locomotion, while a high frame rate increases computational demands. Finally, we note that sensor-related limitations—such as rolling shutter effects, limited field-of-view, and sensor noise, interact with environmental and locomotion factors to impact tracking accuracy; we analyze these effects in detail in Section~\ref{sec:sensor-eval}.

\vspace{-0.3cm}
\subsubsection{\textbf{Locomotion-based Challenges}} 
The movement of the objects being tracked significantly impacts the accuracy of a tracking approach. 
The movement of an object can be categorized based on its speed, predictability, directionality, and regularity.
Tracking objects at high speeds or with significant variations in speed significantly impacts the accuracy of the tracking systems. 
For example, tracking humans in XR is challenging, as humans adapt their speed depending on the situation and can stay still, walk, jog, or run. 
The erratic movements in all possible directions, i.e., omnidirectional movements, change the system state in an unpredictable and significant way, impacting the accuracy of tracking~\cite{pupilli2006real-erratic-locomotion, yu2021visual-locomotion}.
Finally, these are magnified in XR scenarios involving humans whose motion is characterized by frequent starts and stops~\cite{tian2015influence-locomotion}. 
Some of the tracked objects, such as drones and humans, can also have irregular movement patterns.

\subsubsection{\textbf{Algorithmic Challenges.}} 
The fields of tracking systems using SLAM- and odometry-based have made significant strides over the years. 
However, numerous aspects of tracking algorithms must consistently adapt to handle the new applications in increasingly complex environments. 
The accuracy of the tracking algorithm is dependent on the system initialization, which determines the accuracy of downstream operations~\cite{deep-vo-survey-jeong2021comparison}. 
Handling the environmental dynamics requires parameter tuning, which involves determining how frequently to adjust the parameters and the scale of changes to be made.
As various characteristics of the XR environment can change, such as the size of objects in the environment, the algorithms must be robust across varied scenarios~\cite{taheri2021slam-survey} that XR tracking systems may face.

Another key challenge is scaling tracking algorithms to expanded maps and complex environments, which requires significant computational resources.  
For example, Loop Closure Detection requires identifying previously visited locales, which scales as the environment expands~\cite{latif2013robustloopclousure}.
These issues are exacerbated in XR environments that pose the additional challenges of constrained resources~\cite{system-level-quant-analysis-karthik-2022}. 
Additionally, XR headsets often have multiple cameras, which introduces synchronization and calibration challenges~\cite{domain-adaptive-tracking2021PIC, saeedi2016multiple-robot-survey,depth-camera-issues-halmetschlager2018empirical}. 
These issues impact the quality and consistency of the available data to the algorithms and the data used to train the models. 
The performance significantly degrades if the input data distribution significantly shifts. 
Also, the robustness of these methods may vary across applications and environmental scenarios. Finally, unlike traditional SLAM methods, learning-based methods are black boxes and offer little to no interpretability. 
For reliable tracking, it is crucial to understand and solve multifaceted challenges for applications, including autonomous driving, robotics, and mixed reality.

\section{Visual SLAM: A Comprehensive Analysis in XR and IoT Ecosystems}
\label{sec:analysis}
In this section, we compare the performance of state-of-the-art visual tracking algorithms outlined in~\autoref{sec:sota-algos} across the applications and datasets described in~\autoref{sec:data-apps} using the metrics presented in~\autoref{sec:metrics}. In presenting our findings, we answer the following questions.
\begin{enumerate}[leftmargin=*, topsep=0.0cm]
    \item How do algorithms compare in their method-level, dataset-level, sequence-level, and sample-level performance?
    \item How does the choice of SLAM components, such as DL vs. traditional, impact end-to-end tracking performance? 
    \item How do SLAM, DL, and Hybrid methods compare in tackling environmental, locomotion, and algorithmic challenges?
    \item How do human factors prevalent in XR impact the performance of the tracking algorithms?
\end{enumerate}

\vspace{-0.3cm}
\subsection{Methodology} 
This section presents our methodology for quantitative comparative analysis of visual SLAM tracking methods. 

\begin{table}[t]
\centering
\scriptsize
\caption{\textbf{\emph{Summary of state-of-the-art tracking methods used in the evaluation.}}}
\vspace{-0.3cm}
\begin{tabular}{|l|l|l|l|l|l|l|}
\hline
\textbf{Algorithm} & \textbf{SLAM} & \textbf{VO} & \textbf{Deep Learning} & \textbf{Key frame-based} & \textbf{Feature-based} & \textbf{Direct} \\ \hline \hline
\textbf{ORBSLAM3-stereo}  & \checkmark &   &   & \checkmark & \checkmark &   \\ \hline
\textbf{ORBSLAM3-mono}  & \checkmark &   &   & \checkmark & \checkmark &   \\ \hline
\textbf{VINS-Fusion} & \checkmark & \checkmark &   & \checkmark & \checkmark  &   \\ \hline
\textbf{DSM}         & \checkmark &   &   &   &   & \checkmark \\ \hline \hline
\textbf{DROIDSLAM}   & \checkmark &   & \checkmark & \checkmark &   & \checkmark \\ \hline
\textbf{SfmLearner}  & \checkmark & \checkmark & \checkmark &   &   & \checkmark \\ \hline
\textbf{KP3D}        &   & \checkmark & \checkmark & \checkmark & \checkmark &   \\ \hline
\textbf{Tartanvo}    &   & \checkmark & \checkmark & \checkmark & \checkmark &   \\ \hline
\textbf{DFVO}        &   & \checkmark & \checkmark & \checkmark & \checkmark &   \\ \hline \hline
\textbf{Deepvo}      &   & \checkmark & \checkmark &   &   &   \\ \hline \hline
\end{tabular}

\vspace{-0.7cm}
\label{tab:methods}
\end{table}

\vspace{-0.2cm}
\subsubsection{\textbf{State-of-the-Art Tracking Algorithms}} 
\label{sec:sota-algos}
We select a representative set of state-of-the-art tracking algorithms that utilize various SLAM components and pipeline approaches that we described in~\autoref{sec:tracking-approaches} to analyze visual SLAM performance comprehensively. Table~\ref{tab:methods} lists the tracking methods we choose and the computational techniques they use.

In traditional SLAM methods, we choose ORB-SLAM3~\cite{ORBSLAM3_TRO} that leverages Oriented FAST and Rotated BRIEF (ORB)~\cite{rublee2011orb} algorithm for feature extraction, tracking, and mapping. We use both monocular and stereo versions of ORB-SLAM3. 
The other two SLAM methods are VINS-Fusion~\cite{vins-fusion-qin2018online}, and Direct Sparse Mapping (DSM)~\cite{DSM-Zubizarreta2020} that use visual odometry (VO) and direct methods, respectively. VINS-Fusion employs unique feature processing to integrate visual information with inertial data. 
DSM does not rely on feature extraction but directly operates on pixel values.

We picked several hybrid methods that combine learning components with traditional methods. DROID-SLAM~\cite{teed2021droid} utilizes deep neural networks for depth prediction, camera pose estimation, and loop closure. 
It uniquely uses both keyframe detection and direct methods.
SfMLearner~\cite{sfmlearner-zhou2017unsupervised} combines SLAM, VO, and deep learning to directly learn structures from motion (SfM)~\cite{sfm-dynamic-survey-saputra2018visual} without needing explicit feature extraction. 
\kpd~\cite{tang2020kp3d} emphasizes keyframe and feature-based methods, leveraging self-supervised deep learning to improve VO task performance. 
TartanVO~\cite{tartanvo2020corl} also uses deep learning for VO and incorporates keyframe detection and feature-based techniques. 
DFVO~\cite{zhan2019dfvo} combines deep learning and feature-based VO to merge neural networks with traditional feature extraction and matching. 

In our analysis, DeepVO~\cite{wang2017deepvo} is the only pure deep learning-based end-to-end VO system, and it does not rely on traditional keyframe detection or explicit feature extraction.

\textbf{Evaluation Pipeline Note.} To ensure consistency across all methods, we adopt a unified evaluation pipeline: all trajectories are processed through the same post-alignment and error computation steps (ATE and RPE). We use the publicly available open-source versions of each SLAM/VO method, with default configurations unless otherwise specified. Importantly, we do not apply method-specific tuning, multiple reruns, or per-sequence parameter optimization. As a result, our reported ATE/RPE may differ from those in the original papers, where such tuning or manual loop closure interventions were used to achieve best-case performance. This design choice ensures a fair and systematic comparison across methods, even if it exposes sensitivity to failure modes.

\subsubsection{\textbf{Datasets and Application Scenarios}} 
\label{sec:data-apps}
We select three datasets that map to the three sample combinations of the environment, object, and motion, shown in~\autoref{fig:overview}. 
While there can be other possible combinations, these three use cases represent the most common tracking applications and the most widely used datasets for these applications. We summarize the key sensor specifications for these datasets in \autoref{tab:sensor_specs}.

\vspace{-0.3cm}
\begin{table}[h!]
\centering
\scriptsize
\caption{\emph{\textbf{Sensor specifications for evaluated datasets.}}}
\vspace{-0.3cm}
\begin{tabular}{|l|l|l|l|l|l|}
\hline
\textbf{Dataset} & \textbf{Camera Type} & \textbf{Resolution} & \textbf{FPS} & \textbf{Shutter Type} & \textbf{FoV} \\
\hline
KITTI   & Stereo grayscale & 1392 $\times$ 512 & 10  & Global & $\sim$81$^\circ$ \\
\hline
EuRoC   & Stereo monochrome & 752 $\times$ 480  & 20  & Global & $\sim$60$^\circ$ \\
\hline
HoloSet & 4$\times$ grayscale + RGB + Depth & Gray: 640 $\times$ 480 & 30 & Rolling & Gray: $\sim$86$^\circ$ \\
        &                            & RGB: 1920 $\times$ 1080 & 30 & Rolling & RGB: $\sim$81$^\circ$ \\
        &                            & Depth: 512 $\times$ 512 & Variable (5-45) & Rolling & Depth: $\sim$75$^\circ$ \\
\hline
\end{tabular}
\label{tab:sensor_specs}
\end{table}

\begin{enumerate}[leftmargin=*]
    \item \textbf{KITTI~\cite{kitti-geiger2013vision}} is a benchmarking dataset that covers urban \emph{outdoor environments} captured from a car with \emph{fluid fast speed motions}. The dataset includes 22 color and grayscale stereo sequences with IMU and depth point clouds, of which 11 sequences are provided with ground truth using a Velodyne laser scanner and a high-precision GPS.

    \item \textbf{EuRoC~\cite{burri2016euroc} Micro Aerial Vehicle (MAV)} covers \emph{indoor environments} across two sites of varying structural, textural, and motion (fast/slow) difficulties. It includes machine halls and rooms with flat and structured environments. It is captured using AscTec Firefly hex-rotor MAV equipped with two stereo cameras and IMU with ground truth, making it ideal for testing applications with \emph{turbulent motion}.

    \noindent 
    \item \textbf{HoloSet~\cite{chandio2022holoset}} is collected using a Mixed Reality headset - Microsoft Hololens2~\cite{hololens2} worn by a human user in both \emph{indoor and outdoor environments}. It provides data from four grayscale, two depth, and one color cameras with IMU and ground truth.  
    It captures \emph{live-form motion} with substantial human factors in data collection reflecting typical human movements, interactions, and occlusions.     
\end{enumerate}

While there can be other possible combinations, these three use cases represent the most common tracking applications and the most widely used datasets for these applications. We acknowledge that other datasets such as ADVIO~\cite{advio-dataset}, OxIOD~\cite{oxiod-dataset}, YTU~\cite{ytu-dataset}, TUM VI~\cite{tum-vi}, ZJU-SenseTime~\cite{jinyu2019survey}, Oxford RobotCar~\cite{maddern20171}, UMA-VI~\cite{zuniga2020vi}, PALVIO~\cite{wang2022lf}, ICL-NUIM~\cite{handa2014benchmark}, and  M{\'a}laga~\cite{blanco2014malaga}. 
Other includes datasets that focus on human gait (MAREA~\cite{khandelwal2017evaluation}, OU-ISIR~\cite{uddin2018isir}), occupancy (LARA~\cite{niemann2020lara}), and activity recognition (USC-HAD~\cite{zhang2012usc}, CMU-MMAC~\cite{de2009guide}, Opportunity~\cite{chavarriaga2013opportunity}) offer complementary challenges, including richer motion types, XR-relevant trajectories, eye tracking and sensor configurations such as visual-inertial streams. However, we restrict our study to datasets that provide visual-only streams to ensure consistency across the selected monocular, stereo, and learning-based SLAM/VO methods. Evaluating cross-modal SLAM methods, such as visual-inertial pipelines, is a promising direction for future work, but beyond the current paper's scope.

\subsubsection{\textbf{Performance Metrics for Evaluation}} 
\label{sec:metrics}

We use the quality of the trajectory estimated by the tracking method to evaluate and compare performance. While an accurate trajectory does not imply a good map or error-free operation, it is the most commonly used metric to measure the accuracy of SLAM methods. We use two widely used trajectory metrics for our quantitative analysis:

\begin{itemize}[leftmargin=*]
    \item \textbf{Absolute Trajectory Error (ATE)} measures the difference between the translation parts of two trajectories after aligning them into a common reference frame\cite{ATE_RPE_salas2015trajectory}. Mathematically, if \( T_{gt} \) is the ground truth trajectory and \( T_{est} \) is the estimated trajectory, then:
    \begin{equation}
    ATE = \sqrt{\frac{1}{N} \sum_{i=1}^{N} || T_{gt}(i) - T_{est}(i) ||^2}
    \end{equation}

    \item \textbf{Relative Pose Error (RPE)} measures the difference between relative transformations at time instances \( i \) and \( i + k \) for different values of \( k \). If \( \Delta T_{gt}(i, i+k) \) and \( \Delta T_{est}(i, i+k) \) are the relative transformations for ground truth and estimated trajectories, respectively, then:
    \begin{equation}
    RPE = \sqrt{\frac{1}{N-k} \sum_{i=1}^{N-k} || \log(\Delta T_{gt}(i, i+k)^{-1} \Delta T_{est}(i, i+k)) ||_{\mathcal{F}}^2}
    \end{equation}
    where \( \log \) denotes the matrix logarithm and \( || . ||_{\mathcal{F}} \) is the Frobenius norm~\cite{bottcher2008frobenius-norm}. 
    This metric is independent of the reference frame, but when the scale of the map is unknown (e.g., monocular mapping), scale alignment needs to be performed before comparing trajectories using RPE~\cite{ATE_RPE_salas2015trajectory}.

    \item \textbf{Summary Metrics}. In addition, we define three summary metrics that allow us to compare the performance of algorithms across sequences and datasets. The 
    three metrics include (i) the number of sequences for which an algorithm yields the lowest error, (ii) the average error across sequences, and (iii) the coefficient of variation (CoV) in error across sequences, computed as the standard deviation over the mean.
\end{itemize}

\begin{table*}[t]
\centering
\caption{\textbf{\emph{ATE [m] and RPE [m] on EuRoC (Note: (--) ATE above 100m or RPE above 1, (xx) complete failure).}}}
\vspace{-0.35cm}
\resizebox{\textwidth}{!}{%
\begin{tabular}{|c|c|c|c|c|c|c|c|c|c|c|c|}
\toprule
\multirow{2}{*}{\textbf{Algorithm}} & \multicolumn{11}{c|}{\textbf{ATE , RPE}} \\ \cline{2-12} 
 & \multicolumn{1}{c|}{\textbf{MH01}} & \multicolumn{1}{c|}{\textbf{MH02}} & \multicolumn{1}{c|}{\textbf{MH03}} & \multicolumn{1}{c|}{\textbf{MH04}} & \multicolumn{1}{c|}{\textbf{MH05}} & \multicolumn{1}{c|}{\textbf{V101}} & \multicolumn{1}{c|}{\textbf{V102}} & \multicolumn{1}{c|}{\textbf{V103}} & \multicolumn{1}{c|}{\textbf{V201}} & \multicolumn{1}{c|}{\textbf{V202}} & \textbf{V203} \\ \midrule \hline
\textbf{ORB-SLAM3 (S)} & \multicolumn{1}{c|}{0.24 , 0.005} & \multicolumn{1}{c|}{3.11 , 0.002} & \multicolumn{1}{c|}{2.54 , 0.024} & \multicolumn{1}{c|}{2.53 , 0.014} & \multicolumn{1}{c|}{2.62 , 0.002} & \multicolumn{1}{c|}{0.49 , 0.003} & \multicolumn{1}{c|}{0.57 , 0.001} & \multicolumn{1}{c|}{0.64 , 0.009} & \multicolumn{1}{c|}{0.67 , 0.007} & \multicolumn{1}{c|}{0.84 , 0.004} & \textbf{0.22} , 0.001 \\ \hline
\textbf{ORB-SLAM3 (M)} & \multicolumn{1}{c|}{4.41 , 0.029} & \multicolumn{1}{c|}{4.99 , 0.034} & \multicolumn{1}{c|}{4.57 , 0.080} & \multicolumn{1}{c|}{2.03 , 0.010} & \multicolumn{1}{c|}{2.29 , 0.011} & \multicolumn{1}{c|}{1.01 , 0.011} & \multicolumn{1}{c|}{1.26 , 0.029} & \multicolumn{1}{c|}{1.29 , 0.030} & \multicolumn{1}{c|}{1.31 , 0.008} & \multicolumn{1}{c|}{\textbf{0.90} , 0.013} & 1.59 , 0.035 \\ \hline
\textbf{VINS-Fusion} & \multicolumn{1}{c|}{3.91 , 0.056} & \multicolumn{1}{c|}{3.97 , 0.060} & \multicolumn{1}{c|}{2.68 , 0.099} & \multicolumn{1}{c|}{2.57 , 0.042} & \multicolumn{1}{c|}{3.08 , 0.047} & \multicolumn{1}{c|}{1.11 , 0.022} & \multicolumn{1}{c|}{0.77 , 0.027} & \multicolumn{1}{c|}{\textbf{0.32} , 0.009} & \multicolumn{1}{c|}{1.26 , 0.017} & \multicolumn{1}{c|}{1.13 , 0.034} & 1.16 , 0.012 \\ \hline
\textbf{DSM} & \multicolumn{1}{c|}{6.24 , --} & \multicolumn{1}{c|}{\textbf{5.38} , --} & \multicolumn{1}{c|}{6.80 , --} & \multicolumn{1}{c|}{6.04 , --} & \multicolumn{1}{c|}{7.05 , --} & \multicolumn{1}{c|}{7.10 , --} & \multicolumn{1}{c|}{6.48 , --} & \multicolumn{1}{c|}{6.35 , --} & \multicolumn{1}{c|}{6.21 , --} & \multicolumn{1}{c|}{8.53 , --} & 9.06 , -- \\ \hline \hline
\textbf{DROID-SLAM} & \multicolumn{1}{c|}{2.94 , 0.073} & \multicolumn{1}{c|}{4.25 , 0.074} & \multicolumn{1}{c|}{1.86 , 0.145} & \multicolumn{1}{c|}{4.16 , 0.074} & \multicolumn{1}{c|}{3.36 , 0.075} & \multicolumn{1}{c|}{\textbf{1.01} , 0.024} & \multicolumn{1}{c|}{1.63 , 0.041} & \multicolumn{1}{c|}{12.16 , 0.020} & \multicolumn{1}{c|}{1.41 , 0.031} & \multicolumn{1}{c|}{1.20 , 0.054} & 1.43 , 0.092 \\ \hline
\textbf{SfMLearner} & \multicolumn{1}{c|}{4.49 , 0.673} & \multicolumn{1}{c|}{4.76 , 0.750} & \multicolumn{1}{c|}{3.45 , --} & \multicolumn{1}{c|}{5.50 , 0.459} & \multicolumn{1}{c|}{5.46 , 0.550} & \multicolumn{1}{c|}{\textbf{1.65} , 0.524} & \multicolumn{1}{c|}{1.84 , 0.675} & \multicolumn{1}{c|}{1.90 , 0.882} & \multicolumn{1}{c|}{2.04 , --} & \multicolumn{1}{c|}{2.11 , 0.869} & 2.11 , 0.885 \\ \hline
\textbf{KP3D} & \multicolumn{1}{c|}{0.22 , 0.064} & \multicolumn{1}{c|}{\textbf{0.21} , 0.084} & \multicolumn{1}{c|}{0.20 , 0.158} & \multicolumn{1}{c|}{0.18 , 0.053} & \multicolumn{1}{c|}{0.25 , 0.025} & \multicolumn{1}{c|}{0.37 , 0.022} & \multicolumn{1}{c|}{0.54 , 0.021} & \multicolumn{1}{c|}{0.71 , 0.021} & \multicolumn{1}{c|}{0.48 , 0.049} & \multicolumn{1}{c|}{1.07 , 0.070} & 0.39 , 0.033 \\ \hline
\textbf{TartanVO} & \multicolumn{1}{c|}{1.67 , 0.003} & \multicolumn{1}{c|}{1.62 , 0.003} & \multicolumn{1}{c|}{2.97 , 0.009} & \multicolumn{1}{c|}{2.37 , 0.005} & \multicolumn{1}{c|}{2.15 , 0.004} & \multicolumn{1}{c|}{0.54 , 0.001} & \multicolumn{1}{c|}{0.69 , 0.006} & \multicolumn{1}{c|}{\textbf{0.53} , 0.003} & \multicolumn{1}{c|}{1.10 , 0.004} & \multicolumn{1}{c|}{1.37 , 0.009} & 1.16 , 0.006 \\ \hline
\textbf{DFVO} & \multicolumn{1}{c|}{--, --} & \multicolumn{1}{c|}{--, --} & \multicolumn{1}{c|}{--, --} & \multicolumn{1}{c|}{39.1 , 0.160} & \multicolumn{1}{c|}{60.9 , 0.054} & \multicolumn{1}{c|}{\textbf{6.35} , 0.006} & \multicolumn{1}{c|}{--, --} & \multicolumn{1}{c|}{--, --} & \multicolumn{1}{c|}{24.7 , 0.021} & \multicolumn{1}{c|}{--, --} & --, -- \\ \hline \hline
\textbf{DeepVO} & \multicolumn{1}{c|}{1.67 , 0.091} & \multicolumn{1}{c|}{1.59 , 0.109} & \multicolumn{1}{c|}{1.65 , 0.176} & \multicolumn{1}{c|}{\textbf{1.56} , 0.122} & \multicolumn{1}{c|}{1.48 , 0.116} & \multicolumn{1}{c|}{1.84 , 0.028} & \multicolumn{1}{c|}{2.11 , 0.021} & \multicolumn{1}{c|}{1.71 , 0.039} & \multicolumn{1}{c|}{2.02 , 0.029} & \multicolumn{1}{c|}{2.18 , 0.098} & 1.85 , 0.075 \\ \hline \hline
\end{tabular}%
\label{tab:euroc}
}
\vspace{-0.4cm}
\end{table*}
\setlength{\tabcolsep}{4pt}
\begin{table*}[t]
\centering
\caption{\textbf{\emph{ATE [m] and RPE [m] on KITTI. Note: DSM failed on all sequences.}}}
\vspace{-0.35cm}
\resizebox{\textwidth}{!}
{%
\begin{tabular}{|c|c|c|c|c|c|c|c|c|c|c|c|}
\toprule
\multirow{2}{*}{\textbf{Algorithm}} & \multicolumn{11}{c|}{\textbf{ATE , RPE}} \\ \cline{2-12} 
 & \multicolumn{1}{r|}{\textbf{00}} & \multicolumn{1}{r|}{\textbf{01}} & \multicolumn{1}{r|}{\textbf{02}} & \multicolumn{1}{r|}{\textbf{03}} & \multicolumn{1}{r|}{\textbf{04}} & \multicolumn{1}{r|}{\textbf{05}} & \multicolumn{1}{r|}{\textbf{06}} & \multicolumn{1}{r|}{\textbf{07}} & \multicolumn{1}{r|}{\textbf{08}} & \multicolumn{1}{r|}{\textbf{09}} & \multicolumn{1}{r|}{\textbf{10}} \\ \midrule \hline
\textbf{ORB-SLAM3 (S)} & \multicolumn{1}{c|}{4.80 , 0.022} & \multicolumn{1}{c|}{11.2 , 0.061} & \multicolumn{1}{c|}{9.59 , 0.039} & \multicolumn{1}{c|}{4.59 , 0.016} & \multicolumn{1}{c|}{\textbf{3.30 }, 0.016} & \multicolumn{1}{c|}{4.72 , 0.013} & \multicolumn{1}{c|}{4.60 , 0.019} & \multicolumn{1}{c|}{xx , xx} & \multicolumn{1}{c|}{7.27 , 0.031} & \multicolumn{1}{c|}{6.80 , 0.028} & 6.43 , 0.021 \\ \hline
\textbf{ORB-SLAM3 (M)} & \multicolumn{1}{c|}{15.6 , 0.151} & \multicolumn{1}{c|}{10.1 , 0.073} & \multicolumn{1}{c|}{17.5 , 0.101} & \multicolumn{1}{c|}{8.94 , 0.078} & \multicolumn{1}{c|}{\textbf{2.72 }, 0.090} & \multicolumn{1}{c|}{5.94 , 0.080} & \multicolumn{1}{c|}{16.8 , 0.247} & \multicolumn{1}{c|}{11.6 , 0.143} & \multicolumn{1}{c|}{5.22 , 0.077} & \multicolumn{1}{c|}{13.1 , 0.130} & 15.8 , 0.136 \\ \hline
\textbf{VINS-Fusion} & \multicolumn{1}{c|}{-- , --} & \multicolumn{1}{c|}{-- , --} & \multicolumn{1}{c|}{-- , --} & \multicolumn{1}{c|}{\textbf{14.4} , 0.139} & \multicolumn{1}{c|}{32.6 , 0.454} & \multicolumn{1}{c|}{73.5 , 0.400} & \multicolumn{1}{c|}{38.1 , 0.383} & \multicolumn{1}{c|}{26.6 , 0.212} & \multicolumn{1}{c|}{41.2 , 0.216} & \multicolumn{1}{c|}{53.1 , 0.406} & 39.6 , 0.258 \\ \hline

\textbf{DSM} & \multicolumn{1}{c|}{xx , xx} & \multicolumn{1}{c|}{xx , xx} & \multicolumn{1}{c|}{xx , xx} & \multicolumn{1}{c|}{xx , xx} & \multicolumn{1}{c|}{xx , xx} & \multicolumn{1}{c|}{xx , xx} & \multicolumn{1}{c|}{xx , xx} & \multicolumn{1}{c|}{xx , xx} & \multicolumn{1}{c|}{xx , xx} & \multicolumn{1}{c|}{xx , xx} & xx , xx \\ \hline \hline

\textbf{DROID-SLAM} & \multicolumn{1}{c|}{-- , --} & \multicolumn{1}{c|}{-- , --} & \multicolumn{1}{c|}{-- , --} & \multicolumn{1}{c|}{-- , --} & \multicolumn{1}{c|}{-- , --} & \multicolumn{1}{c|}{-- , --} & \multicolumn{1}{c|}{-- , --} & \multicolumn{1}{c|}{85.5 , 2.305} & \multicolumn{1}{c|}{-- , --} & \multicolumn{1}{c|}{-- , --} & -- , -- \\ \hline
\textbf{SfMLearner} & \multicolumn{1}{c|}{-- , --} & \multicolumn{1}{c|}{-- , --} & \multicolumn{1}{c|}{-- , --} & \multicolumn{1}{c|}{-- , --} & \multicolumn{1}{c|}{-- , --} & \multicolumn{1}{c|}{-- , --} & \multicolumn{1}{c|}{-- , --} & \multicolumn{1}{c|}{91.4 , --} & \multicolumn{1}{c|}{-- , --} & \multicolumn{1}{c|}{-- , --} & -- , -- \\ \hline

\textbf{KP3D} & \multicolumn{1}{c|}{15.2 , 0.074} & \multicolumn{1}{c|}{47.6 , 0.682} & \multicolumn{1}{c|}{34.2 , 0.134} & \multicolumn{1}{c|}{3.04 , 0.061} & \multicolumn{1}{c|}{\textbf{1.94} , 0.116} & \multicolumn{1}{c|}{15.6 , 0.080} & \multicolumn{1}{c|}{4.36 , 0.055} & \multicolumn{1}{c|}{3.87 , 0.043} & \multicolumn{1}{c|}{13.4 , 0.071} & \multicolumn{1}{c|}{9.07 , 0.075} & 10.2 , 0.069 \\ \hline

\textbf{TartanVO} & \multicolumn{1}{c|}{85.8 , 0.361} & \multicolumn{1}{c|}{48.2 , 0.329} & \multicolumn{1}{c|}{-- , --} & \multicolumn{1}{c|}{\textbf{2.69} , 0.046} & \multicolumn{1}{c|}{2.30 , 0.068} & \multicolumn{1}{c|}{54.9 , 0.206} & \multicolumn{1}{c|}{6.96 , 0.067} & \multicolumn{1}{c|}{14.7 , 0.114} & \multicolumn{1}{c|}{65.4 , 0.293} & \multicolumn{1}{c|}{34.9 , 0.156} & 13.1 , 0.098 \\ \hline

\textbf{DFVO} & \multicolumn{1}{c|}{-- , --} & \multicolumn{1}{c|}{-- , --} & \multicolumn{1}{c|}{-- , --} & \multicolumn{1}{c|}{\textbf{23.1}, 0.280} & \multicolumn{1}{c|}{93.9 , --} & \multicolumn{1}{c|}{78.3, 0.320} & \multicolumn{1}{c|}{-- , --} & \multicolumn{1}{c|}{40.4 , 0.190} & \multicolumn{1}{c|}{89.4 , 0.230} & \multicolumn{1}{c|}{-- , --} & -- , -- \\ \hline \hline
\textbf{DeepVO} & \multicolumn{1}{c|}{-- , --} & \multicolumn{1}{c|}{-- , --} & \multicolumn{1}{c|}{-- , --} & \multicolumn{1}{c|}{-- , --} & \multicolumn{1}{c|}{-- , --} & \multicolumn{1}{c|}{-- , --} & \multicolumn{1}{c|}{-- , --} & \multicolumn{1}{c|}{90.9 , --} & \multicolumn{1}{c|}{-- , --} & \multicolumn{1}{c|}{-- , --} & -- , -- \\ \hline \hline
\end{tabular}%
}
\vspace{-0.3cm}
\label{tab:kitti}
\end{table*}
\begin{table*}[t]
\centering
\footnotesize
\caption{\textbf{\emph{ATE [m] and RPE [m] on HoloSet. Note: ORBSLAM3 and DSM failed on all sequences.}}}
\vspace{-0.4cm}
\begin{tabular}{|c|cccccc|}
\toprule
\multirow{2}{*}{\textbf{Algorithm}} & \multicolumn{6}{c|}{\textbf{ATE , RPE}} \\ \cline{2-7} 
 & \multicolumn{1}{c|}{\textbf{campus-center-seq1}} & \multicolumn{1}{c|}{\textbf{campus-center-seq2}} & \multicolumn{1}{c|}{\textbf{suburbs-jog-seq1}} & \multicolumn{1}{c|}{\textbf{suburbs-jog-seq2}} & \multicolumn{1}{c|}{\textbf{suburbs-seq1}} & \textbf{suburbs-seq2} \\ \midrule \hline
\textbf{ORB-SLAM3 (S)} & \multicolumn{1}{c|}{xx , xx} & \multicolumn{1}{c|}{xx , xx} & \multicolumn{1}{c|}{xx , xx} & \multicolumn{1}{c|}{xx , xx} & \multicolumn{1}{c|}{xx , xx} & xx , xx \\ \hline
\textbf{ORB-SLAM3 (M)} & \multicolumn{1}{c|}{xx , xx} & \multicolumn{1}{c|}{xx , xx} & \multicolumn{1}{c|}{xx , xx} & \multicolumn{1}{c|}{xx , xx} & \multicolumn{1}{c|}{xx , xx} & xx , xx \\ \hline
\textbf{VINS-Fusion} & \multicolumn{1}{c|}{14.2 , 0.565} & \multicolumn{1}{c|}{13.2 , 0.002} & \multicolumn{1}{c|}{26.1 , 1.000} & \multicolumn{1}{c|}{\textbf{2.68} , 0.267} & \multicolumn{1}{c|}{-- , --} & 4.05 , 0.102 \\ \hline
\textbf{DSM} & \multicolumn{1}{c|}{xx , xx} & \multicolumn{1}{c|}{xx , xx} & \multicolumn{1}{c|}{xx , xx} & \multicolumn{1}{c|}{xx , xx} & \multicolumn{1}{c|}{xx , xx} & xx , xx \\ \hline \hline
\textbf{DROID-SLAM} & \multicolumn{1}{c|}{-- , --} & \multicolumn{1}{c|}{\textbf{15.9} , 0.524} & \multicolumn{1}{c|}{32.8 , 1.797} & \multicolumn{1}{c|}{37.3 , 6.384} & \multicolumn{1}{c|}{-- , --} & 72.9 , 0.787 \\ \hline

\textbf{SfMLearner} & \multicolumn{1}{c|}{24.7 , --} & \multicolumn{1}{c|}{\textbf{17.4} , --} & \multicolumn{1}{c|}{33.6 , --} & \multicolumn{1}{c|}{38.4 , --} & \multicolumn{1}{c|}{-- , --} & 79.3 , -- \\ \hline

\textbf{KP3D} & \multicolumn{1}{c|}{1.40 , 0.005} & \multicolumn{1}{c|}{1.36 , 0.007} & \multicolumn{1}{c|}{1.18 , 0.005} & \multicolumn{1}{c|}{\textbf{1.03} , 0.004} & \multicolumn{1}{c|}{10.3 , 0.008} & 1.91 , 0.002 \\ \hline
\textbf{TartanVO} & \multicolumn{1}{c|}{8.11 , 0.037} & \multicolumn{1}{c|}{15.3 , 0.072} & \multicolumn{1}{c|}{7.92 , 0.045} & \multicolumn{1}{c|}{\textbf{4.77} , 0.030} & \multicolumn{1}{c|}{21.5 , 0.091} & 45.1 , 0.084 \\ \hline
\textbf{DFVO} & \multicolumn{1}{c|}{-- , --} & \multicolumn{1}{c|}{-- , --} & \multicolumn{1}{c|}{-- , --} & \multicolumn{1}{c|}{95.2 , 1.645} & \multicolumn{1}{c|}{-- , --} & -- , -- \\ \hline \hline
\textbf{DeepVO} & \multicolumn{1}{c|}{25.0 , 0.577} & \multicolumn{1}{c|}{16.9 , 1.163} & \multicolumn{1}{c|}{33.0 , 0.716} & \multicolumn{1}{c|}{37.1 , 2.141} & \multicolumn{1}{c|}{-- , --} & 77.7 , 2.734 \\ \hline \hline
\end{tabular}%
\label{tab:holoset}
\vspace{-0.3cm}
\end{table*}
\begin{table*}[t]
\centering
\footnotesize
\caption{\textbf{\emph{Algorithm Performance Summary Across Datasets Computed Using Data in~\autoref{tab:euroc},~\autoref{tab:kitti}, and~\autoref{tab:holoset}. }}}
\vspace{-0.3cm}
\begin{tabular}{|c|wc{1cm}wc{1cm}wc{1cm}|wc{1cm}wc{1cm}wc{1cm}|wc{1cm}wc{1cm}wc{1cm}|}
\toprule
\multirow{2}{*}{\textbf{Algorithm}} & \multicolumn{3}{c|}{\textbf{EuRoC}} & \multicolumn{3}{c|}{\textbf{KITTI}} & \multicolumn{3}{c|}{\textbf{HoloSet}} \\ \cline{2-10} 
 & \multicolumn{1}{c|}{\textbf{Top}} & \multicolumn{1}{c|}{\textbf{Avg ATE (m)}} & \multicolumn{1}{c|}{\textbf{CoV}} & \multicolumn{1}{c|}{\textbf{Top}} & \multicolumn{1}{c|}{\textbf{Avg ATE (m)}} & \multicolumn{1}{c|}{\textbf{CoV}} & \multicolumn{1}{c|}{\textbf{Top}} & \multicolumn{1}{c|}{\textbf{Avg ATE (m)}} & \multicolumn{1}{c|}{\textbf{CoV}} \\ \midrule \hline
 
\textbf{ORB-SLAM3 (S)} & 3 & 1.03 & 0.85 & \textbf{5} & \textbf{6.23} & \textbf{0.38} & xx & xx & xx \\ \hline
\textbf{ORB-SLAM3 (M)} & 0 & 2.64 & 0.50 & 0 & 10.72 & 0.47 & xx & xx & xx \\ \hline
\textbf{VINS-Fusion} & 0 & 2.49 & 0.49 & 0 & 39.24 & 0.52 & 1 & 12.81 & 0.75 \\ \hline
\textbf{DSM} & 0 & 6.56 & 0.14 & xx & xx & xx & xx & xx & xx \\ \hline \hline
\textbf{DROID-SLAM} & 0 & 3.02 & 1.04 & 0 & -- & -- & 0 & 39.72 & 0.67 \\ \hline
\textbf{SfMLearner} & 0 & 3.27 & 0.47 & 0 & -- & -- & 0 & 38.68 & 0.62 \\ \hline
\textbf{KP3D} & \textbf{7} & \textbf{0.45} & 0.58 & 3 & 14.94 & 0.91 & \textbf{5} & \textbf{2.53} & 1.39 \\ \hline
\textbf{TartanVO} & 1 & 1.54 & 0.45 & 2 & 35.03 & 0.84 & 0 & 17.11 & 0.91 \\ \hline
\textbf{DFVO} & 0 & 32.26 & 0.57 & 1 & 61.70 & 0.54 & 0 & 95.20 & -- \\ \hline \hline
\textbf{DeepVO} & 0 & 1.77 & \textbf{0.13} & 0 & -- & -- & 0 & 37.94 & \textbf{0.61} \\ \hline \hline
\end{tabular}%
\label{tab:performance_summary}
\vspace{-0.4cm}
\end{table*}


\subsection{Head-to-Head Performance Comparison at All Levels}
\label{sec:}
We run all tracking methods on all datasets for a comparative quantitative analysis. All methods, except ORB-SLAM3 (stereo), use monocular images from the datasets. We post-process the results for all methods to calibrate the scales and recover poses.  We report ATE as a measure of global consistency and RPE as a measure of local trajectory accuracy in \autoref{tab:euroc} for EuRoC, \autoref{tab:kitti} for KITTI, and \autoref{tab:holoset} for HoloSet. We provide the summary results in~\autoref{tab:performance_summary}.
If an algorithm fails on a given sequence, we mark the failure as ``xx , xx''. If an algorithm runs but yields an ATE value above 100m or RPE of above 1, we mark that as ``-- , --''.

\vspace{0.4cm}
\noindent\emph{\textbf{1 -- Method-level Question: Are there clear winners in a head-to-head comparison across tracking methods?} }
To answer this question, we primarily analyze the EuRoC dataset results from~\autoref{tab:euroc} and ~\autoref{tab:performance_summary}, a dataset on which all methods run, and most have errors within reasonable bounds. \kpd performs the best of 7 out of the 11 sequences, and the next best method is \orbslamS, which beats \kpd on 3 of the remaining four sequences. 
\kpd and \orbslamS are the top two based on the average ATE values of 0.45m and 1.03m, respectively.
However, most methods have a very low ATE across sequences: 4 algorithms have errors less than 2m, and only \dvfo has an error above 7m.

\noindent \emph{\textbf{Key Takeaway.} No single tracking method consistently outperforms others, though most methods have low errors.}

\begin{figure*}[t]
     \centering
        \begin{subfigure}[b]{0.31\linewidth}
         \centering
         \includegraphics[width=\textwidth]{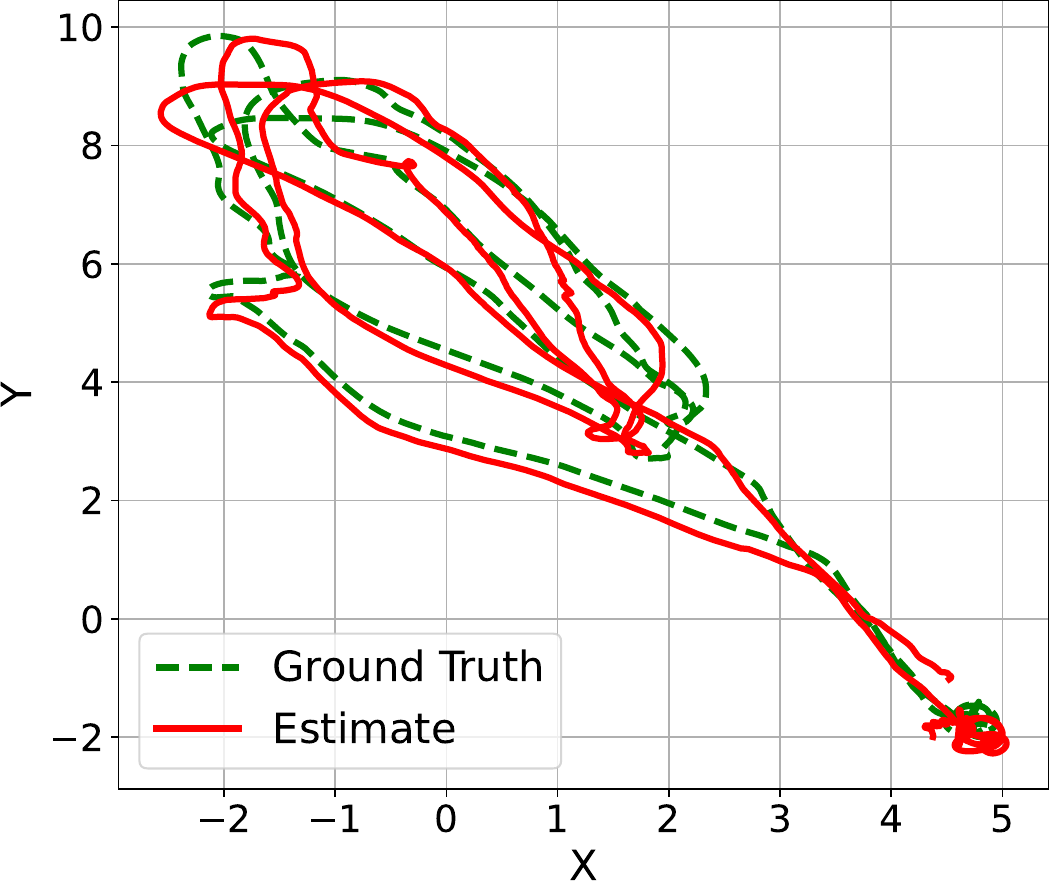}
         \vspace{-0.5cm}
         \caption{EuRoC MH02 sequence}
         \label{fig:euroc_mh02}
     \end{subfigure}
     \begin{subfigure}[b]{0.31\linewidth}
         \centering
         \includegraphics[width=\textwidth]{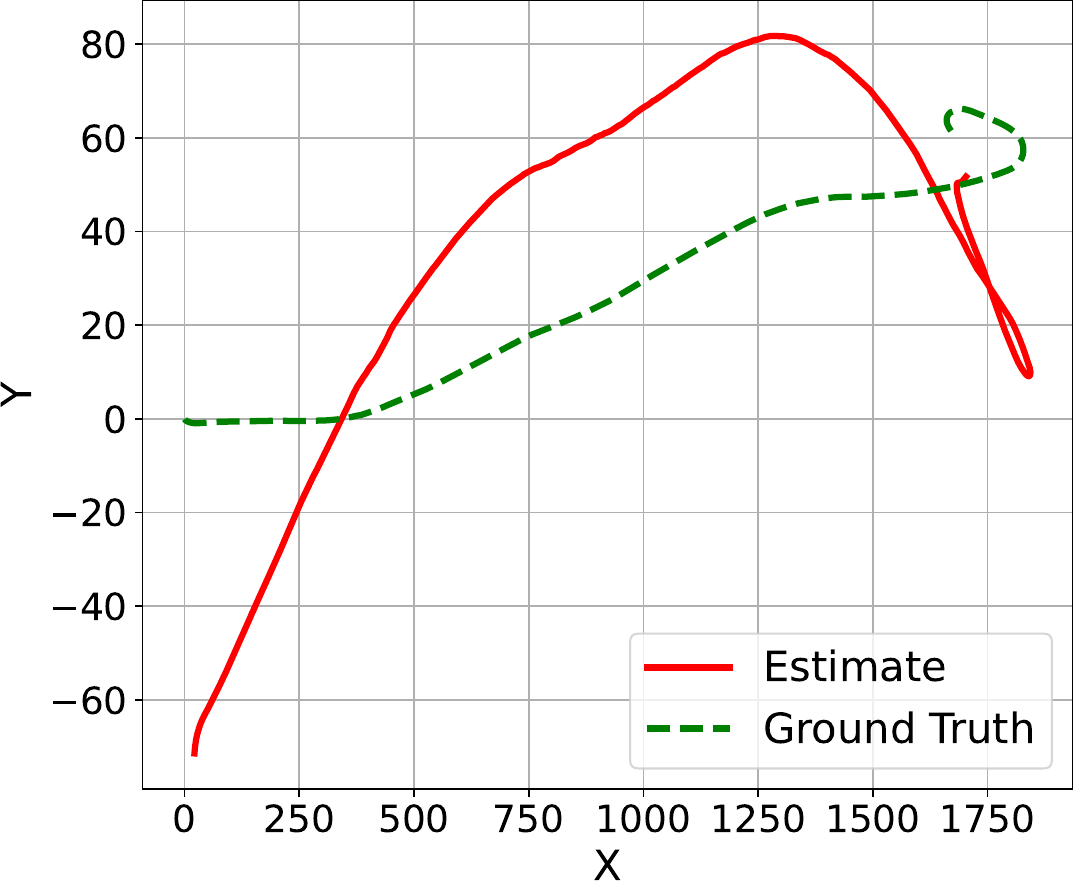}
         \vspace{-0.5cm}
         \caption{KITTI 01 sequence}
         \label{fig:kitti_01}
     \end{subfigure}
     \begin{subfigure}[b]{0.31\linewidth}
         \centering
         \includegraphics[width=\textwidth]{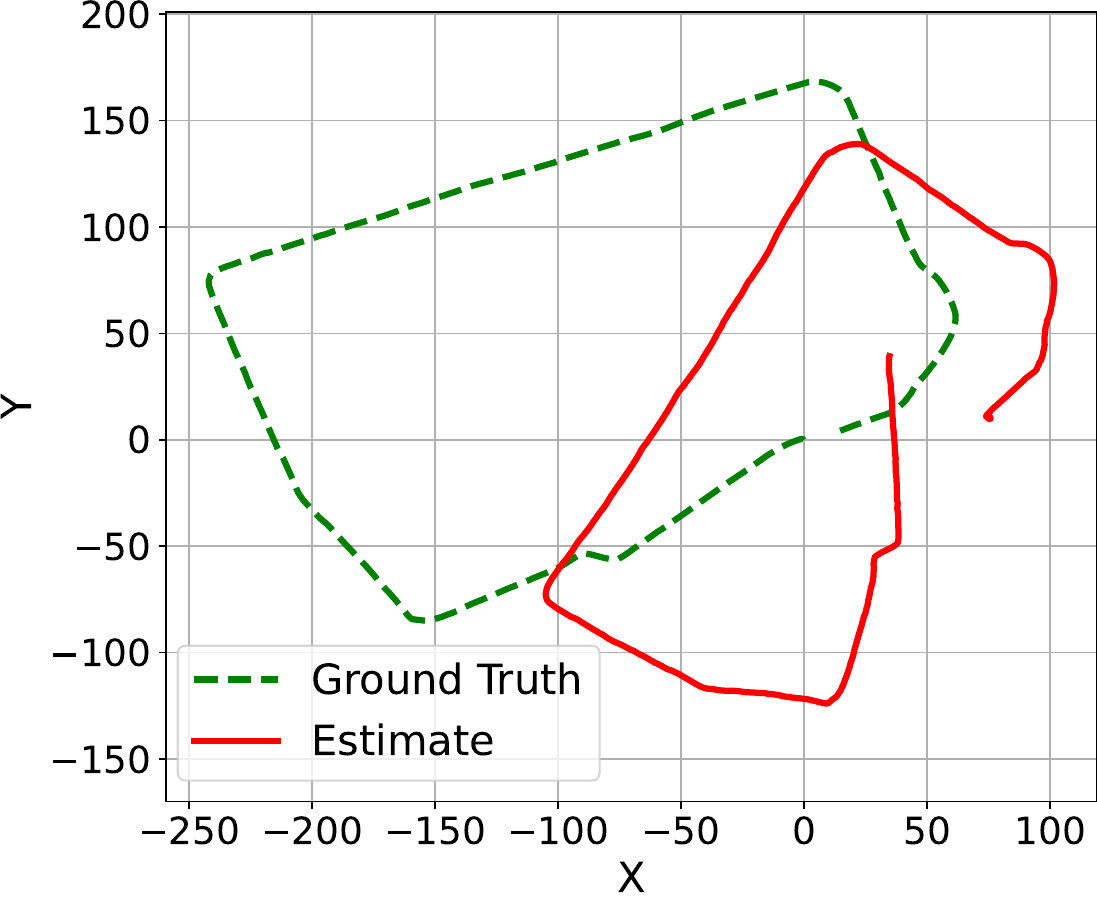}
         \vspace{-0.5cm}
         \caption{HoloSet suburb-seq1 sequence}
         \label{fig:holoset_sub_seq1}
     \end{subfigure}
     \vspace{-0.3cm}
    \caption{\textbf{\emph{Qualitative examples from representative datasets with an estimated trajectory from TartanVO with ground truth.}}}
    \label{fig:trajectories}
    \vspace{-0.35cm}
\end{figure*}

\vspace{0.1cm}
\noindent\emph{\textbf{2 -- Dataset-Level Question: Do tracking methods show robustness to variations in dataset characteristics?}}
To explore this, we analyzed the performance of algorithms across three distinct datasets, as shown in~\autoref{tab:performance_summary}. Interestingly, the best-performing algorithm was not the same across the datasets. 
For the KITTI dataset, \orbslamS emerged as the strongest, performing best on 5 of the 11 sequences, while \kpd followed closely with 3 sequences. However, when applied to the HoloSet dataset, \orbslamS failed completely, unable to generate trajectories for any sequence. 
SLAM pipelines such as \orbslamS rely on explicit loop closure and global optimization to maintain tracking consistency. However, human-controlled motion in XR environments (HoloSet) involves rapid, multi-directional head movements and limited scene revisitation, making loop closure ineffective and causing drift to accumulate rapidly. Moreover, HoloSet employs a rolling shutter camera, which introduces motion-induced artifacts during fast head rotations. These sensor characteristics, combined with fast angular motion, violate key assumptions of traditional SLAM systems, causing failure to initialize or maintain reliable tracking. SLAM methods tuned for global shutter and vehicle-like motion (EuRoC, KITTI) are fragile when exposed to a rolling shutter and rapid, 6-DoF head motion (HoloSet).
This pattern was reinforced by examining ATE values and the coefficient of variation (CoV), indicating that strong performance on one dataset does not necessarily translate to others, particularly when faced with different environmental conditions, motion dynamics, and object types. In part, this is due to differences in sensor characteristics (see Section~\ref{sec:sensor-eval}) and motion dynamics across datasets. For example, EuRoC and KITTI use global shutter cameras with relatively stable, smooth motion patterns (vehicle or drone-based). In contrast, HoloSet uses a rolling shutter camera and exhibits fast rotational motion from human head movements. SLAM methods tuned for global shutter and deep-learning methods trained on vehicle datasets often fail to generalize to these challenging XR motion conditions.

We also visually demonstrate the performance of a sample algorithm across sequences from different datasets. 
In~\autoref{fig:trajectories}, TartanVO’s performance is illustrated across three datasets, showing significant variation across sequences within each dataset. While it performs reasonably well in the EuRoC MH02 sequence with structured, textured surfaces, its accuracy drops notably in KITTI’s outdoor high-speed sequence, where low texture and fast motion introduce more complexity. The HoloSet suburban walk sequence further highlights this inconsistency, where TartanVO struggles with scale estimation and trajectory alignment due to the absence of loop closure detection. Pure deep-learning methods like TartanVO perform inference based on temporal image sequences but do not explicitly model global consistency or geometric structure. As a result, they can generalize well to familiar motion patterns but exhibit instability when exposed to motion regimes not seen in training data. In particular, XR head motion includes abrupt changes in rotation and translation that differ significantly from the vehicle-based or drone-based motions typically present in SLAM training datasets. This mismatch leads to degraded performance, particularly in sequences with fast or unpredictable human movement.

\noindent \emph{\textbf{Key Takeaway.} Tracking methods generally lack robustness when faced with diverse environmental, motion, and object characteristics across datasets, indicating variability in performance based on the dataset context.}

\vspace{0.1cm}
\noindent\emph{\textbf{3 -- Sequence-Level Question: Do algorithms show consistent performance across sequences within a single dataset?}} 
We analyzed the results across sequences to explore this, focusing on the data in~\autoref{tab:euroc} and~\autoref{tab:performance_summary}.
Among the tracking methods, \deepvo demonstrates the most consistent performance, with the lowest coefficient of variation (CoV) across sequences at 0.13. Its average ATE of 1.77m is relatively low, positioning it among the better-performing methods despite not being the top performer for any sequence. In contrast, top-performing methods like \kpd and \orbslamS exhibit higher CoV values of 0.58 and 0.85, indicating more variability in performance across sequences. DL methods are less sensitive to scene texture but cannot handle rapid viewpoint changes or scale ambiguity.
As a result, they produce consistent errors but often higher overall ATE, reflecting model bias toward learned motion patterns.

TartanVO’s performance is illustrated across three datasets, showing significant variation across sequences within each dataset. While it performs reasonably well in the EuRoC MH02 sequence with structured, textured surfaces, its accuracy drops notably in KITTI’s outdoor high-speed sequence, where low texture and fast motion introduce more complexity. The HoloSet suburban walk sequence further highlights this inconsistency, where TartanVO struggles with scale estimation and trajectory alignment due to the absence of loop closure detection.

\noindent \emph{\textbf{Key Takeaway.} Even though some tracking methods achieve low error for a dataset, this does not necessarily translate to consistent performance across all sequences.}

\begin{figure*}[t]
     \centering
     \begin{subfigure}[b]{0.16\linewidth}
         \centering
         \includegraphics[width=\textwidth]{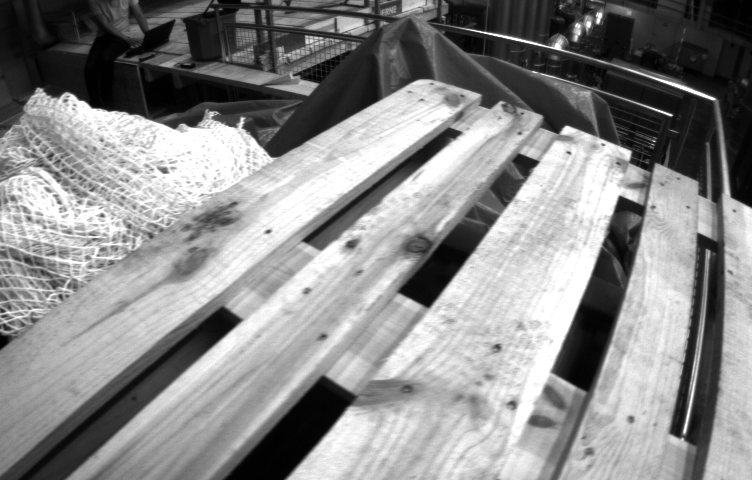}
         \label{fig:euroc_easy}
     \end{subfigure}
        \begin{subfigure}[b]{0.16\linewidth}
         \centering
         \includegraphics[width=\textwidth]{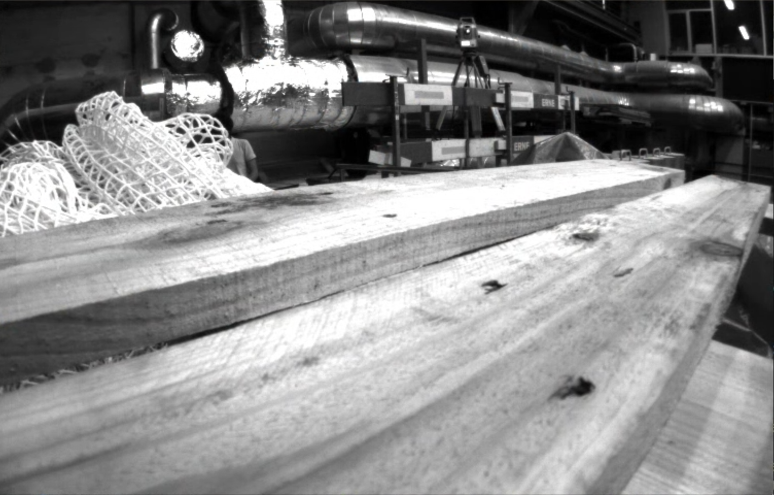}
         \label{fig:euroc_occlusion}
     \end{subfigure}
     \begin{subfigure}[b]{0.16\linewidth}
         \centering
         \includegraphics[width=\textwidth]{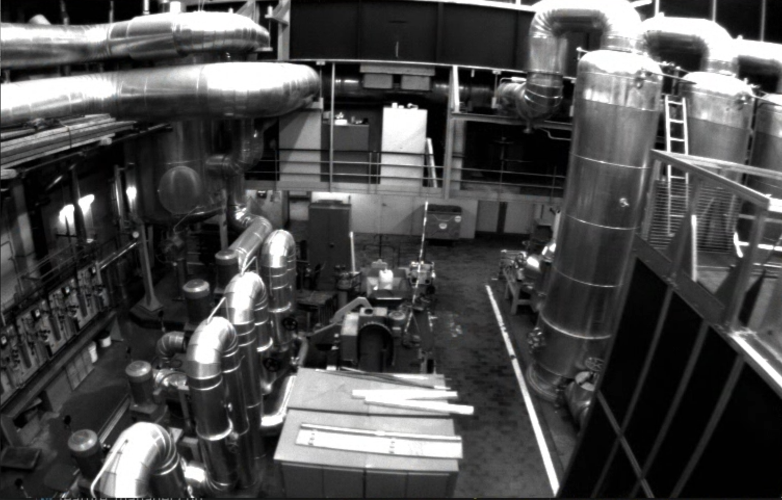}
         \label{fig:euroc_reflective}
     \end{subfigure}
     \begin{subfigure}[b]{0.16\linewidth}
         \centering
         \includegraphics[width=\textwidth]{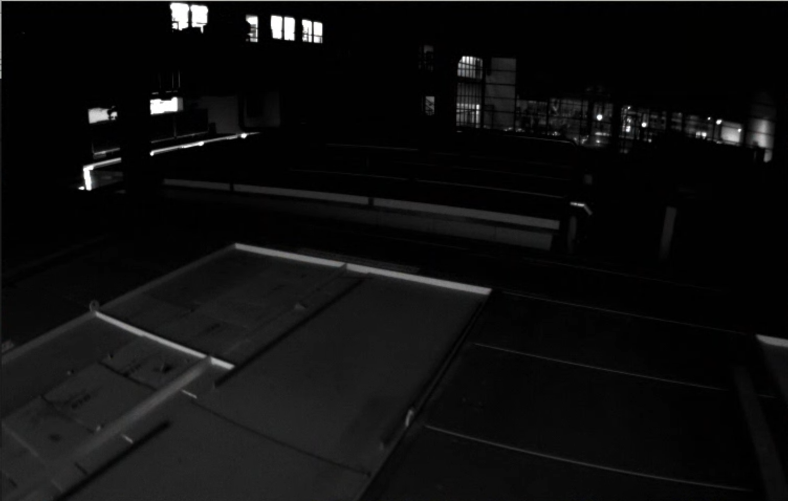}
         \label{fig:euroc_dark}
     \end{subfigure}
        \begin{subfigure}[b]{0.16\linewidth}
         \centering
         \includegraphics[width=\textwidth]{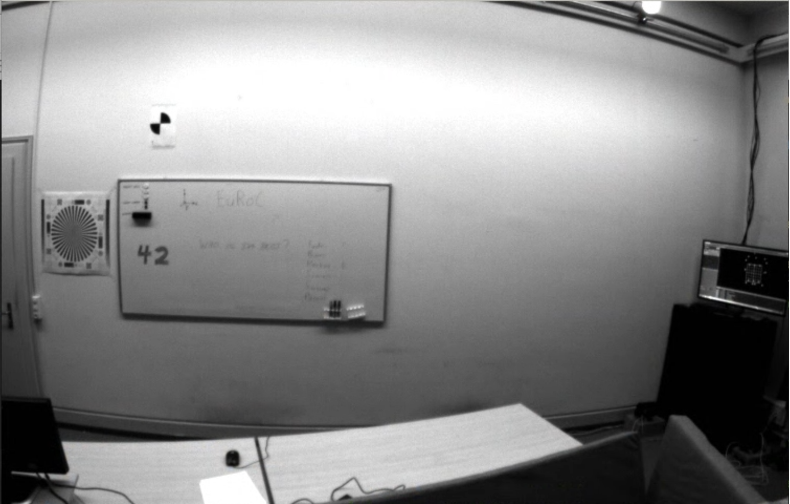}
         \label{fig:euroc_texture}
     \end{subfigure}
     \begin{subfigure}[b]{0.16\linewidth}
         \centering
         \includegraphics[width=\textwidth]{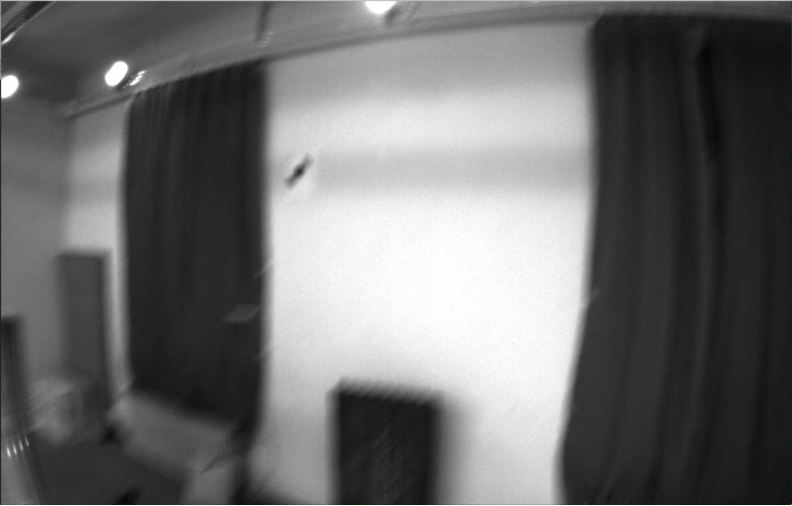}
         \label{fig:euroc_blurred}
     \end{subfigure}
          \begin{subfigure}[b]{0.16\linewidth}
         \centering
         \includegraphics[width=\textwidth]{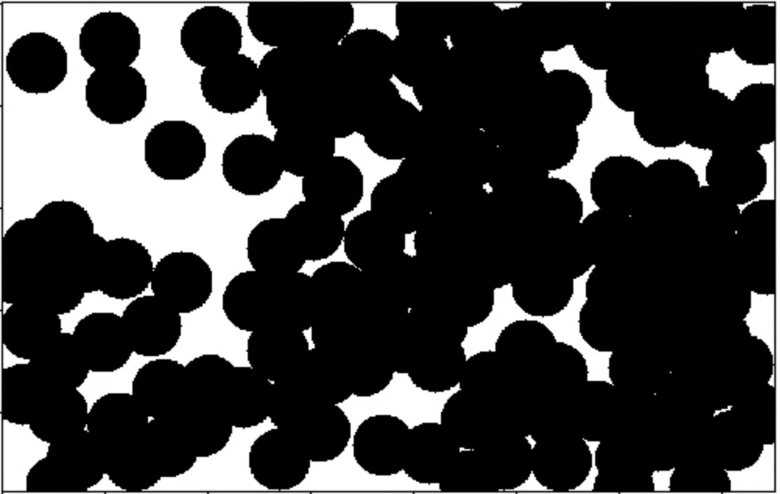}
         \caption{plain}
         \label{fig:euroc_easy_mask}
     \end{subfigure}
          \begin{subfigure}[b]{0.16\linewidth}
         \centering
         \includegraphics[width=\textwidth]{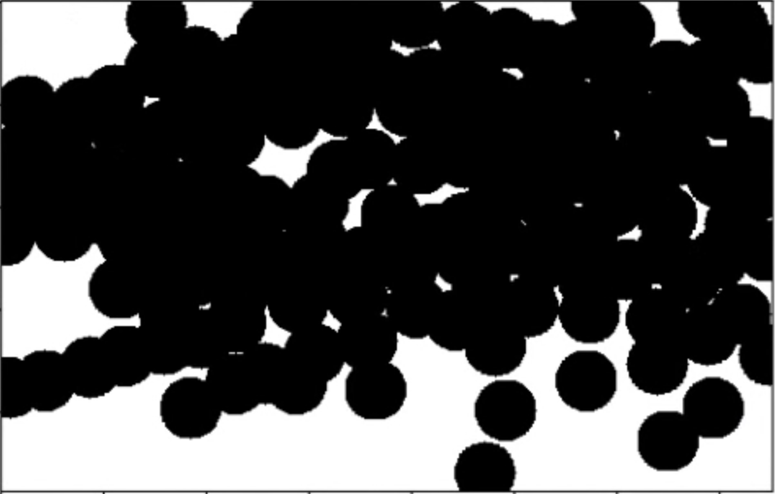}
         \caption{occlusion}
         \label{fig:euroc_occlusion_mask}
     \end{subfigure}
            \begin{subfigure}[b]{0.16\linewidth}
         \centering
         \includegraphics[width=\textwidth]{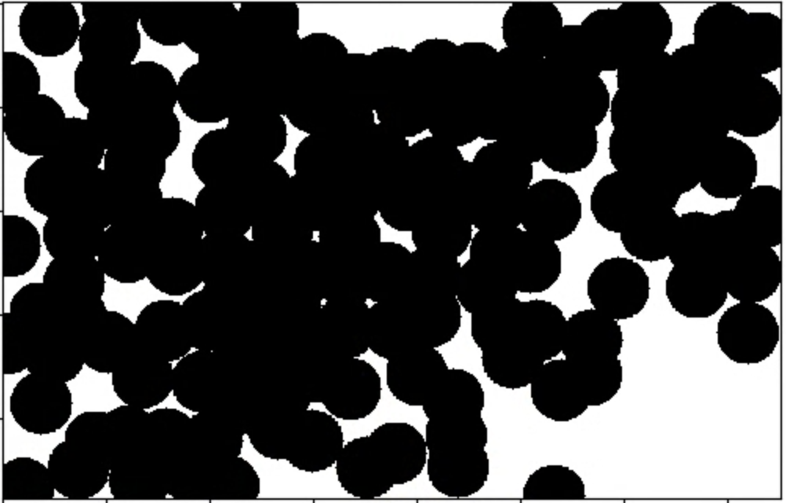}
         \caption{reflective surfaces}
         \label{fig:euroc_reflective_mask}
     \end{subfigure}
            \begin{subfigure}[b]{0.16\linewidth}
         \centering
         \includegraphics[width=\textwidth]{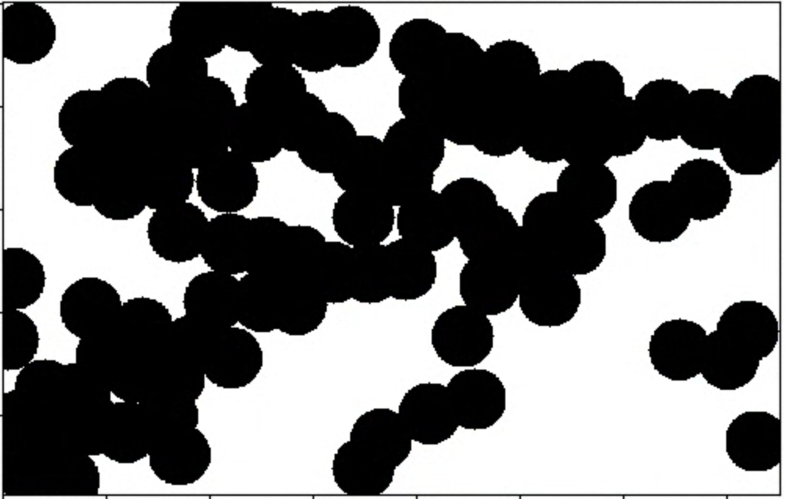}
         \caption{low light}
         \label{fig:euroc_dark_mask} 
     \end{subfigure}
    \begin{subfigure}[b]{0.16\linewidth}
         \centering
         \includegraphics[width=\textwidth]{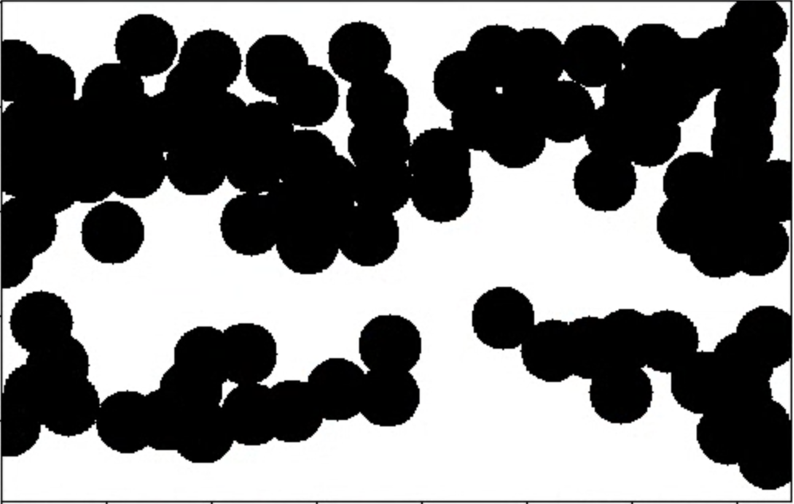}
         \caption{low texture}
         \label{fig:euroc_texture_mask}
     \end{subfigure}
        \begin{subfigure}[b]{0.16\linewidth}
         \centering
         \includegraphics[width=\textwidth]{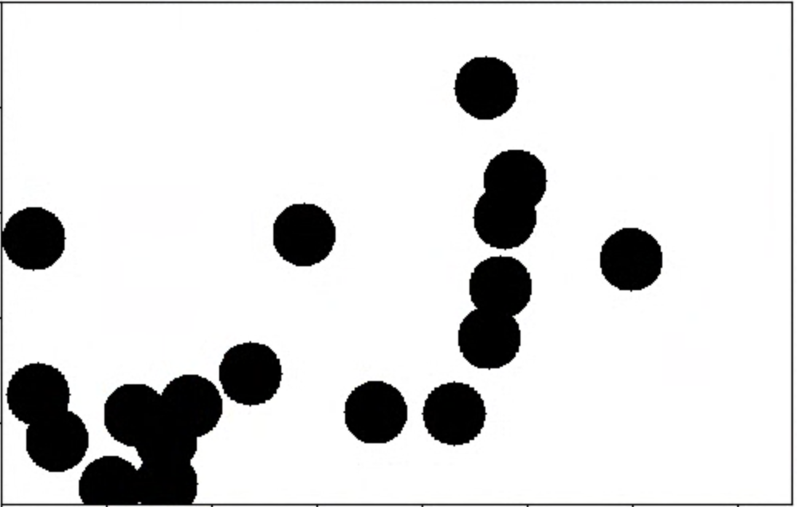}
         \caption{blurriness}
         \label{fig:euroc_blur_mask}
     \end{subfigure}
     \vspace{-0.25cm} 
     \caption{\textbf{\emph{Example samples from EuRoC~\cite{burri2016euroc} dataset, varied environmental conditions (top) and their corresponding feature ($\bullet$) mask  (bottom) generated with VINS-Fusion~\cite{vins-fusion-qin2018online} feature extractor.}}}
     \vspace{-0.35cm}
     \label{fig:sample-level-eval}
\end{figure*}

\vspace{0.1cm}
\noindent \emph{\textbf{4 -- Sample-level Question: Can the methods handle the variations in the samples within a sequence without sample-level fine-tuning?}}
Despite the good performance of some visual SLAM methods under ``ideal conditions'', there is no clear winner on \emph{method}, \emph{dataset}, and \emph{sequence} levels. Our evaluation has unveiled several \emph{algorithmic}, \emph{environmental}, and \emph{locomotion} challenges that SLAM methods face on different levels of the pipeline. To investigate further, we evaluate \emph{sample-level} attributes. We took six samples from EuRoC sequences (MH01, MH03, V203): one baseline vanilla sample and five samples with common visual features essential for data characterization in SLAM~\cite{dataset-characterization-ali2022we, saeedi2019characterizing}. These features include occlusion, lighting, reflective surface, image motion blur, and texture variability. We tracked the feature of each sample with the VINS-Fusion~\cite{vins-fusion-qin2018online} feature extractor as shown in Figure~\ref{fig:sample-level-eval}.  

We observe that the number of features ranges from 127 per frame to a mere 17 features per frame. While certain attributes might favor the algorithms, leading to reliable tracking (as shown in Figure~\ref{fig:euroc_easy_mask}), others can significantly strain the system, as seen in Figure~\ref {fig:euroc_blur_mask}. Any sudden movement (causing a blurry image), low lighting, low textured environment, or occlusions can hinder the methods' view of certain features. Whereas shiny reflective surfaces can create false features, compromising tracking accuracy. A method's ability to consistently recognize and adapt to these varying attributes without manual intervention or fine-tuning signifies its robustness.

\vspace{0.1cm}
\noindent{\emph{\textbf{Key Takeaway.} The disparity in results across the sequences, samples, and visual features indicates that achieving a truly generalized method for handling the broad spectrum of sample attributes remains an open question. A robust and domain-adaptive SLAM can be developed by understanding data at various levels, from the method down to the sample level.}}

\subsection{Impact of Choices in Visual SLAM Components and Pipelines} 

To analyze how the choice of SLAM components, such as deep learning (DL) versus traditional methods, impacts end-to-end tracking performance, we examine the results from~\autoref{tab:performance_summary}. 
The performance of traditional, hybrid, and deep learning approaches is evaluated based on their average ATE, the number of sequences for which they perform the best (Top), and the consistency of their performance as indicated by the CoV.

\vspace{0.1cm}
\noindent\textbf{Traditional Methods.} Traditional SLAM approaches, including \orbslamS and \orbslamM, generally perform well on specific datasets, particularly on KITTI and EuRoC. \orbslamS, for instance, is the top performer on five sequences in KITTI and three sequences in EuRoC. 
However, these methods struggle with robustness across datasets; for example, \orbslamS fails to work on HoloSet entirely. 
The CoV for these algorithms is relatively low on KITTI but higher for EuRoC, indicating more variability in performance across sequences within the EuRoC dataset.

\vspace{0.1cm}
\noindent\textbf{Hybrid Methods.}
Hybrid approaches, which combine traditional and deep learning components, tend to outperform traditional and DL-only methods on overall performance and consistency. 
\kpd, one of the top hybrid algorithms, has the lowest average ATE on EuRoC (0.45m) and performs well on KITTI, although with a higher CoV of 0.91. 
These hybrid methods dominate in average ATE and are the top performers in multiple sequences across different datasets, showing the most balanced performance. 
However, hybrid methods also struggle on HoloSet, with higher ATE and more variability compared to their performance on EuRoC and KITTI.

\vspace{0.1cm}
\noindent\textbf{Deep Learning Methods.}
The deep learning-based method, \deepvo, stands out for its consistency across sequences within a single dataset, demonstrated by the lowest CoV of 0.13 on EuRoC. However, its end-to-end performance in terms of average ATE is not as competitive, with values such as 1.77m on EuRoC and 37.94m on HoloSet, indicating that while it performs consistently, its overall error is higher than most hybrid and traditional approaches. Additionally, deep learning methods tend to struggle on datasets with more complex environmental or motion variations, as indicated by the large ATE on HoloSet. The low CoV of \deepvo reflects its ability to produce stable predictions across sequences, but its relatively higher ATE indicates a systematic bias or model drift. Since \deepvo lacks explicit modeling of scale or 3D structure, its learned motion estimation is less sensitive to texture or loop closure but prone to accumulating global errors. As a result, while the method appears consistent, its absolute accuracy suffers, particularly in longer or more dynamic sequences.

\noindent \emph{\textbf{Key Takeaway.} 
Hybrid methods exhibit the best overall performance, consistently achieving low ATE across multiple datasets and sequences. They outperform traditional and deep learning methods in terms of accuracy and robustness. While deep learning approaches offer more consistency within sequences, they generally produce higher errors, suggesting that traditional and hybrid methods are still more effective for end-to-end tracking in varied conditions.}

\vspace{-0.3cm}
\subsection{Performance in Tackling Environmental, Locomotion, and Algorithmic Challenges} 

To compare how SLAM, deep learning (DL), and hybrid methods tackle environmental, locomotion, and algorithmic challenges, we analyze their performance across the datasets (EuRoC, KITTI, and HoloSet) summarized in \autoref{tab:performance_summary}, in addition to the individual results for the datasets. 
These datasets capture a range of conditions, from indoor to outdoor environments, controlled settings to dynamic scenes, and varying frame rates and resolutions. Each method's ability to adapt to these challenges reveals insights into the strengths and weaknesses of different computational approaches.

\vspace{0.1cm}
\noindent\textbf{Traditional Methods.} 
SLAM methods like \orbslamS and \orbslamM perform well in controlled environments but struggle as environmental and motion complexity increases. 
On EuRoC’s easier sequences, such as MH01, \orbslamS achieves a low ATE of 0.24m, aided by EuRoC’s compact resolution and controlled environment, which facilitates smoother loop closures and reduces scalability challenges. 
However, as sequences become more turbulent (e.g., V202), \orbslamS’s ATE increases to 0.84m, reflecting the limitations of traditional methods in more complex indoor environments. 
In KITTI, while SLAM methods benefit from high-quality calibration and ground truth data, broader variability in lighting and environmental factors challenges their robustness. For instance, \orbslamS records an ATE of 4.80m in sequence 00, but as environmental complexity increases in sequence 02, the ATE jumps to 11.2m. This highlights SLAM’s struggle with outdoor settings where rapid changes in lighting and environment occur. On HoloSet, SLAM methods fail entirely, and they are unable to handle the unpredictable, multi-directional motion of the human-controlled XR headset. Both \orbslamS and \orbslamM are unable to produce valid trajectories, reflecting their difficulty in handling high frame rates and the uncontrolled dynamics of human motion. Furthermore, traditional SLAM methods are optimized for feature-rich environments; the sparse, texture-less backgrounds common in XR indoor scenes (white walls, floors) cause feature tracking to fail, leading to the inability to maintain global pose estimation. Furthermore, the higher frame rate and rolling shutter of the XR headset exacerbate feature tracking errors under fast head rotations, leading to divergence of SLAM pipelines.

\vspace{0.1cm}
\noindent\textbf{Hybrid Methods.}
Hybrid methods demonstrate superior adaptability across indoor and outdoor environments, effectively balancing precision and flexibility to handle varied conditions. For instance, \kpd achieves an ATE of 0.21m in EuRoC’s MH01 sequence and continues to perform well in the challenging V203 sequence with an ATE of 0.39m. The combination of feature extraction techniques and loop closure enables hybrid methods to handle both the compact resolution and high frame rate of EuRoC, maintaining low error rates even in turbulent drone motion. In KITTI, hybrid methods continue to outperform SLAM approaches, handling fluid car motion and complex environmental fluctuations more effectively. \kpd achieves an ATE of 14.9m in sequence 04, demonstrating resilience to the broader variability in outdoor conditions that challenge traditional methods. Their sophisticated feature extraction allows hybrid methods to manage the scene complexity and outdoor lighting fluctuations that typically degrade SLAM’s performance. In HoloSet, hybrid methods like \kpd excel in managing human-centered environments, with an ATE of 10.4m in campus-center-seq1. Although the ATE values increase due to the dynamic, uncontrolled environment and higher frame rates, hybrid methods remain the most robust option, able to handle conditions involving unpredictable human motion better. Their adaptability across domains -- from indoor corridors (EuRoC) to urban streets (KITTI) to suburban areas (HoloSet) -- positions hybrid methods as the most versatile and reliable for complex tracking tasks.

\vspace{0.1cm}
\noindent\textbf{Deep Learning Methods.} 
DL methods exhibit consistency across sequences but generally incur higher error rates, especially in environments with significant variability. 
While DL methods provide consistent performance, their ATE values are higher than hybrid methods. For instance, in MH01, \deepvo records an ATE of 1.67m -- consistent but less accurate compared to \kpd’s 0.21m. This indicates that DL methods are less precise in handling turbulent motion within controlled indoor environments. 
On KITTI, DL methods struggle more with outdoor complexity. For example, \deepvo cannot produce results for most sequences (denoted by “--” in the table), highlighting the difficulty of applying DL methods to rapidly changing outdoor environments, particularly when lighting conditions and scene complexity fluctuate.
On the HoloSet, DL methods perform better than SLAM but still lag behind hybrid methods. For example, \deepvo records an ATE of 33.0m in suburb-jog-seq1, significantly higher than hybrid methods like \kpd. While DL methods can handle unpredictable human motion, their error rates remain higher than hybrid methods, reflecting their limitations in highly dynamic, human-centered environments.
DL models lack explicit modeling of rotation and scene geometry, so fast rotations (head yaw/pitch) cause temporal inconsistencies that accumulate into trajectory errors.
Additionally, HoloLens' camera frame rate and latency differ from training data, impacting the temporal alignment of predictions.

\vspace{0.1cm}
\noindent \emph{\textbf{Key Takeaway.} 
Hybrid methods consistently outperform SLAM and DL approaches by balancing adaptability and accuracy across diverse environments, from controlled indoor settings to dynamic outdoor scenes. While SLAM methods excel in structured environments, they struggle in more complex scenarios, such as human-controlled movements in HoloSet, where they often fail. DL methods offer consistency but at the cost of higher errors, particularly in outdoor and unpredictable environments. Overall, hybrid methods demonstrate superior versatility, handling various challenges with lower error rates, making them the most robust option for complex tracking tasks.}

\vspace{-0.3cm}
\subsection{Effect of Human Factors in XR} 
The HoloSet dataset introduces significant challenges related to human factors, particularly the unpredictable and multi-directional motion associated with a human wearing an XR headset. This variability in movement, combined with changing environments from indoors to suburban outdoors, presents a unique set of challenges for tracking methods.

\vspace{0.1cm}
\noindent\textbf{Traditional Methods.} SLAM methods, such as \orbslamS and \orbslamM, struggle to handle the dynamic nature of human-controlled motion. These methods rely heavily on stable, predictable environments and predefined trajectories, which are difficult to maintain when the human’s head movements introduce abrupt and irregular changes in position and angle. Consequently, SLAM methods fail entirely in HoloSet, unable to produce any valid trajectories due to the unpredictable nature of human-controlled XR interactions.

\vspace{0.1cm}
\noindent\textbf{Hybrid Methods.}  Hybrid methods, like \kpd, demonstrate much better resilience to the human factors inherent in XR. While their ATE values increase compared to more controlled environments, they still manage reasonable performance in HoloSet. For instance, \kpd achieves an ATE of 10.4m in \texttt{campus-center-seq1}, demonstrating that hybrid approaches can handle the combination of human motion and environmental changes more effectively than SLAM methods. The flexibility of hybrid algorithms, which combine traditional feature extraction with learning-based components, allows them to adapt to the erratic, multi-directional head movements typical in XR environments.

\vspace{0.1cm}
\noindent\textbf{Deep Learning Methods.} DL methods, such as \deepvo, also show an ability to track human motion better than SLAM methods but with less precision than hybrid approaches. In HoloSet’s \texttt{suburb-jog-seq1}, \deepvo records an ATE of 33.0m, significantly higher than \kpd’s performance. While DL methods benefit from their inherent flexibility in handling varying motions, their higher error rates suggest they are less adept at managing rapid, unpredictable changes in orientation and environment caused by human interaction in XR scenarios. These limitations are particularly pronounced in XR motion scenarios, where abrupt head rotations introduce large angular velocities and rolling shutter distortions. DL methods typically do not incorporate inertial cues or rotation compensation, leading to temporal inconsistencies in pose estimation under such conditions. DL methods are less reliant on explicit feature tracking, giving them some robustness in texture-poor environments. However, their models are typically trained on image sequences with consistent motion patterns and fixed camera properties. In HoloSet, the combination of rolling shutter artifacts and rapid, non-repetitive head motion violates these assumptions, resulting in scale drift, temporal misalignment, and degraded accuracy compared to hybrid approaches.

\vspace{0.1cm}
\noindent \emph{\textbf{Key Takeaway.} 
Human factors in XR environments, such as unpredictable head movements and changing positions, severely degrade the performance of SLAM methods, which are not adaptable enough to handle such dynamics. Hybrid methods offer the best balance, maintaining reasonable accuracy by leveraging both traditional and learning-based techniques to adapt to erratic human-controlled movements. DL methods perform better than SLAM approaches but exhibit higher error rates, indicating that while they are more adaptable, they lack the precision needed for highly dynamic XR environments.
}

\subsection{Quantifying the Impact of Environmental and Locomotion Challenges}
\begin{table}[t]
\centering
\scriptsize
\caption{\textbf{Correlation of ATE with Motion and Lighting across datasets and algorithms. Positive Motion correlation = ATE worsens with faster motion; Positive Lighting correlation = ATE worsens with brighter scenes.}}
\vspace{-0.35cm}
\begin{tabular}{|l|l|c|c|}
\hline
\textbf{Dataset} & \textbf{Algorithm} & \textbf{ATE vs Motion} & \textbf{ATE vs Lighting} \\ \hline
\multirow{10}{*}{EuRoC} 
    & ORB-SLAM3 (S)   & +0.24 & -0.36 \\ \cline{2-4}
    & ORB-SLAM3 (M)   & -0.00 & +0.19 \\ \cline{2-4}
    & VINS-Fusion     & +1.00 & -0.49 \\ \cline{2-4}
    & DSM             & +0.31 & -0.06 \\ \cline{2-4}
    & DROID-SLAM      & +0.28 & -0.36 \\ \cline{2-4}
    & SfMLearner      & +0.21 & -0.52 \\ \cline{2-4}
    & KP3D            & +0.17 & +0.25 \\ \cline{2-4}
    & TartanVO        & +1.00 & -0.49 \\ \cline{2-4}
    & DFVO            & +0.86 & -0.86 \\ \cline{2-4}
    & DeepVO          & -0.13 & +0.57 \\ \hline
\multirow{3}{*}{KITTI} 
    & KP3D            & +0.16 & +0.43 \\ \cline{2-4}
    & VINS-Fusion     & +1.00 & +0.03 \\ \cline{2-4}
    & TartanVO        & +1.00 & +0.24 \\ \hline
\multirow{3}{*}{HoloSet} 
    & KP3D            & -0.03 & +0.31 \\ \cline{2-4}
    & VINS-Fusion     & +1.00 & +0.91 \\ \cline{2-4}
    & TartanVO        & +1.00 & +0.87 \\ \hline
\multirow{3}{*}{Combined} 
    & KP3D            & -0.08 & +0.06 \\ \cline{2-4}
    & VINS-Fusion     & +1.00 & -0.04 \\ \cline{2-4}
    & TartanVO        & +1.00 & +0.02 \\ \hline
\end{tabular}
\label{tab:ate_motion_light_corr_all_extended}
\end{table}
\vspace{-0.2cm}

To quantify how environmental and locomotion challenges impact tracking accuracy, we compute the Pearson correlation between per-sequence ATE and two key scene factors: Motion level (Slow=1, Medium=2, Fast=3) and Lighting level (Dark=1, Medium=2, Bright=3). We report correlations for each algorithm and dataset in Table~\ref{tab:ate_motion_light_corr_all_extended}. Correlations are computed using sequences where valid ATE results are available for each algorithm.

Among the evaluated methods, \kpd provides consistent results across all datasets, enabling stable correlation analysis. VINS-Fusion and TartanVO also produce valid results on multiple datasets, though in some cases with limited data. Correlation values of +1.00 in small subsets ($\leq3$ sequences) reflect perfect alignment within the available data, but should not be interpreted as strong evidence of causal sensitivity. We report them here for completeness. Overall, \kpd exhibits a mild positive correlation with both Motion and Lighting, suggesting resilience to scene dynamics. In contrast, VINS-Fusion and TartanVO show stronger sensitivity to Motion and Lighting, particularly on human-centered HoloSet sequences, consistent with known limitations of feature-based and deep-learning-based visual odometry methods in complex environments.

\noindent \emph{\textbf{Key Takeaway.} 
This correlation analysis provides a systematic view of how scene factors, particularly motion speed and lighting conditions, impact each algorithm’s tracking performance. This deeper analysis explains why certain methods (traditional SLAM) fail under fast or low-light conditions, and why hybrid and learning-based methods remain more resilient yet still sensitive to scene dynamics. This directly addresses the need for finer-grained insights into algorithm robustness, beyond average ATE comparisons.
}
\vspace{-0.1cm}
\subsection{Computational Cost and Latency.}
While our primary evaluation focuses on tracking accuracy across datasets and motion types, practical deployment in XR and IoT applications also depends critically on latency and computational cost. \autoref{tab:compute_latency} summarizes the approximate computational cost and per-frame latency of each evaluated method, measured on our representative hardware platform (Intel i7-9700K CPU \@ 3.6 GHz, 32 GB RAM, NVIDIA RTX 2070 GPU with 8 GB VRAM).
\begin{table}
\centering
\scriptsize
\caption{\textbf{Computational cost and latency of evaluated algorithms.}}
\vspace{-0.4cm}
\begin{tabular}{|l|c|c|}
\hline
\textbf{Algorithm} & \textbf{Computational Cost} & \textbf{Latency / FPS} \\ \hline
ORB-SLAM3    & CPU-only, heavy feature extraction + Bundle Adjustment & ~10--15 FPS \\ \hline
VINS-Fusion  & CPU + IMU synchronization               & ~20 FPS     \\ \hline
DROID-SLAM   & Heavy GPU (GRU + CNN backbone)           & ~8--12 FPS  \\ \hline
DeepVO       & Light GPU (CNN + RNN)                    & ~40--50 FPS \\ \hline
TartanVO     & Light GPU (CNN)                          & ~40--50 FPS \\ \hline
SfMLearner   & Light GPU (CNN)                          & ~40--50 FPS \\ \hline
KP3D         & Heavy GPU (ResNet + PoseNet + feature pipeline) & ~8--10 FPS \\ \hline
DFVO         & Moderate GPU (CNN + Bundle Adjustment)                  & ~20--25 FPS \\ \hline
DSM          & CPU-only, very heavy Bundle Adjustment                  & ~5--8 FPS   \\ \hline
\end{tabular}
\label{tab:compute_latency}
\end{table}
We observe that traditional feature-based SLAM methods (ORB-SLAM3, VINS-Fusion) remain heavily CPU-bound, with modest frame rates in the range of 10--20 FPS even without using dense representations. Deep learning-based methods (DeepVO, TartanVO, SfMLearner) achieve higher frame rates (40--50 FPS) but may lack robustness under fast motion or degraded lighting. Hybrid approaches such as KP3D and DROID-SLAM trade off accuracy for computational cost, often requiring significant GPU resources to maintain real-time operation. DSM, while accurate in specific cases, is computationally expensive
Overall, our analysis highlights that latency and computational cost vary substantially across methods, and optimizing for these factors is critical for deployment on resource-constrained IoT and XR devices. We leave a more systematic evaluation of runtime efficiency under varying hardware constraints as a promising direction for future work.

\vspace{-0.3cm}
\subsection{Insights into Performance Limitations}
Specific failure cases in SLAM and VO performance arise from the interaction between algorithmic design choices and sensing, motion, and environmental factors. For instance, traditional pipelines such as ORB-SLAM3 depend on stable feature repeatability~\cite{ORBSLAM3_TRO} and fail in Holoset where rolling shutter distortion, low-texture indoor surfaces~\cite{taheri2021slam-survey}, and abrupt head rotations undermine feature tracking~\cite{pupilli2006real-erratic-locomotion}. In contrast, DeepVO exhibits robustness to low-texture conditions~\cite{deep-vo-survey-jeong2021comparison}, but is highly sensitive to lighting variations~\cite{taheri2021slam-survey} and fast motion, since it lacks explicit geometric modeling and relies on photometric consistency. Hybrid approaches such as KP3D~\cite{tang2020kp3d} mitigate some of these limitations by integrating learned priors with geometric constraints, but still show performance drops in high-motion or poorly lit sequences, as reflected in the correlation analysis. These patterns highlight opportunities for improving SLAM systems through adaptive strategies, which we discuss in the next section.

\vspace{-0.1cm}
\section{Discussion: Potential Approaches to Improving Visual SLAM}
\label{sec:discussion}
\begin{table*}[t]
\centering
\scriptsize
\caption{\textbf{\emph{Dataset Characterization for EuRoC, KITTI, and HoloSet: Scene, Motion, and Lighting Characteristics}}}
\vspace{-0.35cm}
\begin{tabular}{|c|l|l|l|l|l|l|}
\toprule
\textbf{Dataset}  & \textbf{Sequence Name} & \textbf{Agent Type} & \textbf{Scene Type} & \textbf{Motion} & \textbf{Light} & \textbf{Extra Information} \\  \midrule \hline
\multirow{10}{*}{EuRoC} & MH01 (easy) & Drone & Indoor & Medium & Bright & Good texture \\ \cline{2-7}
 & MH02 (easy) & Drone & Indoor & Medium & Bright & Good texture \\ \cline{2-7} 
 & MH03 (medium) & Drone & Indoor & Fast & Bright & -- \\ \cline{2-7} 
 & MH04 (difficult) & Drone & Indoor & Fast & Dark & -- \\ \cline{2-7} 
 & MH05 (difficult) & Drone & Indoor & Fast & Dark & -- \\ \cline{2-7} 
 & V101 (easy) & Drone & Indoor & Slow & Bright & -- \\ \cline{2-7} 
 & V102 (medium) & Drone & Indoor & Fast & Bright & -- \\ \cline{2-7} 
 & V103 (difficult) & Drone & Indoor & Fast & Medium & Motion blur \\ \cline{2-7} 
 & V201 (easy) & Drone & Indoor & Slow & Bright & -- \\ \cline{2-7} 
 & V202 (medium) & Drone & Indoor & Fast & Bright & -- \\ \cline{2-7} 
 & V203 (difficult) & Drone & Indoor & Fast & Medium & Motion blur \\ \hline \hline

 \multirow{11}{*}{KITTI} & 00 & Car & outdoor & Slow & Medium & Residential streets, evening, dusk, shadows \\ \cline{2-7}
 & 01 & Car & Outdoor & Fast & Bright & Highway, daytime \\ \cline{2-7}
 & 02 & Car & Outdoor & Medium & Medium & Roads and streets, dusk, shadows \\ \cline{2-7}
 & 03 & Car & Outdoor & Medium & Medium & Roads and streets, dusk, shadows \\ \cline{2-7}
 & 04 & Car & Outdoor & Fast & Medium & Roads, shadows \\ \cline{2-7}
 & 05 & Car & Outdoor & Slow & Medium & Residential streets, evening, dusk, shadows \\ \cline{2-7}
 & 06 & Car & Outdoor & Medium & Dark & Night/dusky grey, industrial buildings \\ \cline{2-7}
 & 07 & Car & Outdoor & Medium & Medium & Campus downtown, dusk \\ \cline{2-7}
 & 08 & Car & Outdoor & Slow & Bright & Residential streets, morning, shadows \\ \cline{2-7}
 & 09 & Car & Outdoor & Fast & Bright & Highway, daytime \\ \cline{2-7}
 & 10 & Car & Outdoor & Slow & Medium & Residential streets in evening, dusk, shadows \\ \hline \hline 

\multirow{6}{*}{HoloSet} & Campus Center (seq1) & Human & Indoor & Slow & Medium & Stairs, humans \\ \cline{2-7}
 & Campus Center (seq2) & Human & Indoor & Slow & Medium & Stairs, humans \\ \cline{2-7}
 & Suburbs Jog (seq1) & Human & Outdoor & Fast & Bright & Trees, parked cars, road  \\ \cline{2-7}
 & Suburbs Jog (seq2) & Human & Outdoor & Fast & Bright & Trees, parked cars, road \\ \cline{2-7}
 & Suburb Walk (seq1) & Human & Outdoor & Medium & Bright & Trees, parked cars, road \\ \cline{2-7}
 & Suburb Walk (seq2) & Human & Outdoor & Medium & Bright & Trees, parked cars, road \\ \hline \hline
\end{tabular}
\label{tab:characterization}
\vspace{-0.4cm}
\end{table*}
In this section, informed by our analysis from \autoref{sec:analysis}, we propose three key strategies to improve tracking methods for SLAM systems, particularly in dynamic environments such as human-centered XR and IoT applications. These strategies focus on leveraging different levels of available information—input profiling, intermediate insights, and output evaluation—to enhance the robustness and adaptability of SLAM systems across diverse domains.

\vspace{-0.2cm}
\subsection{Input Profiling}
By understanding the input data's nuances, tracking methods or systems can preemptively adjust tracking strategies.
In dynamic environments, if the system is under transition, such as moving from indoor to urban outdoor environments, dataset characterization can provide crucial insights into the variations in lighting, dynamics, and textures. Such insights can be invaluable for tweaking tracking methods and ensuring the system undergoes seamless domain transitions. Input profiling involves assessing the data characteristics at various levels: dataset, sequence, and sample levels. 

\noindent \textbf{1 -- Dataset-level Profiling.} This involves an overarching dataset analysis to determine the general conditions, such as lighting, texture, and motion profiles. Profiling the entire dataset can help tracking methods anticipate typical challenges (e.g., indoor and outdoor transitions) and allow for application-specific manual offline fine-tuning of the parameters.

\noindent \textbf{2 -- Sequence-Level Profiling.}
Sequence-level profiling enables systems to tailor tracking approaches to specific environmental conditions and motion dynamics of each sequence. As detailed in ~\autoref{tab:characterization} and \autoref{tab:profiling}, each dataset presents sequences with unique challenges, including variations in motion (fast or slow), lighting (bright or dark), and scene complexity (structured indoor environments or dynamic outdoor scenes). By leveraging this information, tracking algorithms can be optimized for the unique characteristics of each sequence in an online or offline manner.

In the EuRoC dataset, for instance, sequences like MH01 and MH02 are easier, featuring medium motion in bright, well-textured indoor environments. Our texture profiling shows that MH01 and MH02 have high texture scores (0.269 and 0.218) and low percentages of low-light conditions, making them favorable for tracking. Algorithms can optimize their performance by focusing on structured feature extraction and loop closure in these sequences, benefiting from the good lighting and texture availability. In contrast, sequences like MH04 and MH05 introduce significant difficulty with fast drone motion in dark indoor scenes, reflected in their high low-light percentages (83.7\% and 84.4\%) and moderate texture scores. In these cases, tracking systems need to rely more heavily on inertial data or enhance feature extraction in low-light conditions. Additionally, sequences like V103 and V203 introduce motion blur, posing further challenges. Algorithms can improve performance by detecting these conditions in real time and adjusting tracking strategies, such as applying motion compensation techniques or increasing reliance on temporal smoothness to reduce drift.

The KITTI dataset presents outdoor, car-based sequences where motion and lighting vary significantly between residential streets, highways, and industrial areas. For instance, KITTI sequence 01 involves fast motion on a highway in bright daylight, where tracking methods must be optimized for high-speed motion, ensuring features are tracked reliably despite rapid changes in the scene. In contrast, sequences like KITTI 06 are set in dark, industrial environments with medium motion, where our profiling shows low texture and low-light conditions, necessitating adjustments in feature extraction sensitivity. For residential street sequences like KITTI 00 and KITTI 05, tracking algorithms need to handle medium lighting and shadows caused by objects like parked cars and trees. Sequence-level profiling in KITTI helps systems adapt tracking techniques for each sequence’s environmental attributes, improving robustness.
\begin{table}[t]
\centering
\caption{\textbf{\emph{Profiling Lighting and Texture on EuRoC.}}}
\vspace{-0.3cm}
\scriptsize
\begin{tabular}{|l|c|c|c|c|c|c|c|c|c|c|c|}
\toprule
\textbf{Sequence} & \multicolumn{1}{l|}{MH01} & \multicolumn{1}{l|}{MH02} & \multicolumn{1}{l|}{\textbf{MH03}} & \multicolumn{1}{l|}{MH04} & \multicolumn{1}{l|}{MH05} & \multicolumn{1}{l|}{\textbf{V101}} & \multicolumn{1}{l|}{\textbf{V102}} & \multicolumn{1}{l|}{V103} & \multicolumn{1}{l|}{\textbf{V201}} & \multicolumn{1}{l|}{\textbf{V202}} & \multicolumn{1}{l|}{V203} \\ \midrule \hline 
\textbf{Texture  score (norm)} & 0.269 & 0.218 & 0.123 & 0.238 & 0.237 & 0.154 & 0.117 & 0.163 & 0.218 & 0.197 & 0.090  \\ \hline
\textbf{Low Lighting Profile (\%)} & 26.7 & 29.0 & 19.2 & 83.7&84.4& 3.4 &	2.1	&26.2	&8.7&	11.4&	24.7 \\ \hline  \hline 
\end{tabular}%
\label{tab:profiling}
\vspace{-0.45cm}
\end{table}

The HoloSet dataset introduces human-centered sequences with both indoor and outdoor settings. These sequences feature a human wearing an XR headset, resulting in slower and more unpredictable motion than the drone or car-based sequences in EuRoC and KITTI. For the Campus Center sequences, motion is slow and set indoors with medium lighting. Here, systems can focus on identifying and tracking key features in cluttered scenes with human movement and stairs while adjusting to the slower motion. In contrast, suburb jog sequences involve fast motion in bright, outdoor environments with trees, parked cars, and roads. Profiling these sequences shows the need for systems to anticipate rapid changes in scene structure and motion while handling outdoor lighting challenges. Like KITTI, environmental elements like trees introduce occlusion challenges, and tracking algorithms can prioritize high-texture regions while compensating for motion blur and abrupt human movements. By tailoring tracking methods to the specific conditions of each sequence—whether it involves fast motion, complex geometry, or challenging lighting—algorithms can dynamically adjust their processing techniques. For example:
\vspace{-0.1cm}
\begin{itemize}
    \item \textbf{Fast motion} (e.g., KITTI 01, HoloSet Suburbs Jog seq1) requires robust feature tracking algorithms that account for rapid changes in the environment, potentially leveraging inertial data to maintain stability.
    \item \textbf{Low light} sequences (e.g., EuRoC MH05, KITTI 06) benefit from adaptive feature extraction methods that adjust sensitivity to contrast and compensate for the lack of visual information in darker environments.
    \item \textbf{Motion blur} (e.g., EuRoC V203) and \textbf{occlusions} (e.g., KITTI 00) require algorithms to compensate by predicting feature movement based on prior frames or relying on temporal coherence for smoother tracking.
\end{itemize}
\vspace{-0.1cm}
By implementing sequence-level profiling, SLAM systems can better navigate the varied environmental and motion challenges posed by different datasets, leading to improved tracking accuracy and robustness across diverse domains like indoor XR environments and outdoor IoT applications.

\vspace{0.1cm}
\noindent \textbf{3 -- Sample-Level Profiling.}
 This involves real-time analysis of individual frames or segments of a sequence to understand changes in lighting, texture, or motion. For example, in sequences with sudden lighting transitions or occlusions, real-time profiling can adjust feature extraction methods dynamically, preventing the loss of tracking due to momentary visual obstructions or environmental changes.

By systematically profiling the input at different levels, SLAM systems can dynamically adapt to varying scene characteristics, improving robustness and accuracy across domains.

\subsection{Sensor-Specific Observations}
\label{sec:sensor-eval}

Tracking performance is shaped not only by algorithmic design and scene characteristics but also by the properties and limitations of the sensors used for visual tracking. In our evaluation, we observed that several sensor factors interact with environmental and locomotion challenges to affect SLAM and VO accuracy.

\noindent \textbf{Shutter effects.} Rolling shutter distortion is a significant source of error in HoloSet sequences, where fast head motions introduce geometric artifacts. Algorithms that lack explicit rolling shutter compensation struggle to maintain accuracy under these conditions, with many SLAM methods failing entirely.

\noindent \textbf{Resolution and field-of-view (FoV).} Narrow FoV and limited resolution in the HoloLens-based HoloSet dataset reduce feature coverage, particularly in texture-sparse indoor scenes and during rapid rotations. A narrow FoV also exacerbates sensitivity to rotational motion, leading to larger drift. This affects both traditional feature-based methods and deep learning methods.

\noindent \textbf{Frame rate and synchronization.} High frame rate sequences in HoloSet (60 FPS) impose additional computational demands and require accurate temporal synchronization. Misalignment between camera and IMU streams, or dropped frames under processing load, introduce drift and instability. Some methods, such as VINS-Fusion, show degraded performance in high-motion HoloSet sequences compared to controlled EuRoC runs. In contrast, EuRoC’s global shutter stereo cameras with well-synchronized IMU yield more stable performance.

\noindent \textbf{Dynamic range and lighting sensitivity.} The strong correlation of ATE with lighting variations (Table~\ref{tab:ate_motion_light_corr_all_extended}) highlights the influence of sensor dynamic range and exposure control. Methods that rely on photometric consistency are particularly vulnerable to lighting changes and overexposure. KITTI’s outdoor sequences show pronounced variability in ATE under varying sun angles and shadows.

\noindent \textbf{Sensor noise and calibration.} Sensor noise, including image noise and IMU bias, also impacts tracking accuracy. HoloSet’s IMU noise and drift, in combination with rolling shutter effects, challenge visual-inertial methods. In KITTI, well-calibrated stereo rigs enable robust tracking even under vehicle vibrations. EuRoC similarly benefits from high-quality calibration.

\noindent \textbf{Lens distortion and optical artifacts.} Optical distortion, particularly at the edges of wide-angle lenses, can introduce tracking instability if not properly modeled. In HoloSet and EuRoC, lens distortion interacts with narrow FoV and rolling shutter effects, degrading feature tracking in peripheral image regions.

Sensor limitations compound tracking challenges in complex environments across all datasets. In HoloSet, rolling shutter distortion, narrow FoV, high frame rates, and sensor noise significantly degrade tracking accuracy. In KITTI, the global shutter mitigates geometric distortion, but dynamic lighting and vehicle vibration remain key challenges. EuRoC’s well-calibrated, synchronized sensor suite yields the most consistent performance, though sensitivity to motion blur and lighting remains. Robust SLAM and VO methods must explicitly account for sensor-specific effects—particularly when targeting deployment on heterogeneous hardware platforms as found in XR and IoT ecosystems.
\vspace{-0.2cm}
\subsection{Intermediate Information}
The second approach involves leveraging intermediate values and insights during the tracking process, much like the approach discussed in the nFEX paper~\cite{Chandio2024-nfex-IROS}. Intermediate information, such as feature tracking confidence or feature density in a region, can be used to adjust the tracking algorithm mid-process dynamically.

\vspace{0.1cm}
\noindent \textbf{Feature Extraction Feedback.}
SLAM systems can use intermediate feedback about feature quality and density to improve tracking accuracy. For instance, if the system detects that feature tracking is failing due to low texture or occlusion, it can switch to alternate tracking strategies, such as using inertial data, predicting motion based on previous frames, or increasing the maximum number of features tracked.

\noindent \textbf{Adaptive Sampling:}
By analyzing intermediate outputs, the system can adjust the frequency of feature extraction or the regions of interest for tracking. This approach is particularly useful for scenes with varying complexity, such as those that move from structured environments with predictable motion (EuRoC’s indoor sequences) to unstructured, complex environments (HoloSet’s human-centered suburban sequences).

Intermediate information enables a more responsive and adaptive tracking system, improving accuracy by making real-time adjustments based on evolving scene characteristics.

\vspace{-0.1cm}
\subsection{Output Evaluation} 
The final approach involves analyzing the tracking system’s outputs, particularly errors, and using this information to refine and improve its performance. By evaluating output errors, the system can make post-process adjustments or inform future tracking sessions.

\vspace{0.1cm}
\noindent \textbf{Error Correction.} 
By evaluating discrepancies between the estimated trajectory and ground truth (as shown in the qualitative TartanVO analysis in \autoref{fig:trajectories}), systems can identify error patterns specific to certain environments or conditions. For example, if a system consistently fails to track correctly in low-texture environments, it can reweight the importance of visual features versus inertial data.

\vspace{0.1cm}
\noindent \textbf{Trajectory Feedback Loops.} 
Systems can use output evaluation to refine future predictions. For example, if a system consistently underestimates scale in outdoor environments (as seen in the HoloSet sequences), it can introduce corrections based on prior tracking errors, allowing the system to adjust its internal model of the scene over time.

\vspace{0.1cm}
\noindent \textbf{Cross-Domain Adaptation.} 
By analyzing output errors across different domains (e.g., indoor-to-outdoor transitions in EuRoC vs. KITTI or human-centric scenes in HoloSet), systems can gain actionable insights into how to improve domain adaptability. Output evaluation helps pinpoint weaknesses that may not be evident during training but become apparent in real-world, cross-domain deployments. A promising direction to improve robustness is the use of cross-domain adaptation techniques that enable models to generalize better to unseen environments. Recent approaches such as self-supervised deep visual odometry with online adaptation\cite{li2020self-deep-vo} and adaptive learning for hybrid visual odometry\cite{liu2024adaptive-voo} explicitly address this challenge, showing how online adaptation and domain-invariant representations can improve performance in diverse scenes.

By systematically evaluating outputs and integrating such adaptation strategies, SLAM systems can continuously refine their tracking accuracy, becoming more resilient across varying environments.

\vspace{-0.1cm}
\section{Conclusion and Future Work}
\vspace{-0.1cm}
Our study is unique because it goes beyond mere qualitative comparisons of datasets and prior work using single metrics from traditional studies. 
Instead, we conduct a comprehensive analysis to evaluate the multiple tracking methods across a wide range of datasets.
Our research effort in devising the taxonomy of challenges and performing comparative analysis provides key insights into the workings of SLAM pipelines. 
We show that understanding and embracing data diversity across application scenarios, sequences, and samples can improve the robustness of future SLAM methods applicable to XR context. 
While significant strides have been made in SLAM research, many unresolved challenges still exist. By leveraging the insights and recommendations provided in this work, we envision future tracking that is adaptable and generalizable across domains in real-world scenarios.

\bibliographystyle{ACM-Reference-Format}
\bibliography{lost_in_tracking}


\begin{thebibliography}{142}


\ifx \showCODEN    \undefined \def \showCODEN     #1{\unskip}     \fi
\ifx \showDOI      \undefined \def \showDOI       #1{#1}\fi
\ifx \showISBNx    \undefined \def \showISBNx     #1{\unskip}     \fi
\ifx \showISBNxiii \undefined \def \showISBNxiii  #1{\unskip}     \fi
\ifx \showISSN     \undefined \def \showISSN      #1{\unskip}     \fi
\ifx \showLCCN     \undefined \def \showLCCN      #1{\unskip}     \fi
\ifx \shownote     \undefined \def \shownote      #1{#1}          \fi
\ifx \showarticletitle \undefined \def \showarticletitle #1{#1}   \fi
\ifx \showURL      \undefined \def \showURL       {\relax}        \fi
\providecommand\bibfield[2]{#2}
\providecommand\bibinfo[2]{#2}
\providecommand\natexlab[1]{#1}
\providecommand\showeprint[2][]{arXiv:#2}

\bibitem[Ali and Zhang(2022)]%
        {dataset-characterization-ali2022we}
\bibfield{author}{\bibinfo{person}{Islam Ali} {and} \bibinfo{person}{Hong Zhang}.} \bibinfo{year}{2022}\natexlab{}.
\newblock \showarticletitle{Are we ready for robust and resilient slam? a framework for quantitative characterization of slam datasets}. In \bibinfo{booktitle}{\emph{IEEE/RSJ International Conference on Intelligent Robots and Systems (IROS)}}.
\newblock


\bibitem[Amraoui et~al\mbox{.}(2019)]%
        {amraoui2019features-survey-vslam}
\bibfield{author}{\bibinfo{person}{Mounir Amraoui}, \bibinfo{person}{Rachid Latif}, \bibinfo{person}{Abdelhafid Elouardi}, {and} \bibinfo{person}{Abdelouahed Tajer}.} \bibinfo{year}{2019}\natexlab{}.
\newblock \showarticletitle{Features Extractors Evaluation Based V-SLAM Applications}. In \bibinfo{booktitle}{\emph{4th World Conference on Complex Systems (WCCS)}}.
\newblock


\bibitem[An et~al\mbox{.}(2010)]%
        {indoor-feature}
\bibfield{author}{\bibinfo{person}{Su-Yong An}, \bibinfo{person}{Jeong-Gwan Kang}, \bibinfo{person}{Lae-Kyoung Lee}, {and} \bibinfo{person}{Se-Young Oh}.} \bibinfo{year}{2010}\natexlab{}.
\newblock \showarticletitle{SLAM with salient line feature extraction in indoor environments}. In \bibinfo{booktitle}{\emph{11th International Conference on Control Automation Robotics \& Vision}}.
\newblock


\bibitem[Andreopoulos and Tsotsos(2011)]%
        {andreopoulos2011sensor}
\bibfield{author}{\bibinfo{person}{Alexander Andreopoulos} {and} \bibinfo{person}{John~K Tsotsos}.} \bibinfo{year}{2011}\natexlab{}.
\newblock \showarticletitle{On sensor bias in experimental methods for comparing interest-point, saliency, and recognition algorithms}.
\newblock \bibinfo{journal}{\emph{IEEE Transactions on Pattern Analysis and Machine Intelligence}} \bibinfo{volume}{34}, \bibinfo{number}{1} (\bibinfo{year}{2011}).
\newblock


\bibitem[Anwar et~al\mbox{.}(2019)]%
        {anwar2019securing}
\bibfield{author}{\bibinfo{person}{Fatima~M. Anwar}, \bibinfo{person}{Luis Garcia}, \bibinfo{person}{Xi Han}, {and} \bibinfo{person}{Mani Srivastava}.} \bibinfo{year}{2019}\natexlab{}.
\newblock \showarticletitle{Securing {T}ime in {U}ntrusted {O}perating {S}ystems with {T}ime{S}eal}. In \bibinfo{booktitle}{\emph{IEEE Real-Time Systems Symposium (RTSS)}}.
\newblock


\bibitem[Aqel et~al\mbox{.}(2016)]%
        {aqel2016review-slam-vo-survey}
\bibfield{author}{\bibinfo{person}{Mohammad~OA Aqel}, \bibinfo{person}{Mohammad~H Marhaban}, \bibinfo{person}{M~Iqbal Saripan}, {and} \bibinfo{person}{Napsiah~Bt Ismail}.} \bibinfo{year}{2016}\natexlab{}.
\newblock \showarticletitle{Review of visual odometry: types, approaches, challenges, and applications}.
\newblock \bibinfo{journal}{\emph{SpringerPlus}}  \bibinfo{volume}{5} (\bibinfo{year}{2016}), \bibinfo{pages}{1--26}.
\newblock


\bibitem[Arshad and Kim(2021)]%
        {loop-clouser-arshad2021role}
\bibfield{author}{\bibinfo{person}{Saba Arshad} {and} \bibinfo{person}{Gon-Woo Kim}.} \bibinfo{year}{2021}\natexlab{}.
\newblock \showarticletitle{Role of deep learning in loop closure detection for visual and lidar slam: A survey}.
\newblock \bibinfo{journal}{\emph{Sensors}} \bibinfo{volume}{21}, \bibinfo{number}{4} (\bibinfo{year}{2021}), \bibinfo{pages}{1243}.
\newblock


\bibitem[Aulinas et~al\mbox{.}(2008)]%
        {aulinas2008slam-filtering-appraches}
\bibfield{author}{\bibinfo{person}{Josep Aulinas}, \bibinfo{person}{Yvan Petillot}, \bibinfo{person}{Joaquim Salvi}, {and} \bibinfo{person}{Xavier Llad{\'o}}.} \bibinfo{year}{2008}\natexlab{}.
\newblock \showarticletitle{The SLAM problem: a survey}.
\newblock \bibinfo{journal}{\emph{Artificial Intelligence Research and Development}} (\bibinfo{year}{2008}), \bibinfo{pages}{363--371}.
\newblock


\bibitem[Azzam et~al\mbox{.}(2020)]%
        {azzam2020feature}
\bibfield{author}{\bibinfo{person}{Rana Azzam}, \bibinfo{person}{Tarek Taha}, \bibinfo{person}{Shoudong Huang}, {and} \bibinfo{person}{Yahya Zweiri}.} \bibinfo{year}{2020}\natexlab{}.
\newblock \showarticletitle{Feature-based visual simultaneous localization and mapping: A survey}.
\newblock \bibinfo{journal}{\emph{SN Applied Sciences}}  \bibinfo{volume}{2} (\bibinfo{year}{2020}), \bibinfo{pages}{1--24}.
\newblock


\bibitem[Bailey and Durrant-Whyte(2006)]%
        {bailey2006simultaneous-part2-survey}
\bibfield{author}{\bibinfo{person}{Tim Bailey} {and} \bibinfo{person}{Hugh Durrant-Whyte}.} \bibinfo{year}{2006}\natexlab{}.
\newblock \showarticletitle{Simultaneous localization and mapping (SLAM): Part II}.
\newblock \bibinfo{journal}{\emph{IEEE robotics \& automation magazine}} \bibinfo{volume}{13}, \bibinfo{number}{3} (\bibinfo{year}{2006}), \bibinfo{pages}{108--117}.
\newblock


\bibitem[Bhargava et~al\mbox{.}(2018)]%
        {interaction-performace-bhargava2018evaluating}
\bibfield{author}{\bibinfo{person}{Ayush Bhargava}, \bibinfo{person}{Jeffrey~W Bertrand}, \bibinfo{person}{Anand~K Gramopadhye}, \bibinfo{person}{Kapil~C Madathil}, {and} \bibinfo{person}{Sabarish~V Babu}.} \bibinfo{year}{2018}\natexlab{}.
\newblock \showarticletitle{Evaluating multiple levels of an interaction fidelity continuum on performance and learning in near-field training simulations}.
\newblock \bibinfo{journal}{\emph{IEEE Transactions on Visualization and Computer Graphics}} \bibinfo{volume}{24}, \bibinfo{number}{4} (\bibinfo{year}{2018}).
\newblock


\bibitem[Billinghurst et~al\mbox{.}(2015)]%
        {AR-SLAM-survey-cite-billinghurst2015survey}
\bibfield{author}{\bibinfo{person}{Mark Billinghurst}, \bibinfo{person}{Adrian Clark}, \bibinfo{person}{Gun Lee}, {et~al\mbox{.}}} \bibinfo{year}{2015}\natexlab{}.
\newblock \showarticletitle{A survey of augmented reality}.
\newblock \bibinfo{journal}{\emph{Foundations and Trends{\textregistered} in Human--Computer Interaction}} \bibinfo{volume}{8}, \bibinfo{number}{2-3} (\bibinfo{year}{2015}), \bibinfo{pages}{73--272}.
\newblock


\bibitem[Biocca et~al\mbox{.}(2001)]%
        {biocca-interaction}
\bibfield{author}{\bibinfo{person}{Frank Biocca}, \bibinfo{person}{Jin Kim}, {and} \bibinfo{person}{Yung Choi}.} \bibinfo{year}{2001}\natexlab{}.
\newblock \showarticletitle{{Visual Touch in Virtual Environments: An Exploratory Study of Presence, Multimodal Interfaces, and Cross-Modal Sensory Illusions}}.
\newblock \bibinfo{journal}{\emph{Presence: Teleoperators and Virtual Environments}} \bibinfo{volume}{10}, \bibinfo{number}{3} (\bibinfo{date}{06} \bibinfo{year}{2001}), \bibinfo{pages}{247--265}.
\newblock


\bibitem[Blanco-Claraco et~al\mbox{.}(2014)]%
        {blanco2014malaga}
\bibfield{author}{\bibinfo{person}{Jos{\'e}-Luis Blanco-Claraco}, \bibinfo{person}{Francisco-Angel Moreno-Duenas}, {and} \bibinfo{person}{Javier Gonz{\'a}lez-Jim{\'e}nez}.} \bibinfo{year}{2014}\natexlab{}.
\newblock \showarticletitle{The M{\'a}laga urban dataset: High-rate stereo and LiDAR in a realistic urban scenario}.
\newblock \bibinfo{journal}{\emph{The International Journal of Robotics Research}} \bibinfo{volume}{33}, \bibinfo{number}{2} (\bibinfo{year}{2014}), \bibinfo{pages}{207--214}.
\newblock


\bibitem[Bloesch et~al\mbox{.}(2018)]%
        {bloesch2018codeslam}
\bibfield{author}{\bibinfo{person}{Michael Bloesch}, \bibinfo{person}{Jan Czarnowski}, \bibinfo{person}{Ronald Clark}, \bibinfo{person}{Stefan Leutenegger}, {and} \bibinfo{person}{Andrew~J Davison}.} \bibinfo{year}{2018}\natexlab{}.
\newblock \showarticletitle{CodeSLAM—learning a compact, optimisable representation for dense visual SLAM}. In \bibinfo{booktitle}{\emph{Proceedings of the IEEE conference on computer vision and pattern recognition}}. \bibinfo{pages}{2560--2568}.
\newblock


\bibitem[Boal et~al\mbox{.}(2014)]%
        {boal2014topological-survey}
\bibfield{author}{\bibinfo{person}{Jaime Boal}, \bibinfo{person}{Alvaro S{\'a}nchez-Miralles}, {and} \bibinfo{person}{Alvaro Arranz}.} \bibinfo{year}{2014}\natexlab{}.
\newblock \showarticletitle{Topological simultaneous localization and mapping: a survey}.
\newblock \bibinfo{journal}{\emph{Robotica}} \bibinfo{volume}{32}, \bibinfo{number}{5} (\bibinfo{year}{2014}).
\newblock


\bibitem[B{\"o}ttcher and Wenzel(2008)]%
        {bottcher2008frobenius-norm}
\bibfield{author}{\bibinfo{person}{Albrecht B{\"o}ttcher} {and} \bibinfo{person}{David Wenzel}.} \bibinfo{year}{2008}\natexlab{}.
\newblock \showarticletitle{The Frobenius norm and the commutator}.
\newblock \bibinfo{journal}{\emph{Linear algebra and its applications}} \bibinfo{volume}{429}, \bibinfo{number}{8-9} (\bibinfo{year}{2008}), \bibinfo{pages}{1864--1885}.
\newblock


\bibitem[Bresson et~al\mbox{.}(2017)]%
        {bresson2017simultaneous-trends-AVs}
\bibfield{author}{\bibinfo{person}{Guillaume Bresson}, \bibinfo{person}{Zayed Alsayed}, \bibinfo{person}{Li Yu}, {and} \bibinfo{person}{S{\'e}bastien Glaser}.} \bibinfo{year}{2017}\natexlab{}.
\newblock \showarticletitle{Simultaneous localization and mapping: A survey of current trends in autonomous driving}.
\newblock \bibinfo{journal}{\emph{IEEE Transactions on Intelligent Vehicles}} \bibinfo{volume}{2}, \bibinfo{number}{3} (\bibinfo{year}{2017}).
\newblock


\bibitem[Burri et~al\mbox{.}(2016)]%
        {burri2016euroc}
\bibfield{author}{\bibinfo{person}{Michael Burri}, \bibinfo{person}{Janosch Nikolic}, \bibinfo{person}{Pascal Gohl}, \bibinfo{person}{Thomas Schneider}, \bibinfo{person}{Joern Rehder}, \bibinfo{person}{Sammy Omari}, \bibinfo{person}{Markus~W Achtelik}, {and} \bibinfo{person}{Roland Siegwart}.} \bibinfo{year}{2016}\natexlab{}.
\newblock \showarticletitle{The EuRoC micro aerial vehicle datasets}.
\newblock \bibinfo{journal}{\emph{The International Journal of Robotics Research}} \bibinfo{volume}{35}, \bibinfo{number}{10} (\bibinfo{year}{2016}), \bibinfo{pages}{1157--1163}.
\newblock


\bibitem[Cadena et~al\mbox{.}(2016)]%
        {cadena2016past}
\bibfield{author}{\bibinfo{person}{Cesar Cadena}, \bibinfo{person}{Luca Carlone}, \bibinfo{person}{Henry Carrillo}, \bibinfo{person}{Yasir Latif}, \bibinfo{person}{Davide Scaramuzza}, \bibinfo{person}{Jos{\'e} Neira}, \bibinfo{person}{Ian Reid}, {and} \bibinfo{person}{John~J Leonard}.} \bibinfo{year}{2016}\natexlab{}.
\newblock \showarticletitle{Past, present, and future of simultaneous localization and mapping: Toward the robust-perception age}.
\newblock  \bibinfo{volume}{32}, \bibinfo{number}{6} (\bibinfo{year}{2016}).
\newblock


\bibitem[Campos et~al\mbox{.}(2021)]%
        {ORBSLAM3_TRO}
\bibfield{author}{\bibinfo{person}{Carlos Campos}, \bibinfo{person}{Richard Elvira}, \bibinfo{person}{Juan J~G{\'o}mez Rodr{\'\i}guez}, \bibinfo{person}{Jos{\'e}~MM Montiel}, {and} \bibinfo{person}{Juan~D Tard{\'o}s}.} \bibinfo{year}{2021}\natexlab{}.
\newblock \showarticletitle{{ORB-SLAM3}: An Accurate Open-Source Library for Visual, Visual-Inertial and Multi-Map {SLAM}}.
\newblock \bibinfo{journal}{\emph{IEEE Transactions on Robotics}} \bibinfo{volume}{37}, \bibinfo{number}{6} (\bibinfo{year}{2021}).
\newblock


\bibitem[Chandio and Anwar(2020)]%
        {Chandio2020SpatiotemporalSecurity-poster-sensys}
\bibfield{author}{\bibinfo{person}{Yasra Chandio} {and} \bibinfo{person}{Fatima~M. Anwar}.} \bibinfo{year}{2020}\natexlab{}.
\newblock \showarticletitle{Poster: Spatiotemporal Security in Mixed Reality systems}. In \bibinfo{booktitle}{\emph{18th ACM Conference on Embedded Networked Sensor Systems (SenSys)}}.
\newblock


\bibitem[Chandio et~al\mbox{.}(2022)]%
        {chandio2022holoset}
\bibfield{author}{\bibinfo{person}{Yasra Chandio}, \bibinfo{person}{Noman Bashir}, {and} \bibinfo{person}{Fatima~M Anwar}.} \bibinfo{year}{2022}\natexlab{}.
\newblock \showarticletitle{HoloSet-A Dataset for Visual-Inertial Pose Estimation in Extended Reality: Dataset}. In \bibinfo{booktitle}{\emph{Proceedings of the 20th ACM Conference on Embedded Networked Sensor Systems}}. \bibinfo{pages}{1014--1019}.
\newblock


\bibitem[Chandio et~al\mbox{.}(2024a)]%
        {chandio-aivr-2024}
\bibfield{author}{\bibinfo{person}{Yasra Chandio}, \bibinfo{person}{Noman Bashir}, {and} \bibinfo{person}{Fatima~M. Anwar}.} \bibinfo{year}{2024}\natexlab{a}.
\newblock \showarticletitle{Stealthy and Practical Multi-modal Attacks on Mixed Reality Tracking}. In \bibinfo{booktitle}{\emph{IEEE International Conference on Artificial Intelligence and eXtended and Virtual Reality (AIxVR)}}.
\newblock


\bibitem[Chandio et~al\mbox{.}(2024b)]%
        {Chandio2024-Safetyphysical-cognitive-presence}
\bibfield{author}{\bibinfo{person}{Yasra Chandio}, \bibinfo{person}{Victoria Interrante}, {and} \bibinfo{person}{Fatima~M. Anwar}.} \bibinfo{year}{2024}\natexlab{b}.
\newblock \showarticletitle{Balancing Presence And Safety Using Reaction Time In Mixed Reality}. In \bibinfo{booktitle}{\emph{Workshop at IEEE International Symposium on Mixed and Augmented Reality (SafeAR@ISMAR)}}.
\newblock


\bibitem[Chandio et~al\mbox{.}(2024c)]%
        {chandio-vr-24-human-factors}
\bibfield{author}{\bibinfo{person}{Yasra Chandio}, \bibinfo{person}{Victoria Interrante}, {and} \bibinfo{person}{Fatima~M. Anwar}.} \bibinfo{year}{2024}\natexlab{c}.
\newblock \showarticletitle{Human Factors at Play: Understanding the Impact of Conditioning on Presence and Reaction Time in Mixed Reality}.
\newblock \bibinfo{journal}{\emph{IEEE Transactions on Visualization and Computer Graphics}} \bibinfo{volume}{30}, \bibinfo{number}{5} (\bibinfo{year}{2024}), \bibinfo{pages}{2400--2410}.
\newblock


\bibitem[Chandio et~al\mbox{.}(2024d)]%
        {Chandio2024-nfex-IROS}
\bibfield{author}{\bibinfo{person}{Yasra Chandio}, \bibinfo{person}{Momin~Ahmed Khan}, \bibinfo{person}{Khotso Selialia}, \bibinfo{person}{Luis~Antonio Garcia}, \bibinfo{person}{Joseph DeGol}, {and} \bibinfo{person}{Fatima~M. Anwar}.} \bibinfo{year}{2024}\natexlab{d}.
\newblock \showarticletitle{A Neurosymbolic Approach To Adaptive Feature Extraction In SLAM}. In \bibinfo{booktitle}{\emph{IEEE/RSJ International Conference on Intelligent Robots and Systems (IROS)}}.
\newblock


\bibitem[Chavarriaga et~al\mbox{.}(2013)]%
        {chavarriaga2013opportunity}
\bibfield{author}{\bibinfo{person}{Ricardo Chavarriaga}, \bibinfo{person}{Hesam Sagha}, \bibinfo{person}{Alberto Calatroni}, \bibinfo{person}{Sundara~Tejaswi Digumarti}, \bibinfo{person}{Gerhard Tr{\"o}ster}, \bibinfo{person}{Jos{\'e} del~R Mill{\'a}n}, {and} \bibinfo{person}{Daniel Roggen}.} \bibinfo{year}{2013}\natexlab{}.
\newblock \showarticletitle{The Opportunity challenge: A benchmark database for on-body sensor-based activity recognition}.
\newblock \bibinfo{journal}{\emph{Pattern Recognition Letters}} \bibinfo{volume}{34}, \bibinfo{number}{15} (\bibinfo{year}{2013}), \bibinfo{pages}{2033--2042}.
\newblock


\bibitem[Chen et~al\mbox{.}(2021)]%
        {chen2021dynanet}
\bibfield{author}{\bibinfo{person}{Changhao Chen}, \bibinfo{person}{Chris~Xiaoxuan Lu}, \bibinfo{person}{Bing Wang}, \bibinfo{person}{Niki Trigoni}, {and} \bibinfo{person}{Andrew Markham}.} \bibinfo{year}{2021}\natexlab{}.
\newblock \showarticletitle{DynaNet: Neural Kalman dynamical model for motion estimation and prediction}.
\newblock \bibinfo{journal}{\emph{IEEE Transactions on Neural Networks and Learning Systems}} \bibinfo{volume}{32}, \bibinfo{number}{12} (\bibinfo{year}{2021}), \bibinfo{pages}{5479--5491}.
\newblock


\bibitem[Chen et~al\mbox{.}(2020)]%
        {chen2020survey}
\bibfield{author}{\bibinfo{person}{Changhao Chen}, \bibinfo{person}{Bing Wang}, \bibinfo{person}{Chris~Xiaoxuan Lu}, \bibinfo{person}{Niki Trigoni}, {and} \bibinfo{person}{Andrew Markham}.} \bibinfo{year}{2020}\natexlab{}.
\newblock \showarticletitle{A survey on deep learning for localization and mapping: Towards the age of spatial machine intelligence}.
\newblock \bibinfo{journal}{\emph{arXiv preprint arXiv:2006.12567}} (\bibinfo{year}{2020}).
\newblock


\bibitem[Chen et~al\mbox{.}(2018)]%
        {oxiod-dataset}
\bibfield{author}{\bibinfo{person}{Changhao Chen}, \bibinfo{person}{Peijun Zhao}, \bibinfo{person}{Chris~Xiaoxuan Lu}, \bibinfo{person}{Wei Wang}, \bibinfo{person}{Andrew Markham}, {and} \bibinfo{person}{Niki Trigoni}.} \bibinfo{year}{2018}\natexlab{}.
\newblock \showarticletitle{Ox{I}{O}{D}: {T}he {D}ataset for {D}eep {I}nertial {O}dometry}.
\newblock \bibinfo{journal}{\emph{arXiv preprint arXiv:1809.07491}} (\bibinfo{year}{2018}).
\newblock


\bibitem[Chen et~al\mbox{.}(2022)]%
        {traditional-semantic-survey}
\bibfield{author}{\bibinfo{person}{Weifeng Chen}, \bibinfo{person}{Guangtao Shang}, \bibinfo{person}{Aihong Ji}, \bibinfo{person}{Chengjun Zhou}, \bibinfo{person}{Xiyang Wang}, \bibinfo{person}{Chonghui Xu}, \bibinfo{person}{Zhenxiong Li}, {and} \bibinfo{person}{Kai Hu}.} \bibinfo{year}{2022}\natexlab{}.
\newblock \showarticletitle{An Overview on Visual SLAM: From Tradition to Semantic}.
\newblock \bibinfo{journal}{\emph{Remote Sensing}} \bibinfo{volume}{14}, \bibinfo{number}{13} (\bibinfo{year}{2022}).
\newblock


\bibitem[Chong et~al\mbox{.}(2015)]%
        {chong2015sensor}
\bibfield{author}{\bibinfo{person}{TJ Chong}, \bibinfo{person}{XJ Tang}, \bibinfo{person}{CH Leng}, \bibinfo{person}{Mohan Yogeswaran}, \bibinfo{person}{OE Ng}, {and} \bibinfo{person}{YZ Chong}.} \bibinfo{year}{2015}\natexlab{}.
\newblock \showarticletitle{Sensor technologies and simultaneous localization and mapping (SLAM)}.
\newblock \bibinfo{journal}{\emph{Procedia Computer Science}}  \bibinfo{volume}{76} (\bibinfo{year}{2015}), \bibinfo{pages}{174--179}.
\newblock


\bibitem[Cort{\'e}s et~al\mbox{.}(2018)]%
        {advio-dataset}
\bibfield{author}{\bibinfo{person}{Santiago Cort{\'e}s}, \bibinfo{person}{Arno Solin}, \bibinfo{person}{Esa Rahtu}, {and} \bibinfo{person}{Juho Kannala}.} \bibinfo{year}{2018}\natexlab{}.
\newblock \showarticletitle{A{D}{V}{I}{O}: {A}n {A}uthentic {D}ataset for {V}isual-{I}nertial {O}dometry}. In \bibinfo{booktitle}{\emph{Proceedings of the European Conference on Computer Vision (ECCV)}}.
\newblock


\bibitem[De~la Torre et~al\mbox{.}(2009)]%
        {de2009guide}
\bibfield{author}{\bibinfo{person}{Fernando De~la Torre}, \bibinfo{person}{Jessica Hodgins}, \bibinfo{person}{Adam Bargteil}, \bibinfo{person}{Xavier Martin}, \bibinfo{person}{Justin Macey}, \bibinfo{person}{Alex Collado}, {and} \bibinfo{person}{Pep Beltran}.} \bibinfo{year}{2009}\natexlab{}.
\newblock \showarticletitle{Guide to the carnegie mellon university multimodal activity (cmu-mmac) database}.
\newblock  (\bibinfo{year}{2009}).
\newblock


\bibitem[Diaz et~al\mbox{.}(2017)]%
        {depth-perception-diaz2017designing}
\bibfield{author}{\bibinfo{person}{Catherine Diaz}, \bibinfo{person}{Michael Walker}, \bibinfo{person}{Danielle~Albers Szafir}, {and} \bibinfo{person}{Daniel Szafir}.} \bibinfo{year}{2017}\natexlab{}.
\newblock \showarticletitle{Designing for depth perceptions in augmented reality}. In \bibinfo{booktitle}{\emph{IEEE international symposium on mixed and augmented reality (ISMAR)}}.
\newblock


\bibitem[Dressel and Kochenderfer(2019)]%
        {drones}
\bibfield{author}{\bibinfo{person}{Louis Dressel} {and} \bibinfo{person}{Mykel~J. Kochenderfer}.} \bibinfo{year}{2019}\natexlab{}.
\newblock \showarticletitle{Hunting Drones with Other Drones: Tracking a Moving Radio Target}.
\newblock \bibinfo{journal}{\emph{IEEE International Conference on Robotics and Automation (ICRA)}}.
\newblock


\bibitem[Durrant-Whyte and Bailey(2006)]%
        {durrant2006simultaneous-part-1-survey}
\bibfield{author}{\bibinfo{person}{Hugh Durrant-Whyte} {and} \bibinfo{person}{Tim Bailey}.} \bibinfo{year}{2006}\natexlab{}.
\newblock \showarticletitle{Simultaneous localization and mapping: part I}.
\newblock \bibinfo{journal}{\emph{IEEE robotics \& automation magazine}} \bibinfo{volume}{13}, \bibinfo{number}{2} (\bibinfo{year}{2006}), \bibinfo{pages}{99--110}.
\newblock


\bibitem[Erickson et~al\mbox{.}(2020)]%
        {lighting-ost}
\bibfield{author}{\bibinfo{person}{Austin Erickson}, \bibinfo{person}{Kangsoo Kim}, \bibinfo{person}{Gerd Bruder}, {and} \bibinfo{person}{Gregory~F. Welch}.} \bibinfo{year}{2020}\natexlab{}.
\newblock \showarticletitle{Exploring the Limitations of Environment Lighting on Optical See-Through Head-Mounted Displays}. In \bibinfo{booktitle}{\emph{Symposium on Spatial User Interaction}}.
\newblock


\bibitem[Fan et~al\mbox{.}(2024)]%
        {fan2024cueing-trajectory-safety-prediction}
\bibfield{author}{\bibinfo{person}{Chao Fan}, \bibinfo{person}{Weike Ding}, \bibinfo{person}{Kun Qian}, \bibinfo{person}{Hao Tan}, {and} \bibinfo{person}{Zihan Li}.} \bibinfo{year}{2024}\natexlab{}.
\newblock \showarticletitle{Cueing Flight Object Trajectory and Safety Prediction Based on SLAM Technology}.
\newblock \bibinfo{journal}{\emph{Journal of Theory and Practice of Engineering Science}} \bibinfo{volume}{4}, \bibinfo{number}{05} (\bibinfo{year}{2024}), \bibinfo{pages}{1--8}.
\newblock


\bibitem[Fu et~al\mbox{.}(2021)]%
        {activityaware-fu2021hawatcher}
\bibfield{author}{\bibinfo{person}{Chenglong Fu}, \bibinfo{person}{Qiang Zeng}, {and} \bibinfo{person}{Xiaojiang Du}.} \bibinfo{year}{2021}\natexlab{}.
\newblock \showarticletitle{Hawatcher: Semantics-aware anomaly detection for appified smart homes}. In \bibinfo{booktitle}{\emph{USENIX Security)}}.
\newblock


\bibitem[Fua and Lepetit(2007)]%
        {fua2007vision-mr-tracking}
\bibfield{author}{\bibinfo{person}{Pascal Fua} {and} \bibinfo{person}{Vincent Lepetit}.} \bibinfo{year}{2007}\natexlab{}.
\newblock \showarticletitle{Vision based 3D tracking and pose estimation for mixed reality}.
\newblock In \bibinfo{booktitle}{\emph{Emerging technologies of augmented reality: Interfaces and design}}. \bibinfo{publisher}{IGI Global}, \bibinfo{pages}{1--22}.
\newblock


\bibitem[Fuentes-Pacheco et~al\mbox{.}(2015)]%
        {fuentes2015visual-survey}
\bibfield{author}{\bibinfo{person}{Jorge Fuentes-Pacheco}, \bibinfo{person}{Jos{\'e} Ruiz-Ascencio}, {and} \bibinfo{person}{Juan~Manuel Rend{\'o}n-Mancha}.} \bibinfo{year}{2015}\natexlab{}.
\newblock \showarticletitle{Visual simultaneous localization and mapping: a survey}.
\newblock \bibinfo{journal}{\emph{Artificial intelligence review}}  \bibinfo{volume}{43} (\bibinfo{year}{2015}), \bibinfo{pages}{55--81}.
\newblock


\bibitem[Garigipati et~al\mbox{.}(2022)]%
        {lidar-slam-survey-garigipati2022evaluation}
\bibfield{author}{\bibinfo{person}{Bharath Garigipati}, \bibinfo{person}{Nataliya Strokina}, {and} \bibinfo{person}{Reza Ghabcheloo}.} \bibinfo{year}{2022}\natexlab{}.
\newblock \showarticletitle{Evaluation and comparison of eight popular Lidar and Visual SLAM algorithms}. In \bibinfo{booktitle}{\emph{2022 25th International Conference on Information Fusion (FUSION)}}. IEEE, \bibinfo{pages}{1--8}.
\newblock


\bibitem[Geiger et~al\mbox{.}(2013)]%
        {kitti-geiger2013vision}
\bibfield{author}{\bibinfo{person}{Andreas Geiger}, \bibinfo{person}{Philip Lenz}, \bibinfo{person}{Christoph Stiller}, {and} \bibinfo{person}{Raquel Urtasun}.} \bibinfo{year}{2013}\natexlab{}.
\newblock \showarticletitle{Vision meets robotics: The kitti dataset}.
\newblock \bibinfo{journal}{\emph{The International Journal of Robotics Research}} \bibinfo{volume}{32}, \bibinfo{number}{11} (\bibinfo{year}{2013}), \bibinfo{pages}{1231--1237}.
\newblock


\bibitem[Gonzalez-Franco et~al\mbox{.}(2017)]%
        {gonzalez2017immersive}
\bibfield{author}{\bibinfo{person}{Mar Gonzalez-Franco}, \bibinfo{person}{Rodrigo Pizarro}, \bibinfo{person}{Julio Cermeron}, \bibinfo{person}{Katie Li}, \bibinfo{person}{Jacob Thorn}, \bibinfo{person}{Windo Hutabarat}, \bibinfo{person}{Ashutosh Tiwari}, {and} \bibinfo{person}{Pablo Bermell-Garcia}.} \bibinfo{year}{2017}\natexlab{}.
\newblock \showarticletitle{Immersive mixed reality for manufacturing training}.
\newblock \bibinfo{journal}{\emph{Frontiers in Robotics and AI}}  \bibinfo{volume}{4} (\bibinfo{year}{2017}), \bibinfo{pages}{3}.
\newblock


\bibitem[Green and Bavelier(2003)]%
        {gamers-enhanced-cognitive-ablities-green2003action}
\bibfield{author}{\bibinfo{person}{C~Shawn Green} {and} \bibinfo{person}{Daphne Bavelier}.} \bibinfo{year}{2003}\natexlab{}.
\newblock \showarticletitle{Action video game modifies visual selective attention}.
\newblock \bibinfo{journal}{\emph{Nature}} \bibinfo{volume}{423}, \bibinfo{number}{6939} (\bibinfo{year}{2003}), \bibinfo{pages}{534--537}.
\newblock


\bibitem[Grisetti et~al\mbox{.}(2010)]%
        {grisetti2010tutorial-slam}
\bibfield{author}{\bibinfo{person}{Giorgio Grisetti}, \bibinfo{person}{Rainer K{\"u}mmerle}, \bibinfo{person}{Cyrill Stachniss}, {and} \bibinfo{person}{Wolfram Burgard}.} \bibinfo{year}{2010}\natexlab{}.
\newblock \showarticletitle{A tutorial on graph-based SLAM}.
\newblock \bibinfo{journal}{\emph{IEEE Intelligent Transportation Systems Magazine}} \bibinfo{volume}{2}, \bibinfo{number}{4} (\bibinfo{year}{2010}), \bibinfo{pages}{31--43}.
\newblock


\bibitem[Guo et~al\mbox{.}(2022)]%
        {guo2022lidar-slam-featrue-and-kitti-scene-info}
\bibfield{author}{\bibinfo{person}{Shiyi Guo}, \bibinfo{person}{Zheng Rong}, \bibinfo{person}{Shuo Wang}, {and} \bibinfo{person}{Yihong Wu}.} \bibinfo{year}{2022}\natexlab{}.
\newblock \showarticletitle{A LiDAR SLAM with PCA-based feature extraction and two-stage matching}.
\newblock \bibinfo{journal}{\emph{IEEE Transactions on Instrumentation and Measurement}}  \bibinfo{volume}{71} (\bibinfo{year}{2022}).
\newblock


\bibitem[Gupta and Fernando(2022)]%
        {drones6040085}
\bibfield{author}{\bibinfo{person}{Abhishek Gupta} {and} \bibinfo{person}{Xavier Fernando}.} \bibinfo{year}{2022}\natexlab{}.
\newblock \showarticletitle{Simultaneous Localization and Mapping (SLAM) and Data Fusion in Unmanned Aerial Vehicles: Recent Advances and Challenges}.
\newblock \bibinfo{journal}{\emph{Drones}} \bibinfo{volume}{6}, \bibinfo{number}{4} (\bibinfo{year}{2022}).
\newblock


\bibitem[Gurturk et~al\mbox{.}(2021)]%
        {ytu-dataset}
\bibfield{author}{\bibinfo{person}{Mert Gurturk}, \bibinfo{person}{Abdullah Yusefi}, \bibinfo{person}{Muhammet~Fatih Aslan}, \bibinfo{person}{Metin Soycan}, \bibinfo{person}{Akif Durdu}, {and} \bibinfo{person}{Andrea Masiero}.} \bibinfo{year}{2021}\natexlab{}.
\newblock \showarticletitle{The {Y}{T}{U} {D}ataset and {R}ecurrent {N}eural {N}etwork based {V}isual-inertial {O}dometry}.
\newblock \bibinfo{journal}{\emph{Measurement}} (\bibinfo{year}{2021}).
\newblock


\bibitem[Halmetschlager-Funek et~al\mbox{.}(2018)]%
        {depth-camera-issues-halmetschlager2018empirical}
\bibfield{author}{\bibinfo{person}{Georg Halmetschlager-Funek}, \bibinfo{person}{Markus Suchi}, \bibinfo{person}{Martin Kampel}, {and} \bibinfo{person}{Markus Vincze}.} \bibinfo{year}{2018}\natexlab{}.
\newblock \showarticletitle{An empirical evaluation of ten depth cameras: Bias, precision, lateral noise, different lighting conditions and materials, and multiple sensor setups in indoor environments}.
\newblock \bibinfo{journal}{\emph{IEEE Robotics \& Automation Magazine}} \bibinfo{volume}{26}, \bibinfo{number}{1} (\bibinfo{year}{2018}), \bibinfo{pages}{67--77}.
\newblock


\bibitem[Handa et~al\mbox{.}(2014)]%
        {handa2014benchmark}
\bibfield{author}{\bibinfo{person}{Ankur Handa}, \bibinfo{person}{Thomas Whelan}, \bibinfo{person}{John McDonald}, {and} \bibinfo{person}{Andrew~J Davison}.} \bibinfo{year}{2014}\natexlab{}.
\newblock \showarticletitle{A benchmark for RGB-D visual odometry, 3D reconstruction and SLAM}. In \bibinfo{booktitle}{\emph{2014 IEEE international conference on Robotics and automation (ICRA)}}. IEEE, \bibinfo{pages}{1524--1531}.
\newblock


\bibitem[Hu et~al\mbox{.}(2024)]%
        {huapple-vs-meta-spatial-tracking-XR-maria}
\bibfield{author}{\bibinfo{person}{Tianyi Hu}, \bibinfo{person}{Fan Yang}, {and} \bibinfo{person}{Tim Scargill~Maria Gorlatova}.} \bibinfo{year}{2024}\natexlab{}.
\newblock \showarticletitle{Apple vs. Meta: A Comparative Study on Spatial Tracking in SOTA XR Headsets}. In \bibinfo{booktitle}{\emph{2nd ACM Workshop on Mobile Immersive Computing, Networking, and Systems (ImmerCom)}}.
\newblock


\bibitem[Huang and Dissanayake(2016)]%
        {huang2016critique-survey}
\bibfield{author}{\bibinfo{person}{Shoudong Huang} {and} \bibinfo{person}{Gamini Dissanayake}.} \bibinfo{year}{2016}\natexlab{}.
\newblock \showarticletitle{A critique of current developments in simultaneous localization and mapping}.
\newblock \bibinfo{journal}{\emph{International Journal of Advanced Robotic Systems}} \bibinfo{volume}{13}, \bibinfo{number}{5} (\bibinfo{year}{2016}), \bibinfo{pages}{1729881416669482}.
\newblock


\bibitem[H{\"u}bner et~al\mbox{.}(2020)]%
        {hubner2020evaluation}
\bibfield{author}{\bibinfo{person}{Patrick H{\"u}bner}, \bibinfo{person}{Kate Clintworth}, \bibinfo{person}{Qingyi Liu}, \bibinfo{person}{Martin Weinmann}, {and} \bibinfo{person}{Sven Wursthorn}.} \bibinfo{year}{2020}\natexlab{}.
\newblock \showarticletitle{Evaluation of HoloLens tracking and depth sensing for indoor mapping applications}.
\newblock \bibinfo{journal}{\emph{Sensors}} \bibinfo{volume}{20}, \bibinfo{number}{4} (\bibinfo{year}{2020}), \bibinfo{pages}{1021}.
\newblock


\bibitem[Inc.(2020)]%
        {hololens2}
\bibfield{author}{\bibinfo{person}{Microsoft Inc.}} \bibinfo{year}{2020}\natexlab{}.
\newblock \showarticletitle{{Hololens 2}}.
\newblock \bibinfo{howpublished}{\url{https://www.microsoft.com/en-us/hololens/hardware}}.
\newblock  (\bibinfo{year}{2020}).
\newblock
\newblock
\shownote{[Online; accessed September 2024]}.


\bibitem[Jeong et~al\mbox{.}(2021)]%
        {deep-vo-survey-jeong2021comparison}
\bibfield{author}{\bibinfo{person}{Eunju Jeong}, \bibinfo{person}{Jaun Lee}, {and} \bibinfo{person}{Pyojin Kim}.} \bibinfo{year}{2021}\natexlab{}.
\newblock \showarticletitle{A Comparison of Deep Learning-Based Monocular Visual Odometry Algorithms}. In \bibinfo{booktitle}{\emph{Asia-Pacific International Symposium on Aerospace Technology}}. Springer, \bibinfo{pages}{923--934}.
\newblock


\bibitem[Jinyu et~al\mbox{.}(2019)]%
        {jinyu2019survey}
\bibfield{author}{\bibinfo{person}{Li Jinyu}, \bibinfo{person}{Yang Bangbang}, \bibinfo{person}{Chen Danpeng}, \bibinfo{person}{Wang Nan}, \bibinfo{person}{Zhang Guofeng}, {and} \bibinfo{person}{Bao Hujun}.} \bibinfo{year}{2019}\natexlab{}.
\newblock \showarticletitle{Survey and evaluation of monocular visual-inertial SLAM algorithms for augmented reality}.
\newblock \bibinfo{journal}{\emph{Virtual Reality \& Intelligent Hardware}} \bibinfo{volume}{1}, \bibinfo{number}{4} (\bibinfo{year}{2019}), \bibinfo{pages}{386--410}.
\newblock


\bibitem[Khandelwal and Wickstr{\"o}m(2017)]%
        {khandelwal2017evaluation}
\bibfield{author}{\bibinfo{person}{Siddhartha Khandelwal} {and} \bibinfo{person}{Nicholas Wickstr{\"o}m}.} \bibinfo{year}{2017}\natexlab{}.
\newblock \showarticletitle{Evaluation of the performance of accelerometer-based gait event detection algorithms in different real-world scenarios using the MAREA gait database}.
\newblock \bibinfo{journal}{\emph{Gait \& posture}}  \bibinfo{volume}{51} (\bibinfo{year}{2017}), \bibinfo{pages}{84--90}.
\newblock


\bibitem[Klein and Murray(2007)]%
        {SLAMAR-klein2007parallel}
\bibfield{author}{\bibinfo{person}{Georg Klein} {and} \bibinfo{person}{David Murray}.} \bibinfo{year}{2007}\natexlab{}.
\newblock \showarticletitle{Parallel tracking and mapping for small AR workspaces}. In \bibinfo{booktitle}{\emph{6th IEEE and ACM International Symposium on Mixed and Augmented Reality}}.
\newblock


\bibitem[Klein(2019)]%
        {klein2019smartphone}
\bibfield{author}{\bibinfo{person}{Itzik Klein}.} \bibinfo{year}{2019}\natexlab{}.
\newblock \showarticletitle{Smartphone location recognition: A deep learning-based approach}.
\newblock \bibinfo{journal}{\emph{Sensors}} \bibinfo{volume}{20}, \bibinfo{number}{1} (\bibinfo{year}{2019}), \bibinfo{pages}{214}.
\newblock


\bibitem[{Krishna Murthy} et~al\mbox{.}(2020)]%
        {gradslam}
\bibfield{author}{\bibinfo{person}{Jatavallabhula {Krishna Murthy}}, \bibinfo{person}{Soroush Saryazdi}, \bibinfo{person}{Ganesh Iyer}, {and} \bibinfo{person}{Liam Paull}.} \bibinfo{year}{2020}\natexlab{}.
\newblock \showarticletitle{gradSLAM: Dense SLAM meets automatic differentiation}. In \bibinfo{booktitle}{\emph{arXiv preprint arXiv:1910.10672}}.
\newblock


\bibitem[Latif et~al\mbox{.}(2013)]%
        {latif2013robustloopclousure}
\bibfield{author}{\bibinfo{person}{Yasir Latif}, \bibinfo{person}{Cesar Cadena}, {and} \bibinfo{person}{Jos{\'e} Neira}.} \bibinfo{year}{2013}\natexlab{}.
\newblock \showarticletitle{Robust loop closing over time}. In \bibinfo{booktitle}{\emph{Proc. Robotics: Science Systems}}. \bibinfo{pages}{233--240}.
\newblock


\bibitem[{Lebeck} et~al\mbox{.}(2017)]%
        {output_reality}
\bibfield{author}{\bibinfo{person}{K. {Lebeck}}, \bibinfo{person}{K. {Ruth}}, \bibinfo{person}{T. {Kohno}}, {and} \bibinfo{person}{F. {Roesner}}.} \bibinfo{year}{2017}\natexlab{}.
\newblock \showarticletitle{Securing Augmented Reality Output}.
\newblock \bibinfo{journal}{\emph{IEEE S\&P}}.
\newblock


\bibitem[Li et~al\mbox{.}(2022a)]%
        {li2022timing-slam-survey}
\bibfield{author}{\bibinfo{person}{Ao Li}, \bibinfo{person}{Han Liu}, \bibinfo{person}{Jinwen Wang}, {and} \bibinfo{person}{Ning Zhang}.} \bibinfo{year}{2022}\natexlab{a}.
\newblock \showarticletitle{From timing variations to performance degradation: Understanding and mitigating the impact of software execution timing in slam}. In \bibinfo{booktitle}{\emph{IEEE/RSJ International Conference on Intelligent Robots and Systems (IROS)}}.
\newblock


\bibitem[Li and Mourikis(2013)]%
        {spatial-misalogment}
\bibfield{author}{\bibinfo{person}{Mingyang Li} {and} \bibinfo{person}{Anastasios Mourikis}.} \bibinfo{year}{2013}\natexlab{}.
\newblock \showarticletitle{High-precision, {C}onsistent {EKF}-based {V}isual–{I}nertial {O}dometry}.
\newblock \bibinfo{journal}{\emph{Journal of Robotics Research}} (\bibinfo{year}{2013}).
\newblock


\bibitem[Li et~al\mbox{.}(2018)]%
        {li2018ongoing-evolution-survey}
\bibfield{author}{\bibinfo{person}{Ruihao Li}, \bibinfo{person}{Sen Wang}, {and} \bibinfo{person}{Dongbing Gu}.} \bibinfo{year}{2018}\natexlab{}.
\newblock \showarticletitle{Ongoing evolution of visual SLAM from geometry to deep learning: Challenges and opportunities}.
\newblock \bibinfo{journal}{\emph{Cognitive Computation}}  \bibinfo{volume}{10} (\bibinfo{year}{2018}), \bibinfo{pages}{875--889}.
\newblock


\bibitem[Li et~al\mbox{.}(2020)]%
        {li2020self-deep-vo}
\bibfield{author}{\bibinfo{person}{Shunkai Li}, \bibinfo{person}{Xin Wang}, \bibinfo{person}{Yingdian Cao}, \bibinfo{person}{Fei Xue}, \bibinfo{person}{Zike Yan}, {and} \bibinfo{person}{Hongbin Zha}.} \bibinfo{year}{2020}\natexlab{}.
\newblock \showarticletitle{Self-supervised deep visual odometry with online adaptation}. In \bibinfo{booktitle}{\emph{Proceedings of the IEEE/CVF Conference on Computer Vision and Pattern Recognition}}. \bibinfo{pages}{6339--6348}.
\newblock


\bibitem[Li et~al\mbox{.}(2022b)]%
        {li2022overview}
\bibfield{author}{\bibinfo{person}{Shaopeng Li}, \bibinfo{person}{Daqiao Zhang}, \bibinfo{person}{Yong Xian}, \bibinfo{person}{Bangjie Li}, \bibinfo{person}{Tao Zhang}, {and} \bibinfo{person}{Chengliang Zhong}.} \bibinfo{year}{2022}\natexlab{b}.
\newblock \showarticletitle{Overview of deep learning application on visual SLAM}.
\newblock \bibinfo{journal}{\emph{Displays}} (\bibinfo{year}{2022}).
\newblock


\bibitem[Li et~al\mbox{.}(2021)]%
        {structural-irrelagularite-li2021rgb}
\bibfield{author}{\bibinfo{person}{Yanyan Li}, \bibinfo{person}{Raza Yunus}, \bibinfo{person}{Nikolas Brasch}, \bibinfo{person}{Nassir Navab}, {and} \bibinfo{person}{Federico Tombari}.} \bibinfo{year}{2021}\natexlab{}.
\newblock \showarticletitle{RGB-D SLAM with structural regularities}. In \bibinfo{booktitle}{\emph{IEEE International Conference on Robotics and Automation (ICRA)}}.
\newblock


\bibitem[Liu et~al\mbox{.}(2021)]%
        {liu2021simultaneous-datasets-survey}
\bibfield{author}{\bibinfo{person}{Yuanzhi Liu}, \bibinfo{person}{Yujia Fu}, \bibinfo{person}{Fengdong Chen}, \bibinfo{person}{Bart Goossens}, \bibinfo{person}{Wei Tao}, {and} \bibinfo{person}{Hui Zhao}.} \bibinfo{year}{2021}\natexlab{}.
\newblock \showarticletitle{Simultaneous localization and mapping related datasets: A comprehensive survey}.
\newblock \bibinfo{journal}{\emph{arXiv preprint arXiv:2102.04036}} (\bibinfo{year}{2021}).
\newblock


\bibitem[Liu et~al\mbox{.}(2024)]%
        {liu2024adaptive-voo}
\bibfield{author}{\bibinfo{person}{Ziming Liu}, \bibinfo{person}{Ezio Malis}, {and} \bibinfo{person}{Philippe Martinet}.} \bibinfo{year}{2024}\natexlab{}.
\newblock \showarticletitle{Adaptive Learning for Hybrid Visual Odometry}.
\newblock \bibinfo{journal}{\emph{IEEE Robotics and Automation Letters}} (\bibinfo{year}{2024}).
\newblock


\bibitem[Lok(2004)]%
        {merging-real-vr-lok2004toward}
\bibfield{author}{\bibinfo{person}{Benjamin~C Lok}.} \bibinfo{year}{2004}\natexlab{}.
\newblock \showarticletitle{Toward the merging of real and virtual spaces}.
\newblock \bibinfo{journal}{\emph{Commun. ACM}} \bibinfo{volume}{47}, \bibinfo{number}{8} (\bibinfo{year}{2004}).
\newblock


\bibitem[Macario~Barros et~al\mbox{.}(2022)]%
        {macario2022comprehensive-survey}
\bibfield{author}{\bibinfo{person}{Andr{\'e}a Macario~Barros}, \bibinfo{person}{Maugan Michel}, \bibinfo{person}{Yoann Moline}, \bibinfo{person}{Gwenol{\'e} Corre}, {and} \bibinfo{person}{Fr{\'e}d{\'e}rick Carrel}.} \bibinfo{year}{2022}\natexlab{}.
\newblock \showarticletitle{A comprehensive survey of visual slam algorithms}.
\newblock \bibinfo{journal}{\emph{Robotics}} \bibinfo{volume}{11}, \bibinfo{number}{1} (\bibinfo{year}{2022}), \bibinfo{pages}{24}.
\newblock


\bibitem[Maddern et~al\mbox{.}(2017)]%
        {maddern20171}
\bibfield{author}{\bibinfo{person}{Will Maddern}, \bibinfo{person}{Geoffrey Pascoe}, \bibinfo{person}{Chris Linegar}, {and} \bibinfo{person}{Paul Newman}.} \bibinfo{year}{2017}\natexlab{}.
\newblock \showarticletitle{1 year, 1000 km: The Oxford RobotCar dataset}.
\newblock \bibinfo{journal}{\emph{The International Journal of Robotics Research}} \bibinfo{volume}{36}, \bibinfo{number}{1} (\bibinfo{year}{2017}), \bibinfo{pages}{3--15}.
\newblock


\bibitem[Milz et~al\mbox{.}(2018)]%
        {automated-driving-tracking-survey-cvpr-w-milz2018visual}
\bibfield{author}{\bibinfo{person}{Stefan Milz}, \bibinfo{person}{Georg Arbeiter}, \bibinfo{person}{Christian Witt}, \bibinfo{person}{Bassam Abdallah}, {and} \bibinfo{person}{Senthil Yogamani}.} \bibinfo{year}{2018}\natexlab{}.
\newblock \showarticletitle{Visual slam for automated driving: Exploring the applications of deep learning}. In \bibinfo{booktitle}{\emph{Proceedings of the IEEE Conference on Computer Vision and Pattern Recognition Workshops}}. \bibinfo{pages}{247--257}.
\newblock


\bibitem[Mokssit et~al\mbox{.}(2023)]%
        {mokssit2023deep}
\bibfield{author}{\bibinfo{person}{Saad Mokssit}, \bibinfo{person}{Daniel~Bonilla Licea}, \bibinfo{person}{Bassma Guermah}, {and} \bibinfo{person}{Mounir Ghogho}.} \bibinfo{year}{2023}\natexlab{}.
\newblock \showarticletitle{Deep Learning Techniques for Visual SLAM: A Survey}.
\newblock \bibinfo{journal}{\emph{IEEE Access}}  \bibinfo{volume}{11} (\bibinfo{year}{2023}), \bibinfo{pages}{20026--20050}.
\newblock


\bibitem[Muravyev and Yakovlev(2022)]%
        {large-indoor-survey-muravyev2022evaluation}
\bibfield{author}{\bibinfo{person}{Kirill Muravyev} {and} \bibinfo{person}{Konstantin Yakovlev}.} \bibinfo{year}{2022}\natexlab{}.
\newblock \showarticletitle{Evaluation of RGB-D SLAM in Large Indoor Environments}. In \bibinfo{booktitle}{\emph{International Conference on Interactive Collaborative Robotics}}. Springer, \bibinfo{pages}{93--104}.
\newblock


\bibitem[Nardi et~al\mbox{.}(2015)]%
        {nardi2015introducing-SLAMBENCh}
\bibfield{author}{\bibinfo{person}{Luigi Nardi}, \bibinfo{person}{Bruno Bodin}, \bibinfo{person}{M~Zeeshan Zia}, \bibinfo{person}{John Mawer}, \bibinfo{person}{Andy Nisbet}, \bibinfo{person}{Paul~HJ Kelly}, \bibinfo{person}{Andrew~J Davison}, \bibinfo{person}{Mikel Luj{\'a}n}, \bibinfo{person}{Michael~FP O'Boyle}, \bibinfo{person}{Graham Riley}, {et~al\mbox{.}}} \bibinfo{year}{2015}\natexlab{}.
\newblock \showarticletitle{Introducing SLAMBench, a performance and accuracy benchmarking methodology for SLAM}. In \bibinfo{booktitle}{\emph{IEEE international conference on robotics and automation (ICRA)}}.
\newblock


\bibitem[Ngo et~al\mbox{.}(2022)]%
        {cps-work-luis-human-cyber-safety-crtical}
\bibfield{author}{\bibinfo{person}{Steven Ngo}, \bibinfo{person}{Dave DeAngelis}, {and} \bibinfo{person}{Luis Garcia}.} \bibinfo{year}{2022}\natexlab{}.
\newblock \showarticletitle{Modeling Human-Cyber Interactions in Safety-Critical Cyber-Physical/Industrial Control Systems}. In \bibinfo{booktitle}{\emph{IEEE 19th International Conference on Mobile Ad Hoc and Smart Systems (MASS)}}.
\newblock


\bibitem[Niemann et~al\mbox{.}(2020)]%
        {niemann2020lara}
\bibfield{author}{\bibinfo{person}{Friedrich Niemann}, \bibinfo{person}{Christopher Reining}, \bibinfo{person}{Fernando Moya~Rueda}, \bibinfo{person}{Nilah~Ravi Nair}, \bibinfo{person}{Janine~Anika Steffens}, \bibinfo{person}{Gernot~A Fink}, {and} \bibinfo{person}{Michael Ten~Hompel}.} \bibinfo{year}{2020}\natexlab{}.
\newblock \showarticletitle{Lara: Creating a dataset for human activity recognition in logistics using semantic attributes}.
\newblock \bibinfo{journal}{\emph{Sensors}} \bibinfo{volume}{20}, \bibinfo{number}{15} (\bibinfo{year}{2020}), \bibinfo{pages}{4083}.
\newblock


\bibitem[Nwankwo and Rueckert(2023)]%
        {why-slam-fail}
\bibfield{author}{\bibinfo{person}{Linus Nwankwo} {and} \bibinfo{person}{Elmar Rueckert}.} \bibinfo{year}{2023}\natexlab{}.
\newblock \showarticletitle{Understanding Why SLAM Algorithms Fail in Modern Indoor Environments}. In \bibinfo{booktitle}{\emph{Advances in Service and Industrial Robotics}}, \bibfield{editor}{\bibinfo{person}{Tadej Petri{\v{c}}}, \bibinfo{person}{Ale{\v{s}} Ude}, {and} \bibinfo{person}{Leon {\v{Z}}lajpah}} (Eds.). \bibinfo{publisher}{Springer Nature Switzerland}, \bibinfo{address}{Cham}, \bibinfo{pages}{186--194}.
\newblock


\bibitem[Petit et~al\mbox{.}(2015)]%
        {occlusion-attack-petit2015remote}
\bibfield{author}{\bibinfo{person}{Jonathan Petit}, \bibinfo{person}{Bas Stottelaar}, \bibinfo{person}{Michael Feiri}, {and} \bibinfo{person}{Frank Kargl}.} \bibinfo{year}{2015}\natexlab{}.
\newblock \showarticletitle{Remote Attacks on Automated Vehicles Sensors: Experiments on Camera and Lidar}.
\newblock \bibinfo{journal}{\emph{Black Hat Europe}}  \bibinfo{volume}{11} (\bibinfo{year}{2015}).
\newblock


\bibitem[Prokhorov et~al\mbox{.}(2019)]%
        {survey-robustness-prokhorov2019measuring}
\bibfield{author}{\bibinfo{person}{David Prokhorov}, \bibinfo{person}{Dmitry Zhukov}, \bibinfo{person}{Olga Barinova}, \bibinfo{person}{Konushin Anton}, {and} \bibinfo{person}{Anna Vorontsova}.} \bibinfo{year}{2019}\natexlab{}.
\newblock \showarticletitle{Measuring robustness of Visual SLAM}. In \bibinfo{booktitle}{\emph{2019 16th International Conference on Machine Vision Applications (MVA)}}. IEEE, \bibinfo{pages}{1--6}.
\newblock


\bibitem[Pupilli and Calway(2006)]%
        {pupilli2006real-erratic-locomotion}
\bibfield{author}{\bibinfo{person}{Mark Pupilli} {and} \bibinfo{person}{Andrew Calway}.} \bibinfo{year}{2006}\natexlab{}.
\newblock \showarticletitle{Real-time visual slam with resilience to erratic motion}. In \bibinfo{booktitle}{\emph{2006 IEEE Computer Society Conference on Computer Vision and Pattern Recognition (CVPR'06)}}, Vol.~\bibinfo{volume}{1}. IEEE, \bibinfo{pages}{1244--1249}.
\newblock


\bibitem[Qin et~al\mbox{.}(2018)]%
        {vins-mono-qin2017vins}
\bibfield{author}{\bibinfo{person}{Tong Qin}, \bibinfo{person}{Peiliang Li}, {and} \bibinfo{person}{Shaojie Shen}.} \bibinfo{year}{2018}\natexlab{}.
\newblock \showarticletitle{VINS-Mono: A Robust and Versatile Monocular Visual-Inertial State Estimator}.
\newblock \bibinfo{journal}{\emph{IEEE Transactions on Robotics}} \bibinfo{volume}{34}, \bibinfo{number}{4} (\bibinfo{year}{2018}).
\newblock


\bibitem[Qin and Shen(2018)]%
        {vins-fusion-qin2018online}
\bibfield{author}{\bibinfo{person}{Tong Qin} {and} \bibinfo{person}{Shaojie Shen}.} \bibinfo{year}{2018}\natexlab{}.
\newblock \showarticletitle{Online Temporal Calibration for Monocular Visual-Inertial Systems}. In \bibinfo{booktitle}{\emph{IEEE/RSJ International Conference on Intelligent Robots and Systems (IROS)}}.
\newblock


\bibitem[Rogers et~al\mbox{.}(2020)]%
        {rogers2020test-your-slam}
\bibfield{author}{\bibinfo{person}{John~G Rogers}, \bibinfo{person}{Jason~M Gregory}, \bibinfo{person}{Jonathan Fink}, {and} \bibinfo{person}{Ethan Stump}.} \bibinfo{year}{2020}\natexlab{}.
\newblock \showarticletitle{Test your slam! the subt-tunnel dataset and metric for mapping}. In \bibinfo{booktitle}{\emph{IEEE International Conference on Robotics and Automation (ICRA)}}.
\newblock


\bibitem[Rublee et~al\mbox{.}(2011)]%
        {rublee2011orb}
\bibfield{author}{\bibinfo{person}{Ethan Rublee}, \bibinfo{person}{Vincent Rabaud}, \bibinfo{person}{Kurt Konolige}, {and} \bibinfo{person}{Gary Bradski}.} \bibinfo{year}{2011}\natexlab{}.
\newblock \showarticletitle{ORB: An efficient alternative to SIFT or SURF}. In \bibinfo{booktitle}{\emph{International Conference on Computer Vision}}. \bibinfo{pages}{2564--2571}.
\newblock


\bibitem[Saeedi et~al\mbox{.}(2019)]%
        {saeedi2019characterizing}
\bibfield{author}{\bibinfo{person}{Sajad Saeedi}, \bibinfo{person}{Eduardo~DC Carvalho}, \bibinfo{person}{Wenbin Li}, \bibinfo{person}{Dimos Tzoumanikas}, \bibinfo{person}{Stefan Leutenegger}, \bibinfo{person}{Paul~HJ Kelly}, {and} \bibinfo{person}{Andrew~J Davison}.} \bibinfo{year}{2019}\natexlab{}.
\newblock \showarticletitle{Characterizing visual localization and mapping datasets}. In \bibinfo{booktitle}{\emph{International Conference on Robotics and Automation (ICRA)}}.
\newblock


\bibitem[Saeedi et~al\mbox{.}(2016)]%
        {saeedi2016multiple-robot-survey}
\bibfield{author}{\bibinfo{person}{Sajad Saeedi}, \bibinfo{person}{Michael Trentini}, \bibinfo{person}{Mae Seto}, {and} \bibinfo{person}{Howard Li}.} \bibinfo{year}{2016}\natexlab{}.
\newblock \showarticletitle{Multiple-robot simultaneous localization and mapping: A review}.
\newblock \bibinfo{journal}{\emph{Journal of Field Robotics}} \bibinfo{volume}{33}, \bibinfo{number}{1} (\bibinfo{year}{2016}), \bibinfo{pages}{3--46}.
\newblock


\bibitem[Salas et~al\mbox{.}(2015)]%
        {ATE_RPE_salas2015trajectory}
\bibfield{author}{\bibinfo{person}{Marta Salas}, \bibinfo{person}{Yasir Latif}, \bibinfo{person}{Ian~D Reid}, {and} \bibinfo{person}{J Montiel}.} \bibinfo{year}{2015}\natexlab{}.
\newblock \showarticletitle{Trajectory alignment and evaluation in SLAM: Horns method vs alignment on the manifold}. In \bibinfo{booktitle}{\emph{Robotics: Science and Systems Workshop: The problem of mobile sensors}}. sn, \bibinfo{pages}{1--3}.
\newblock


\bibitem[Saputra et~al\mbox{.}(2018)]%
        {sfm-dynamic-survey-saputra2018visual}
\bibfield{author}{\bibinfo{person}{Muhamad Risqi~U Saputra}, \bibinfo{person}{Andrew Markham}, {and} \bibinfo{person}{Niki Trigoni}.} \bibinfo{year}{2018}\natexlab{}.
\newblock \showarticletitle{Visual SLAM and structure from motion in dynamic environments: A survey}.
\newblock \bibinfo{journal}{\emph{ACM Computing Surveys (CSUR)}} \bibinfo{volume}{51}, \bibinfo{number}{2} (\bibinfo{year}{2018}), \bibinfo{pages}{1--36}.
\newblock


\bibitem[Schops et~al\mbox{.}(2019)]%
        {schops2019badslam}
\bibfield{author}{\bibinfo{person}{Thomas Schops}, \bibinfo{person}{Torsten Sattler}, {and} \bibinfo{person}{Marc Pollefeys}.} \bibinfo{year}{2019}\natexlab{}.
\newblock \showarticletitle{Bad slam: Bundle adjusted direct rgb-d slam}. In \bibinfo{booktitle}{\emph{Proceedings of the IEEE/CVF Conference on Computer Vision and Pattern Recognition}}. \bibinfo{pages}{134--144}.
\newblock


\bibitem[Schubert et~al\mbox{.}(2018)]%
        {tum-vi}
\bibfield{author}{\bibinfo{person}{David Schubert}, \bibinfo{person}{Thore Goll}, \bibinfo{person}{Nikolaus Demmel}, \bibinfo{person}{Vladyslav Usenko}, \bibinfo{person}{J{\"o}rg St{\"u}ckler}, {and} \bibinfo{person}{Daniel Cremers}.} \bibinfo{year}{2018}\natexlab{}.
\newblock \showarticletitle{The {T}{U}{M} {V}{I} {B}enchmark for {E}valuating {V}isual-inertial {O}dometry}. In \bibinfo{booktitle}{\emph{IEEE/RSJ International Conference on Intelligent Robots and Systems (IROS)}}.
\newblock


\bibitem[Schwind et~al\mbox{.}(2019)]%
        {abstract-real}
\bibfield{author}{\bibinfo{person}{Valentin Schwind}, \bibinfo{person}{Pascal Knierim}, \bibinfo{person}{Nico Haas}, {and} \bibinfo{person}{Niels Henze}.} \bibinfo{year}{2019}\natexlab{}.
\newblock \showarticletitle{Using Presence Questionnaires in Virtual Reality}. In \bibinfo{booktitle}{\emph{Proceedings of the 2019 CHI Conference on Human Factors in Computing Systems}}.
\newblock


\bibitem[Schöps et~al\mbox{.}(2019)]%
        {keyframe-similar-badslam}
\bibfield{author}{\bibinfo{person}{Thomas Schöps}, \bibinfo{person}{Torsten Sattler}, {and} \bibinfo{person}{Marc Pollefeys}.} \bibinfo{year}{2019}\natexlab{}.
\newblock \showarticletitle{BAD SLAM: Bundle Adjusted Direct RGB-D SLAM}. In \bibinfo{booktitle}{\emph{2019 IEEE/CVF Conference on Computer Vision and Pattern Recognition (CVPR)}}. \bibinfo{pages}{134--144}.
\newblock


\bibitem[Scona et~al\mbox{.}(2018)]%
        {scona2018staticfusion}
\bibfield{author}{\bibinfo{person}{Raluca Scona}, \bibinfo{person}{Mariano Jaimez}, \bibinfo{person}{Yvan~R Petillot}, \bibinfo{person}{Maurice Fallon}, {and} \bibinfo{person}{Daniel Cremers}.} \bibinfo{year}{2018}\natexlab{}.
\newblock \showarticletitle{Staticfusion: Background reconstruction for dense rgb-d slam in dynamic environments}. In \bibinfo{booktitle}{\emph{IEEE international conference on robotics and automation (ICRA)}}.
\newblock


\bibitem[Semenova et~al\mbox{.}(2022)]%
        {system-level-quant-analysis-karthik-2022}
\bibfield{author}{\bibinfo{person}{Sofiya Semenova}, \bibinfo{person}{Steven~Y Ko}, \bibinfo{person}{Yu~David Liu}, \bibinfo{person}{Lukasz Ziarek}, {and} \bibinfo{person}{Karthik Dantu}.} \bibinfo{year}{2022}\natexlab{}.
\newblock \showarticletitle{A quantitative analysis of system bottlenecks in visual SLAM}. In \bibinfo{booktitle}{\emph{Proceedings of the 23rd Annual International Workshop on Mobile Computing Systems and Applications}}. \bibinfo{pages}{74--80}.
\newblock


\bibitem[Servi{\`e}res et~al\mbox{.}(2021)]%
        {servieres2021visual-vislam}
\bibfield{author}{\bibinfo{person}{Myriam Servi{\`e}res}, \bibinfo{person}{Val{\'e}rie Renaudin}, \bibinfo{person}{Alexis Dupuis}, {and} \bibinfo{person}{Nicolas Antigny}.} \bibinfo{year}{2021}\natexlab{}.
\newblock \showarticletitle{Visual and visual-inertial slam: State of the art, classification, and experimental benchmarking}.
\newblock \bibinfo{journal}{\emph{Journal of Sensors}}  \bibinfo{volume}{2021} (\bibinfo{year}{2021}), \bibinfo{pages}{1--26}.
\newblock


\bibitem[Skarbez et~al\mbox{.}(2020)]%
        {skarbez2020immersion}
\bibfield{author}{\bibinfo{person}{Richard Skarbez}, \bibinfo{person}{Frederick~P Brooks}, {and} \bibinfo{person}{Mary~C Whitton}.} \bibinfo{year}{2020}\natexlab{}.
\newblock \showarticletitle{Immersion and coherence: Research agenda and early results}.
\newblock \bibinfo{journal}{\emph{IEEE Transactions on Visualization and Computer Graphics}} \bibinfo{volume}{27}, \bibinfo{number}{10} (\bibinfo{year}{2020}).
\newblock


\bibitem[Slater et~al\mbox{.}(1996)]%
        {perfromance-presence}
\bibfield{author}{\bibinfo{person}{Mel Slater}, \bibinfo{person}{Vasilis Linakis}, \bibinfo{person}{Martin Usoh}, {and} \bibinfo{person}{Rob Kooper}.} \bibinfo{year}{1996}\natexlab{}.
\newblock \showarticletitle{Immersion, Presence and Performance in Virtual Environments: An Experiment with Tri-Dimensional Chess}. In \bibinfo{booktitle}{\emph{Proceedings of the ACM Symposium on Virtual Reality Software and Technology}}.
\newblock
\showISBNx{0897918258}


\bibitem[Sluganovic et~al\mbox{.}(2020)]%
        {man-in-the-middle-attack-sluganovic2020tap}
\bibfield{author}{\bibinfo{person}{Ivo Sluganovic}, \bibinfo{person}{Mihael Liskij}, \bibinfo{person}{Ante Derek}, {and} \bibinfo{person}{Ivan Martinovic}.} \bibinfo{year}{2020}\natexlab{}.
\newblock \showarticletitle{Tap-pair: Using spatial secrets for single-tap device pairing of augmented reality headsets}. In \bibinfo{booktitle}{\emph{ACM CDASP}}.
\newblock


\bibitem[Son et~al\mbox{.}(2015)]%
        {acoustic-dronesson2015rocking}
\bibfield{author}{\bibinfo{person}{Yunmok Son}, \bibinfo{person}{Hocheol Shin}, \bibinfo{person}{Dongkwan Kim}, \bibinfo{person}{Youngseok Park}, \bibinfo{person}{Juhwan Noh}, \bibinfo{person}{Kibum Choi}, \bibinfo{person}{Jungwoo Choi}, {and} \bibinfo{person}{Yongdae Kim}.} \bibinfo{year}{2015}\natexlab{}.
\newblock \showarticletitle{Rocking drones with intentional sound noise on gyroscopic sensors}. In \bibinfo{booktitle}{\emph{USENIX Security}}.
\newblock


\bibitem[Speicher et~al\mbox{.}(2019)]%
        {what-is-mr-chi}
\bibfield{author}{\bibinfo{person}{Maximilian Speicher}, \bibinfo{person}{Brian~D. Hall}, {and} \bibinfo{person}{Michael Nebeling}.} \bibinfo{year}{2019}\natexlab{}.
\newblock \showarticletitle{What is Mixed Reality?}. In \bibinfo{booktitle}{\emph{Proceedings of the CHI Conference on Human Factors in Computing Systems}}.
\newblock


\bibitem[Sualeh and Kim(2019)]%
        {survey-sualeh2019simultaneous}
\bibfield{author}{\bibinfo{person}{Muhammad Sualeh} {and} \bibinfo{person}{Gon-Woo Kim}.} \bibinfo{year}{2019}\natexlab{}.
\newblock \showarticletitle{Simultaneous localization and mapping in the epoch of semantics: a survey}.
\newblock \bibinfo{journal}{\emph{International Journal of Control, Automation and Systems}}  \bibinfo{volume}{17} (\bibinfo{year}{2019}), \bibinfo{pages}{729--742}.
\newblock


\bibitem[Taheri and Xia(2021)]%
        {taheri2021slam-survey}
\bibfield{author}{\bibinfo{person}{Hamid Taheri} {and} \bibinfo{person}{Zhao~Chun Xia}.} \bibinfo{year}{2021}\natexlab{}.
\newblock \showarticletitle{SLAM; definition and evolution}.
\newblock \bibinfo{journal}{\emph{Engineering Applications of Artificial Intelligence}}  \bibinfo{volume}{97} (\bibinfo{year}{2021}), \bibinfo{pages}{104032}.
\newblock


\bibitem[Taketomi et~al\mbox{.}(2017)]%
        {taketomi2017visual-survey-2010-to-2016}
\bibfield{author}{\bibinfo{person}{Takafumi Taketomi}, \bibinfo{person}{Hideaki Uchiyama}, {and} \bibinfo{person}{Sei Ikeda}.} \bibinfo{year}{2017}\natexlab{}.
\newblock \showarticletitle{Visual SLAM algorithms: A survey from 2010 to 2016}.
\newblock \bibinfo{journal}{\emph{IPSJ Transactions on Computer Vision and Applications}} \bibinfo{volume}{9}, \bibinfo{number}{1} (\bibinfo{year}{2017}), \bibinfo{pages}{1--11}.
\newblock


\bibitem[Tang et~al\mbox{.}(2020)]%
        {tang2020kp3d}
\bibfield{author}{\bibinfo{person}{Jiexiong Tang}, \bibinfo{person}{Rares Ambrus}, \bibinfo{person}{Vitor Guizilini}, \bibinfo{person}{Sudeep Pillai}, \bibinfo{person}{Hanme Kim}, \bibinfo{person}{Patric Jensfelt}, {and} \bibinfo{person}{Adrien Gaidon}.} \bibinfo{year}{2020}\natexlab{}.
\newblock \showarticletitle{{Self-Supervised 3D Keypoint Learning for Ego-Motion Estimation}}. In \bibinfo{booktitle}{\emph{Conference on Robot Learning (CoRL)}}.
\newblock


\bibitem[Teed and Deng(2021)]%
        {teed2021droid}
\bibfield{author}{\bibinfo{person}{Zachary Teed} {and} \bibinfo{person}{Jia Deng}.} \bibinfo{year}{2021}\natexlab{}.
\newblock \showarticletitle{Droid-slam: Deep visual slam for monocular, stereo, and rgb-d cameras}.
\newblock \bibinfo{journal}{\emph{Advances in Neural Information Processing Systems}}  \bibinfo{volume}{34} (\bibinfo{year}{2021}), \bibinfo{pages}{16558--16569}.
\newblock


\bibitem[Tian et~al\mbox{.}(2015)]%
        {tian2015influence-locomotion}
\bibfield{author}{\bibinfo{person}{Yang Tian}, \bibinfo{person}{Victor Gomez}, {and} \bibinfo{person}{Shugen Ma}.} \bibinfo{year}{2015}\natexlab{}.
\newblock \showarticletitle{Influence of two SLAM algorithms using serpentine locomotion in a featureless environment}. In \bibinfo{booktitle}{\emph{2015 IEEE International Conference on Robotics and Biomimetics (ROBIO)}}. IEEE, \bibinfo{pages}{182--187}.
\newblock


\bibitem[Trippel et~al\mbox{.}(2017)]%
        {walnut}
\bibfield{author}{\bibinfo{person}{Timothy Trippel}, \bibinfo{person}{Ofir Weisse}, \bibinfo{person}{Wenyuan Xu}, \bibinfo{person}{Peter Honeyman}, {and} \bibinfo{person}{Kevin Fu}.} \bibinfo{year}{2017}\natexlab{}.
\newblock \showarticletitle{{WALNUT}: {W}aging {D}oubt on the {I}ntegrity of {MEMS} {A}ccelerometers with {A}coustic {I}njection {A}ttacks}. In \bibinfo{booktitle}{\emph{IEEE European Symposium on Security and Privacy (EuroS\&P)}}.
\newblock


\bibitem[Trunfio et~al\mbox{.}(2022)]%
        {museums-trunfio2022mixed}
\bibfield{author}{\bibinfo{person}{Mariapina Trunfio}, \bibinfo{person}{Timothy Jung}, {and} \bibinfo{person}{Salvatore Campana}.} \bibinfo{year}{2022}\natexlab{}.
\newblock \showarticletitle{Mixed reality experiences in museums: Exploring the impact of functional elements of the devices on visitors’ immersive experiences and post-experience behaviours}. In \bibinfo{booktitle}{\emph{Information \& Management}}, Vol.~\bibinfo{volume}{59}. \bibinfo{publisher}{Elsevier}.
\newblock


\bibitem[Tsintotas et~al\mbox{.}(2022)]%
        {tsintotas2022revisiting-slam-loop-clouser}
\bibfield{author}{\bibinfo{person}{Konstantinos~A Tsintotas}, \bibinfo{person}{Loukas Bampis}, {and} \bibinfo{person}{Antonios Gasteratos}.} \bibinfo{year}{2022}\natexlab{}.
\newblock \showarticletitle{The revisiting problem in simultaneous localization and mapping: A survey on visual loop closure detection}.
\newblock \bibinfo{journal}{\emph{IEEE Transactions on Intelligent Transportation Systems}} \bibinfo{volume}{23}, \bibinfo{number}{11} (\bibinfo{year}{2022}).
\newblock


\bibitem[Uddin et~al\mbox{.}(2018)]%
        {uddin2018isir}
\bibfield{author}{\bibinfo{person}{Md Uddin}, \bibinfo{person}{Thanh~Trung Ngo}, \bibinfo{person}{Yasushi Makihara}, \bibinfo{person}{Noriko Takemura}, \bibinfo{person}{Xiang Li}, \bibinfo{person}{Daigo Muramatsu}, \bibinfo{person}{Yasushi Yagi}, {et~al\mbox{.}}} \bibinfo{year}{2018}\natexlab{}.
\newblock \showarticletitle{The ou-isir Large Population Gait Database with Real-life Carried Object and its Performance Evaluation}.
\newblock \bibinfo{journal}{\emph{IPSJ Transactions on Computer Vision and Applications}} \bibinfo{volume}{10}, \bibinfo{number}{1} (\bibinfo{year}{2018}), \bibinfo{pages}{1--11}.
\newblock


\bibitem[Ungureanu et~al\mbox{.}(2020)]%
        {ungureanu2020hololens2r}
\bibfield{author}{\bibinfo{person}{Dorin Ungureanu}, \bibinfo{person}{Federica Bogo}, \bibinfo{person}{S. Galliani}, \bibinfo{person}{Pooja Sama}, \bibinfo{person}{Xin Duan}, \bibinfo{person}{Casey Meekhof}, \bibinfo{person}{Jan St{\"u}hmer}, \bibinfo{person}{Thomas~J. Cashman}, \bibinfo{person}{Bugra Tekin}, \bibinfo{person}{Johannes~L. Sch{\"o}nberger}, \bibinfo{person}{Pawel Olszta}, {and} \bibinfo{person}{Marc Pollefeys}.} \bibinfo{year}{2020}\natexlab{}.
\newblock \showarticletitle{HoloLens 2 Research Mode as a Tool for Computer Vision Research}.
\newblock \bibinfo{journal}{\emph{arXiv preprint arXiv:2008.11239}} (\bibinfo{year}{2020}).
\newblock


\bibitem[Valente et~al\mbox{.}(2019)]%
        {valente2019improving-visual-challenges}
\bibfield{author}{\bibinfo{person}{J. Valente}, \bibinfo{person}{K. Bahirat}, \bibinfo{person}{K. Venechanos}, \bibinfo{person}{A.A. Cardenas}, {and} \bibinfo{person}{P. Balakrishnan}.} \bibinfo{year}{2019}\natexlab{}.
\newblock \showarticletitle{Improving the {S}ecurity of {V}isual {C}hallenges}.
\newblock \bibinfo{journal}{\emph{ACM Transactions on Cyber-Physical Systems (TCPS)}} (\bibinfo{year}{2019}).
\newblock


\bibitem[Vávra et~al\mbox{.}(2017)]%
        {surgery_review}
\bibfield{author}{\bibinfo{person}{Petr Vávra}, \bibinfo{person}{Jan Roman}, \bibinfo{person}{P. Zonča}, \bibinfo{person}{Peter Ihnát}, \bibinfo{person}{M. Němec}, \bibinfo{person}{Kumar Jayant}, \bibinfo{person}{Nagy Habib}, {and} \bibinfo{person}{Ahmed El-Gendi}.} \bibinfo{year}{2017}\natexlab{}.
\newblock \showarticletitle{Recent Development of Augmented Reality in Surgery: A Review}.
\newblock \bibinfo{journal}{\emph{Journal of Healthcare Engineering}} (\bibinfo{year}{2017}).
\newblock


\bibitem[Wang et~al\mbox{.}(2022a)]%
        {deep-vo-survey-}
\bibfield{author}{\bibinfo{person}{Ke Wang}, \bibinfo{person}{Sai Ma}, \bibinfo{person}{Junlan Chen}, \bibinfo{person}{Fan Ren}, {and} \bibinfo{person}{Jianbo Lu}.} \bibinfo{year}{2022}\natexlab{a}.
\newblock \showarticletitle{Approaches, Challenges, and Applications for Deep Visual Odometry: Toward Complicated and Emerging Areas}.
\newblock \bibinfo{journal}{\emph{IEEE Transactions on Cognitive and Developmental Systems}} \bibinfo{volume}{14}, \bibinfo{number}{1} (\bibinfo{year}{2022}).
\newblock


\bibitem[Wang et~al\mbox{.}(2017)]%
        {wang2017deepvo}
\bibfield{author}{\bibinfo{person}{Sen Wang}, \bibinfo{person}{Ronald Clark}, \bibinfo{person}{Hongkai Wen}, {and} \bibinfo{person}{Niki Trigoni}.} \bibinfo{year}{2017}\natexlab{}.
\newblock \showarticletitle{Deepvo: Towards end-to-end visual odometry with deep recurrent convolutional neural networks}. In \bibinfo{booktitle}{\emph{IEEE international conference on robotics and automation (ICRA)}}.
\newblock


\bibitem[Wang et~al\mbox{.}(2020)]%
        {tartanvo2020corl}
\bibfield{author}{\bibinfo{person}{Wenshan Wang}, \bibinfo{person}{Yaoyu Hu}, {and} \bibinfo{person}{Sebastian Scherer}.} \bibinfo{year}{2020}\natexlab{}.
\newblock \showarticletitle{TartanVO: A Generalizable Learning-based VO}.
\newblock  (\bibinfo{year}{2020}).
\newblock


\bibitem[Wang et~al\mbox{.}(2022b)]%
        {wang2022lf}
\bibfield{author}{\bibinfo{person}{Ze Wang}, \bibinfo{person}{Kailun Yang}, \bibinfo{person}{Hao Shi}, {and} \bibinfo{person}{Kaiwei Wang}.} \bibinfo{year}{2022}\natexlab{b}.
\newblock \showarticletitle{LF-VIO: A Visual-Inertial-Odometry Framework for Large Field-of-View Cameras with Negative Plane}.
\newblock \bibinfo{journal}{\emph{arXiv preprint arXiv:2202.12613}} (\bibinfo{year}{2022}).
\newblock


\bibitem[Warin and Reinhardt(2022)]%
        {washigntonxrsecurityap}
\bibfield{author}{\bibinfo{person}{Chris Warin} {and} \bibinfo{person}{Delphine Reinhardt}.} \bibinfo{year}{2022}\natexlab{}.
\newblock \showarticletitle{Vision: {U}sable {P}rivacy for {XR} in the {E}ra of the {M}etaverse}. In \bibinfo{booktitle}{\emph{European Symposium on Usable Security (EuroUSEC)}}.
\newblock


\bibitem[Wienrich et~al\mbox{.}(2021)]%
        {context}
\bibfield{author}{\bibinfo{person}{Carolin Wienrich}, \bibinfo{person}{Philipp Komma}, \bibinfo{person}{Stephanie Vogt}, {and} \bibinfo{person}{Marc~E. Latoschik}.} \bibinfo{year}{2021}\natexlab{}.
\newblock \showarticletitle{Spatial Presence in Mixed Realities–Considerations About the Concept, Measures, Design, and Experiments}.
\newblock \bibinfo{journal}{\emph{Frontiers in Virtual Reality}}  \bibinfo{volume}{2} (\bibinfo{year}{2021}).
\newblock


\bibitem[Yan et~al\mbox{.}(2022)]%
        {yan2022dgs-reflective-texture}
\bibfield{author}{\bibinfo{person}{Li Yan}, \bibinfo{person}{Xiao Hu}, \bibinfo{person}{Leyang Zhao}, \bibinfo{person}{Yu Chen}, \bibinfo{person}{Pengcheng Wei}, {and} \bibinfo{person}{Hong Xie}.} \bibinfo{year}{2022}\natexlab{}.
\newblock \showarticletitle{DGS-SLAM: A Fast and Robust RGBD SLAM in Dynamic Environments Combined by Geometric and Semantic Information}.
\newblock \bibinfo{journal}{\emph{Remote Sensing}} \bibinfo{volume}{14}, \bibinfo{number}{3} (\bibinfo{year}{2022}), \bibinfo{pages}{795}.
\newblock


\bibitem[Yarossi et~al\mbox{.}(2021)]%
        {context-of-adaptation}
\bibfield{author}{\bibinfo{person}{Mathew Yarossi}, \bibinfo{person}{Madhur Mangalam}, \bibinfo{person}{Stephanie Naufel}, {and} \bibinfo{person}{Eugene Tunik}.} \bibinfo{year}{2021}\natexlab{}.
\newblock \showarticletitle{Virtual Reality as a Context for Adaptation}.
\newblock \bibinfo{journal}{\emph{Frontiers in Virtual Reality}}  \bibinfo{volume}{2} (\bibinfo{year}{2021}).
\newblock


\bibitem[Younes et~al\mbox{.}(2017)]%
        {survey-keyframe-younes2017keyframe}
\bibfield{author}{\bibinfo{person}{Georges Younes}, \bibinfo{person}{Daniel Asmar}, \bibinfo{person}{Elie Shammas}, {and} \bibinfo{person}{John Zelek}.} \bibinfo{year}{2017}\natexlab{}.
\newblock \showarticletitle{Keyframe-based monocular SLAM: design, survey, and future directions}.
\newblock \bibinfo{journal}{\emph{Robotics and Autonomous Systems}}  \bibinfo{volume}{98} (\bibinfo{year}{2017}), \bibinfo{pages}{67--88}.
\newblock


\bibitem[Yu et~al\mbox{.}(2021b)]%
        {domain-adaptive-tracking2021PIC}
\bibfield{author}{\bibinfo{person}{Qian-Qian Yu}, \bibinfo{person}{Yi-Yang Wang}, \bibinfo{person}{Ke-Qi Fan}, {and} \bibinfo{person}{Yu-Hui Zheng}.} \bibinfo{year}{2021}\natexlab{b}.
\newblock \showarticletitle{Domain Adaptive Visual Tracking with Multi-scale Feature Fusion}. In \bibinfo{booktitle}{\emph{2021 IEEE International Conference on Progress in Informatics and Computing (PIC)}}.
\newblock


\bibitem[Yu et~al\mbox{.}(2021a)]%
        {yu2021visual-locomotion}
\bibfield{author}{\bibinfo{person}{Wenhao Yu}, \bibinfo{person}{Deepali Jain}, \bibinfo{person}{Alejandro Escontrela}, \bibinfo{person}{Atil Iscen}, \bibinfo{person}{Peng Xu}, \bibinfo{person}{Erwin Coumans}, \bibinfo{person}{Sehoon Ha}, \bibinfo{person}{Jie Tan}, {and} \bibinfo{person}{Tingnan Zhang}.} \bibinfo{year}{2021}\natexlab{a}.
\newblock \showarticletitle{Visual-locomotion: Learning to walk on complex terrains with vision}. In \bibinfo{booktitle}{\emph{5th Annual Conference on Robot Learning}}.
\newblock


\bibitem[{Zhan} et~al\mbox{.}(2020)]%
        {zhan2019dfvo}
\bibfield{author}{\bibinfo{person}{H. {Zhan}}, \bibinfo{person}{C.~S. {Weerasekera}}, \bibinfo{person}{J.~W. {Bian}}, {and} \bibinfo{person}{I. {Reid}}.} \bibinfo{year}{2020}\natexlab{}.
\newblock \showarticletitle{Visual Odometry Revisited: What Should Be Learnt?}. In \bibinfo{booktitle}{\emph{IEEE International Conference on Robotics and Automation (ICRA)}}.
\newblock


\bibitem[Zhang et~al\mbox{.}(2022)]%
        {smart-robot}
\bibfield{author}{\bibinfo{person}{Hui Zhang}, \bibinfo{person}{Li~Zhu Liu}, \bibinfo{person}{He Xie}, \bibinfo{person}{Yiming Jiang}, \bibinfo{person}{Jian Zhou}, {and} \bibinfo{person}{Yaonan Wang}.} \bibinfo{year}{2022}\natexlab{}.
\newblock \showarticletitle{Deep Learning-Based Robot Vision: High-End Tools for Smart Manufacturing}.
\newblock \bibinfo{journal}{\emph{IEEE IMM}}.
\newblock


\bibitem[Zhang and Sawchuk(2012)]%
        {zhang2012usc}
\bibfield{author}{\bibinfo{person}{Mi Zhang} {and} \bibinfo{person}{Alexander~A Sawchuk}.} \bibinfo{year}{2012}\natexlab{}.
\newblock \showarticletitle{USC-HAD: A daily activity dataset for ubiquitous activity recognition using wearable sensors}. In \bibinfo{booktitle}{\emph{Proceedings of the 2012 ACM conference on ubiquitous computing}}. \bibinfo{pages}{1036--1043}.
\newblock


\bibitem[Zhang et~al\mbox{.}(2021b)]%
        {zhang2021pose}
\bibfield{author}{\bibinfo{person}{Mingming Zhang}, \bibinfo{person}{Xingxing Zuo}, \bibinfo{person}{Yiming Chen}, \bibinfo{person}{Yong Liu}, {and} \bibinfo{person}{Mingyang Li}.} \bibinfo{year}{2021}\natexlab{b}.
\newblock \showarticletitle{Pose estimation for ground robots: On manifold representation, integration, reparameterization, and optimization}.
\newblock \bibinfo{journal}{\emph{IEEE Transactions on Robotics}} \bibinfo{volume}{37}, \bibinfo{number}{4} (\bibinfo{year}{2021}).
\newblock


\bibitem[Zhang et~al\mbox{.}(2021a)]%
        {rgbd-survey}
\bibfield{author}{\bibinfo{person}{Shishun Zhang}, \bibinfo{person}{Longyu Zheng}, {and} \bibinfo{person}{Wenbing Tao}.} \bibinfo{year}{2021}\natexlab{a}.
\newblock \showarticletitle{Survey and Evaluation of RGB-D SLAM}.
\newblock \bibinfo{journal}{\emph{IEEE Access}}  \bibinfo{volume}{9} (\bibinfo{year}{2021}).
\newblock


\bibitem[Zhao et~al\mbox{.}(2020)]%
        {zhao2020closedloop-system0benchmark-survey}
\bibfield{author}{\bibinfo{person}{Yipu Zhao}, \bibinfo{person}{Justin~S Smith}, \bibinfo{person}{Sambhu~H Karumanchi}, {and} \bibinfo{person}{Patricio~A Vela}.} \bibinfo{year}{2020}\natexlab{}.
\newblock \showarticletitle{Closed-loop benchmarking of stereo visual-inertial SLAM systems: Understanding the impact of drift and latency on tracking accuracy}. In \bibinfo{booktitle}{\emph{IEEE International Conference on Robotics and Automation (ICRA)}}.
\newblock


\bibitem[Zhou et~al\mbox{.}(2018)]%
        {ZUB18deeptam}
\bibfield{author}{\bibinfo{person}{H. Zhou}, \bibinfo{person}{B. Ummenhofer}, {and} \bibinfo{person}{T. Brox}.} \bibinfo{year}{2018}\natexlab{}.
\newblock \showarticletitle{DeepTAM: Deep Tracking and Mapping}. In \bibinfo{booktitle}{\emph{European Conference on Computer Vision (ECCV)}}.
\newblock


\bibitem[Zhou et~al\mbox{.}(2015)]%
        {zhou2015structslam}
\bibfield{author}{\bibinfo{person}{Huizhong Zhou}, \bibinfo{person}{Danping Zou}, \bibinfo{person}{Ling Pei}, \bibinfo{person}{Rendong Ying}, \bibinfo{person}{Peilin Liu}, {and} \bibinfo{person}{Wenxian Yu}.} \bibinfo{year}{2015}\natexlab{}.
\newblock \showarticletitle{Struct{SLAM}: {V}isual {SLAM} with {B}uilding {S}tructure {L}ines}.
\newblock \bibinfo{journal}{\emph{IEEE Transactions on Vehicular Technology}} \bibinfo{volume}{64}, \bibinfo{number}{4} (\bibinfo{year}{2015}).
\newblock


\bibitem[Zhou et~al\mbox{.}(2017)]%
        {sfmlearner-zhou2017unsupervised}
\bibfield{author}{\bibinfo{person}{Tinghui Zhou}, \bibinfo{person}{Matthew Brown}, \bibinfo{person}{Noah Snavely}, {and} \bibinfo{person}{David~G Lowe}.} \bibinfo{year}{2017}\natexlab{}.
\newblock \showarticletitle{Unsupervised learning of depth and ego-motion from video}. In \bibinfo{booktitle}{\emph{Proceedings of the IEEE conference on computer vision and pattern recognition}}. \bibinfo{pages}{1851--1858}.
\newblock


\bibitem[Zou et~al\mbox{.}(2019)]%
        {zou2019structvio}
\bibfield{author}{\bibinfo{person}{Danping Zou}, \bibinfo{person}{Yuanxin Wu}, \bibinfo{person}{Ling Pei}, \bibinfo{person}{Haibin Ling}, {and} \bibinfo{person}{Wenxian Yu}.} \bibinfo{year}{2019}\natexlab{}.
\newblock \showarticletitle{StructVIO: visual-inertial odometry with structural regularity of man-made environments}.
\newblock \bibinfo{journal}{\emph{IEEE Transactions on Robotics}} \bibinfo{volume}{35}, \bibinfo{number}{4} (\bibinfo{year}{2019}).
\newblock


\bibitem[Zubizarreta et~al\mbox{.}(2020)]%
        {DSM-Zubizarreta2020}
\bibfield{author}{\bibinfo{person}{Jon Zubizarreta}, \bibinfo{person}{Iker Aguinaga}, {and} \bibinfo{person}{Jose Maria~Martinez Montiel}.} \bibinfo{year}{2020}\natexlab{}.
\newblock \showarticletitle{Direct Sparse Mapping}.
\newblock \bibinfo{journal}{\emph{IEEE Transactions on Robotics}} (\bibinfo{year}{2020}).
\newblock


\bibitem[Zu{\~n}iga-No{\"e}l et~al\mbox{.}(2020)]%
        {zuniga2020vi}
\bibfield{author}{\bibinfo{person}{David Zu{\~n}iga-No{\"e}l}, \bibinfo{person}{Alberto Jaenal}, \bibinfo{person}{Ruben Gomez-Ojeda}, {and} \bibinfo{person}{Javier Gonzalez-Jimenez}.} \bibinfo{year}{2020}\natexlab{}.
\newblock \showarticletitle{The uma-vi dataset: Visual--inertial odometry in low-textured and dynamic illumination environments}.
\newblock \bibinfo{journal}{\emph{The International Journal of Robotics Research}} \bibinfo{volume}{39}, \bibinfo{number}{9} (\bibinfo{year}{2020}), \bibinfo{pages}{1052--1060}.
\newblock


\end{thebibliography}

\appendix

\end{document}